%% file: arxiv.tex
\documentclass[10pt,journal,compsoc]{IEEEtran}

\ifCLASSOPTIONcompsoc
  \usepackage[nocompress]{cite}
\else
  \usepackage{cite}
\fi

\ifCLASSINFOpdf

\else
\fi

\usepackage{pgfplots}
\pgfplotsset{compat=newest}
\usepackage{tikz}

\usepackage{times}
\usepackage{epsfig}
\usepackage{graphicx}
\usepackage{amsmath}
\usepackage{amssymb}
\usepackage{amsthm}

\usepackage[utf8x]{inputenc}
\usepackage{color}
\usepackage{xcolor}
\usepackage{booktabs,siunitx} 
\usepackage{makecell}
\usepackage{enumitem}
\usepackage[percent]{overpic}
\usepackage[caption=false]{subfig}
\usepackage{rotating}
\usepackage[misc]{ifsym}
\usepackage{algorithm}
\usepackage{algpseudocode}

\usepackage[pagebackref=true,breaklinks=true,colorlinks,bookmarks=false]{hyperref}

\makeatletter
\newcommand\AB@affilsepx{, \protect\Affilfont}
\makeatother

\if@mathematic
   
\else
   
\fi

\definecolor{gta5}                {RGB}{227,111, 38}
\definecolor{synthia}             {RGB}{128, 64,128}
\definecolor{acdc}                {RGB}{ 62,135,207}

\definecolor{road}                {RGB}{128, 64,128}
\definecolor{sidewalk}            {RGB}{244, 35,232}
\definecolor{building}            {RGB}{ 70, 70, 70}
\definecolor{wall}                {RGB}{102,102,156}
\definecolor{fence}               {RGB}{190,153,153}
\definecolor{pole}                {RGB}{153,153,153}
\definecolor{traffic light}       {RGB}{250,170, 30}
\definecolor{traffic sign}        {RGB}{220,220,  0}
\definecolor{vegetation}          {RGB}{107,142, 35}
\definecolor{terrain}             {RGB}{152,251,152}
\definecolor{sky}                 {RGB}{ 70,130,180}
\definecolor{person}              {RGB}{220, 20, 60}
\definecolor{rider}               {RGB}{255,  0,  0}
\definecolor{car}                 {RGB}{  0,  0,142}
\definecolor{truck}               {RGB}{  0,  0, 70}
\definecolor{bus}                 {RGB}{  0, 60,100}
\definecolor{train}               {RGB}{  0, 80,100}
\definecolor{motorcycle}          {RGB}{  0,  0,230}
\definecolor{bicycle}             {RGB}{119, 11, 32}
\definecolor{void}                {RGB}{  0,  0,  0}

\algnewcommand{\LineComment}[1]{\State \(\triangleright\) #1}

\newcommand\ver[1]{\rotatebox[origin=c]{90}{#1}}

\newcommand{\yes}{\checkmark}
\newcommand{\no}{$\times$}

\newcommand{\best}[1]{\textbf{#1}}

\newcommand{\PAR}[1]{\vskip4pt \noindent{\bf #1~}}

\begin{document}
\title{ACDC: The Adverse Conditions Dataset\\%
       with Correspondences\\%
       for Robust Semantic Driving Scene Perception}

\author{Christos~Sakaridis$^*$,
        Haoran~Wang$^*$,
        Ke Li,
        René~Zurbrügg,
        Arpit Jadon,
        Wim~Abbeloos,
        Daniel~Olmeda~Reino,
        Luc~Van~Gool,
        and~Dengxin Dai
\IEEEcompsocitemizethanks{\IEEEcompsocthanksitem C.~Sakaridis, K.~Li, and R.~Zurbrügg are with ETH Z\"urich, Switzerland.\protect\\
\IEEEcompsocthanksitem H.~Wang, A.~Jadon and D.~Dai are with Max Planck Institute for Informatics, Saarland Informatics Campus, Germany.\\
\IEEEcompsocthanksitem W.~Abbeloos and D.~Olmeda~Reino are with Toyota Motor Europe, Belgium.\\
\IEEEcompsocthanksitem L.~Van Gool is with INSAIT, Sofia University St.~Kliment Ohridski, Bulgaria.\\ 
\IEEEcompsocthanksitem C.~Sakaridis and H.~Wang have contributed equally.\\}
}

\markboth{IEEE Transactions on Pattern Analysis and Machine Intelligence,~Vol.~xx, No.~xx,  August~2025}%
{Sakaridis \MakeLowercase{\textit{et al.}}: ACDC: The Adverse Conditions Dataset with Correspondences for Robust Semantic Driving Scene Perception}

\IEEEtitleabstractindextext{%
\begin{abstract}
\input{sections/0_abstract}
\end{abstract}

}

\maketitle

\IEEEdisplaynontitleabstractindextext

\IEEEpeerreviewmaketitle

\ifCLASSOPTIONcompsoc
\IEEEraisesectionheading{\section{Introduction}\label{sec:intro}}
\else
\section{Introduction}
\label{sec:intro}
\fi

\input{sections/1_introduction}

\section{Related Work}
\label{sec:related}
\input{sections/2_related_works}

\section{ACDC Dataset}
\label{sec:dataset}
\input{sections/3_acdc_description}

\section{Normal-to-Adverse Adaptation}
\label{sec:n2a_adapt}
\input{sections/4_adapt2adverse}

\section{Sensor-Level Adaptation}
\label{sec:n2n_adapt}
\input{sections/5_adapt2normal}

\section{Supervised Learning on Adverse Conditions}
\label{sec:supervised}
\input{sections/6_supervised}

\section{Evaluation of Externally Pre-trained Models}
\label{sec:external}
\input{sections/7_external}

\section{Uncertainty-Aware Semantic Segmentation}
\label{sec:uncertain}
\input{sections/8_uncertainty}

\section{Conclusion}
\label{sec:conclusion}
\input{sections/9_conclusion}

\input{appendices}

\ifCLASSOPTIONcompsoc
  \section*{Acknowledgments}
\else
  \section*{Acknowledgment}
\fi

This work is funded by Toyota Motor Europe via the research project TRACE-Z\"urich. We thank Anton Obukhov and Yuhang Lu for their advice on running HRNet and DANet.

\ifCLASSOPTIONcaptionsoff
  \newpage
\fi

\bibliographystyle{IEEEtran}
\bibliography{refs}

\begin{IEEEbiography}[{\includegraphics[width=1in,clip,keepaspectratio]{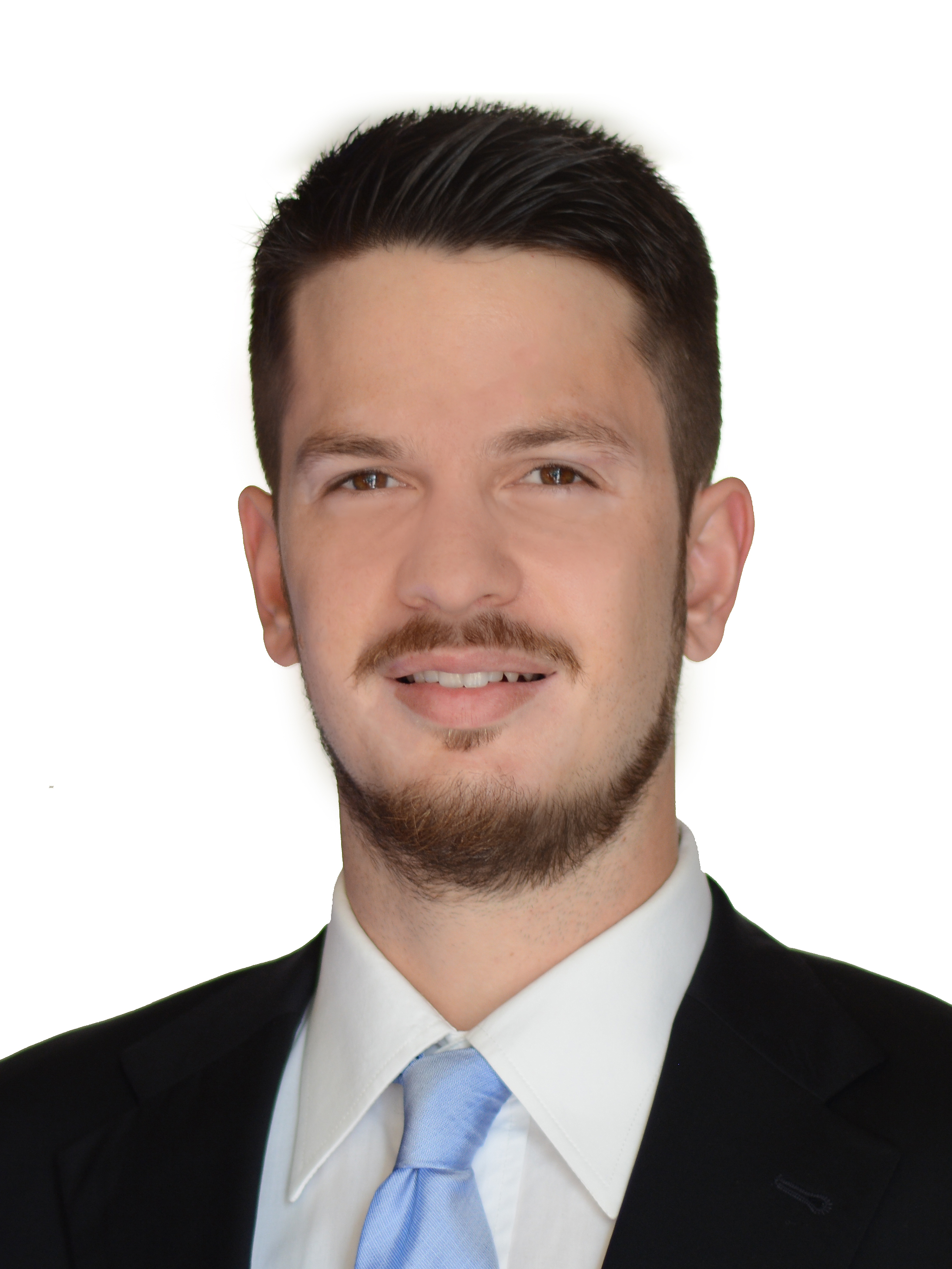}}]{Christos Sakaridis}
is a lecturer and Principal Investigator leading the Artificial Visual Intelligence group in the Photogrammetry and Remote Sensing lab of ETH Z\"urich. His research fields are computer vision, artificial intelligence, and machine learning. The focus of his research is on 3D and semantic visual perception, where he develops hybrid, data-driven yet informed, vision models and representations. Since 2021, he leads Toyota TRACE-Z\"urich, a large-scale project on computer vision for autonomous cars and robots. He received the ETH Zurich Career Seed Award in 2022. He obtained his PhD from ETH Z\"urich in 2021, having worked in Computer Vision Lab. Prior to that, he received his MSc in Computer Science from ETH Z\"urich in 2016 and his Diploma in Electrical and Computer Engineering from National Technical University of Athens in 2014.
\end{IEEEbiography}

\begin{IEEEbiography}[{\includegraphics[width=1in,clip,keepaspectratio]{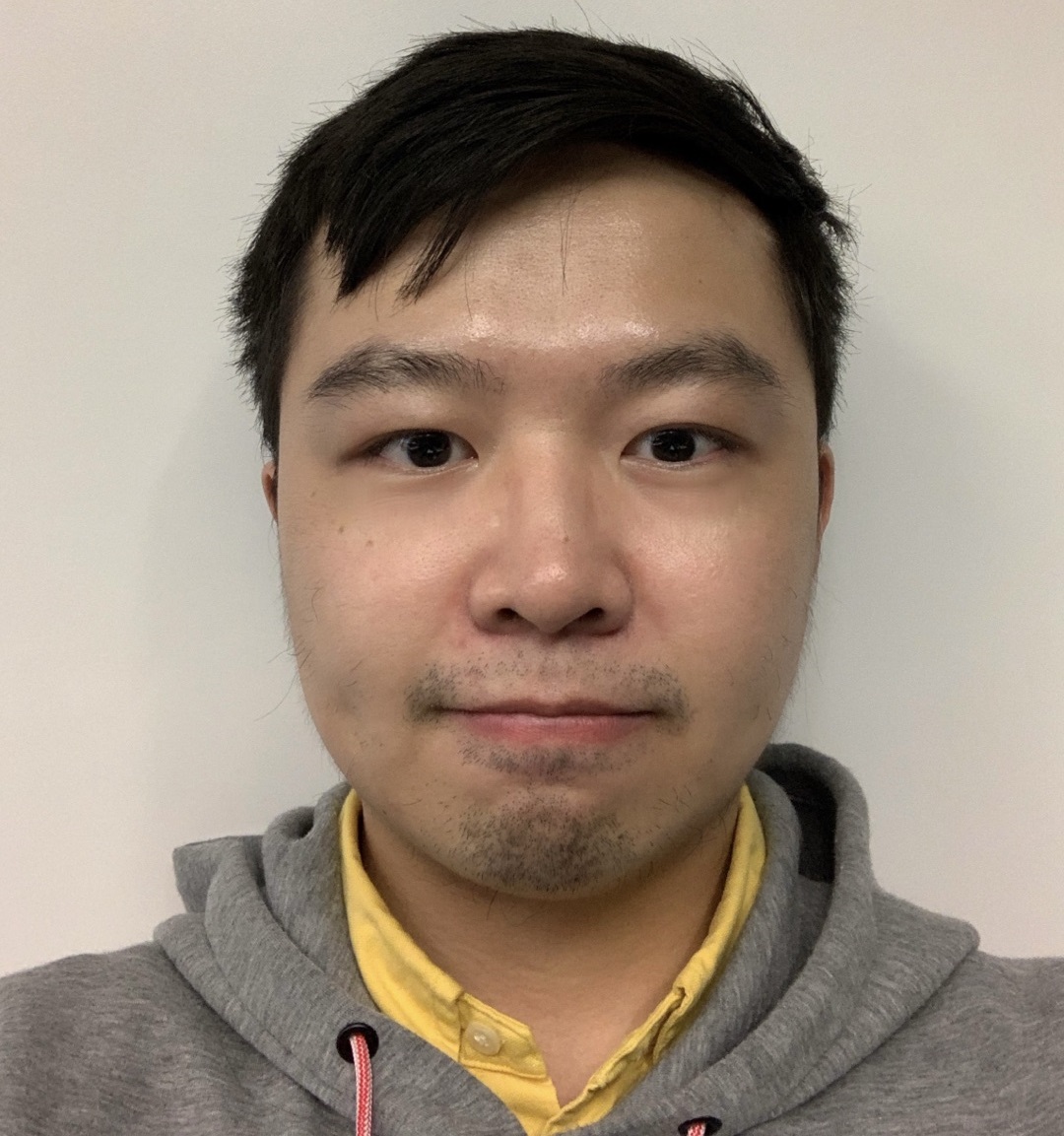}}]{Haoran Wang}
is a PhD student at Max Planck Institute for Informatics. He received his M.Sc degree from ETH Z\"urich in 2020 and B.Eng degree from Sichuan University in 2017. He has served as a reviewer for major computer vision conferences and journals (e.g. CVPR, ICCV, ECCV, IEEE T-PAMI). His research interests include machine learning and computer vision, especially transfer learning and domain adaptation. 
\end{IEEEbiography}

\begin{IEEEbiography}[{\includegraphics[width=1in,clip,keepaspectratio]{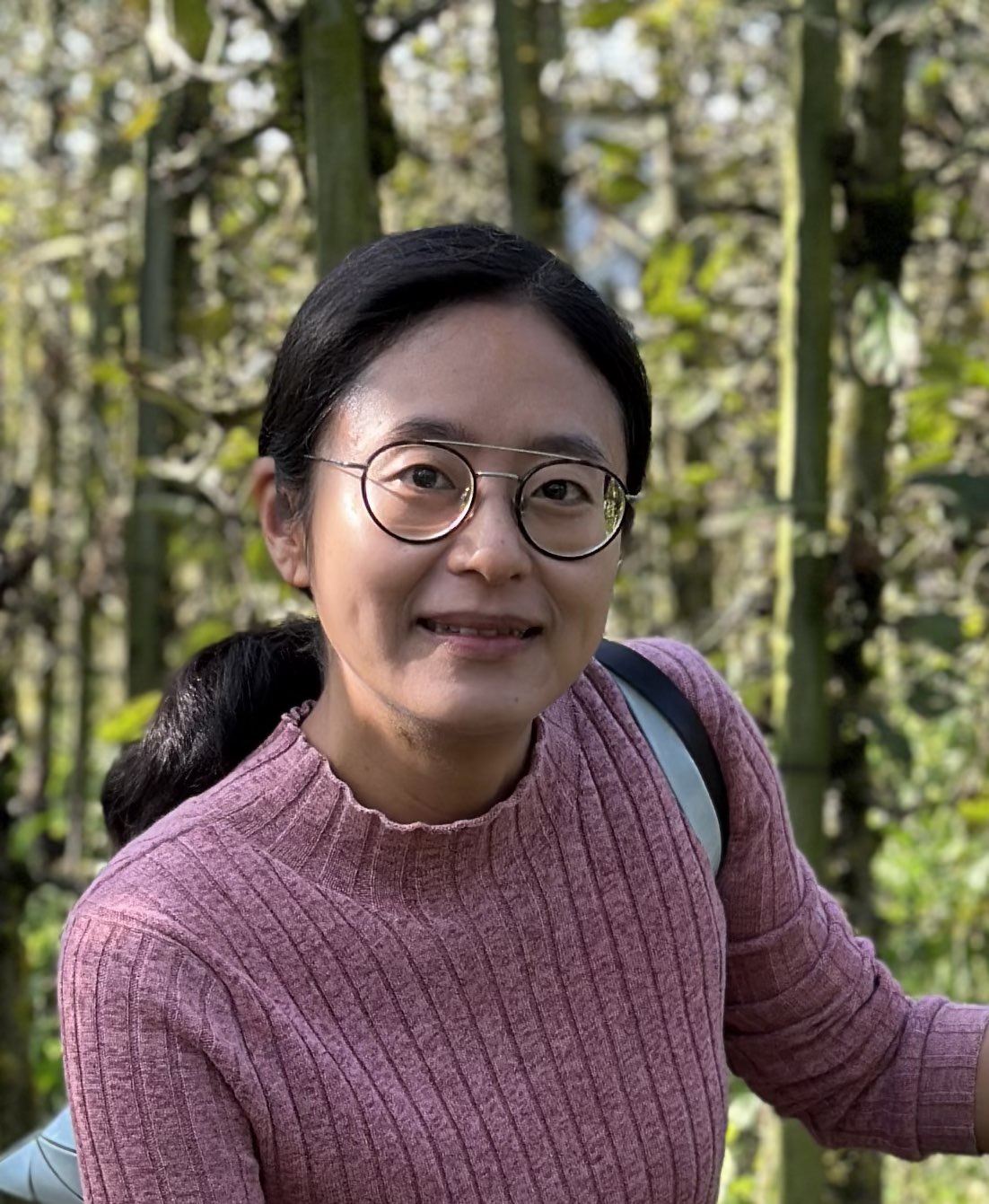}}]{Ke Li}
obtained her PhD in computer vision from ETH Z\"urich in 2024. Previously, she obtained her Bachelor's degree in Clinical Medicine from Tongji Medical School, Huazhong University of Science and Technology, in 2011. She then obtained a MSc in gene expression analysis from University of Basel in 2015 and a MSc in Biostatistics from University of Zurich in 2018. Her current research interests lie in hyperspectral imaging and super-resolution, learning with auxiliary tasks, and domain adaptation.
\end{IEEEbiography}

\begin{IEEEbiography}[{\includegraphics[width=1in,clip,keepaspectratio]{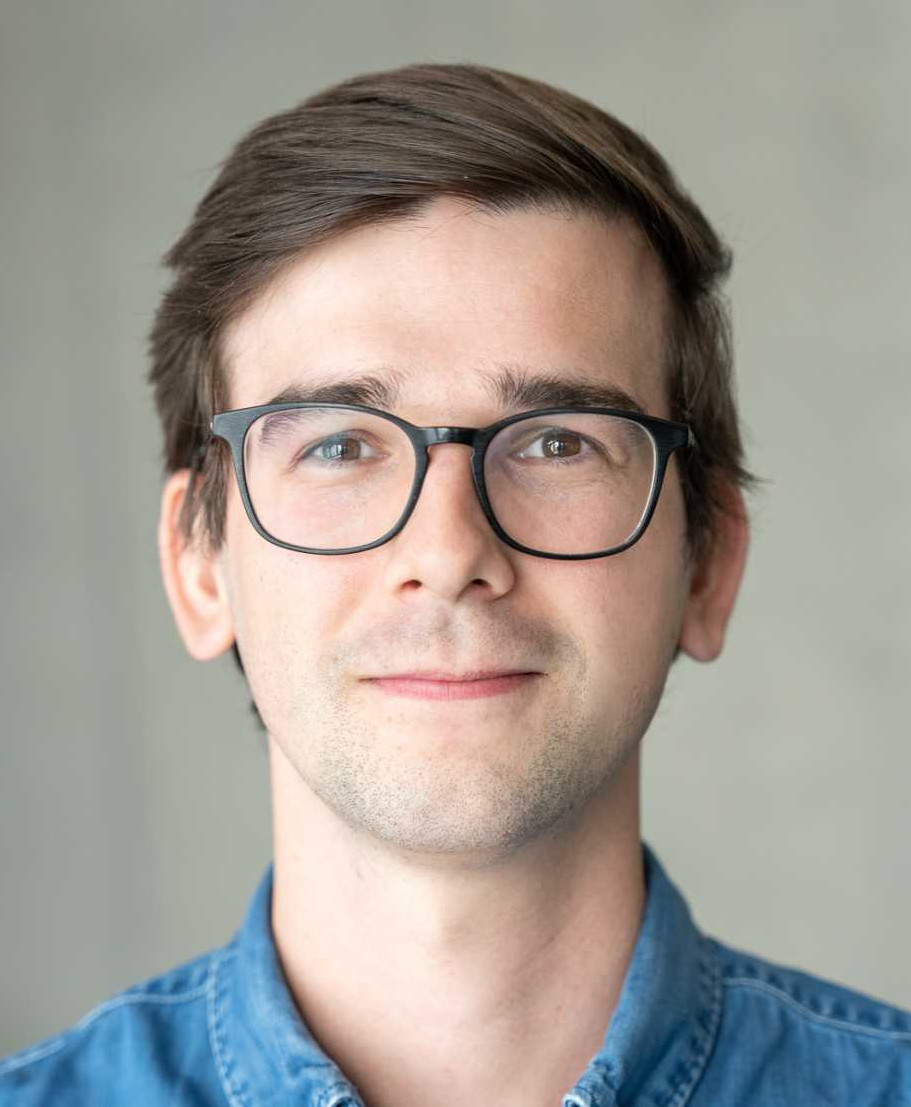}}]{René Zurbrügg}
is a PhD student at the ETH AI Center and part of the Robotic Systems Lab and Computer Vision and Geometry Group. He received his MSc in Robotics, Systems and Control in 2022 and his BSc in Information Technology and Electrical Engineering in 2021, both from ETH Z\"urich. His research interests lie in the field of embodied intelligence, focusing on the interconnection of computer vision and robotics applied to interactive tasks such as manipulation and active exploration.
\end{IEEEbiography}

\begin{IEEEbiography}[{\includegraphics[width=1in,clip,keepaspectratio]{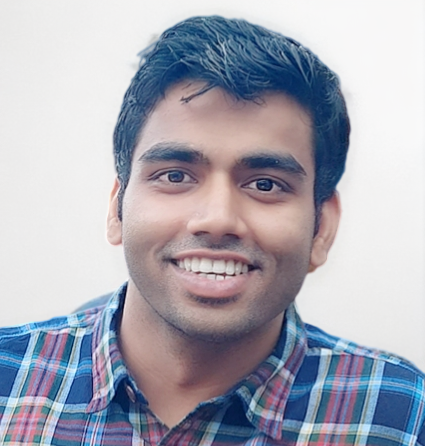}}]{Arpit Jadon}
is an AI Research Engineer at MPI for Informatics. He received his MSc in Computer Science from Saarland University in 2023 and his BSc in Electrical Engineering from Aligarh Muslim University in 2019, where he was also awarded for excellent holistic performance in academics and research. His primary research interests revolve around computer vision and machine learning, with a special focus on domain adaptation and synthetic data generation for autonomous driving.
\end{IEEEbiography}

\begin{IEEEbiography}[{\includegraphics[width=1in,clip,keepaspectratio]{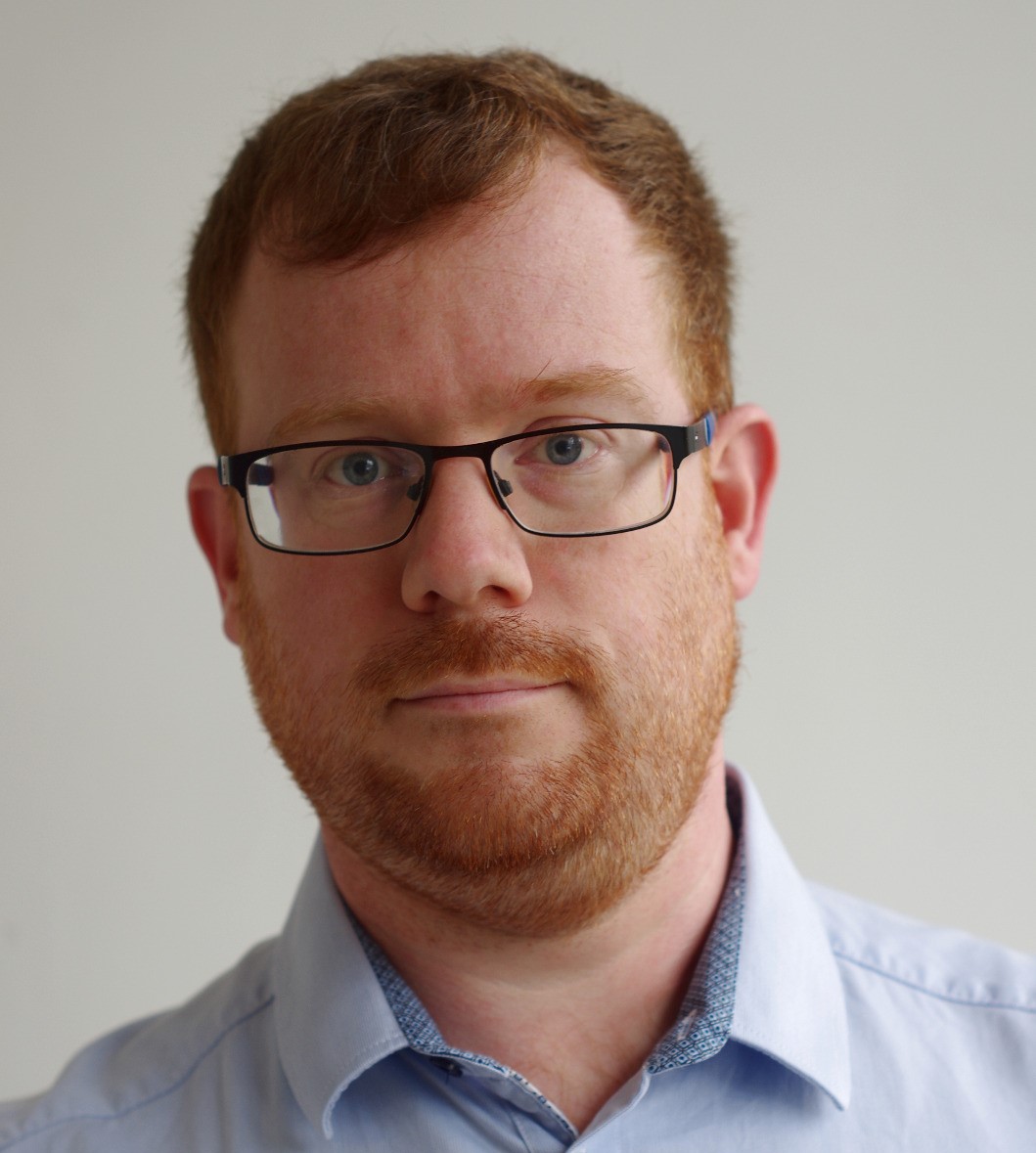}}]{Wim Abbeloos}
obtained a MSc in applied engineering at University of Antwerp in 2011. He then worked as a researcher and PhD student at the Industrial Vision Lab at University of Antwerp and the Embedded and Artificially Intelligent Vision Engineering group at KU Leuven. He joined Toyota Motor Europe (Belgium) in 2018, where he currently coordinates and manages research collaborations with top research institutes in Europe in computer vision and artificial intelligence, including automated driving.
\end{IEEEbiography}

\begin{IEEEbiography}[{\includegraphics[width=1in,clip,keepaspectratio]{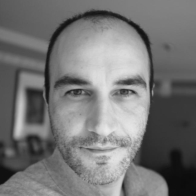}}]{Daniel Olmeda Reino}
received the PhD degree from UC3M, Spain, in 2014, in the field of computer vision. He joined Toyota Motor Europe in 2015, where he currently is a Technical Manager in the AI division. His research interests are in computer vision and machine learning. His latest work is on semi-supervised and continual learning, visual tracking and prediction, novelty detection, domain adaptation, visual localization, and robust perception among others.
\end{IEEEbiography}

\begin{IEEEbiography}[{\includegraphics[width=1in,clip,keepaspectratio]{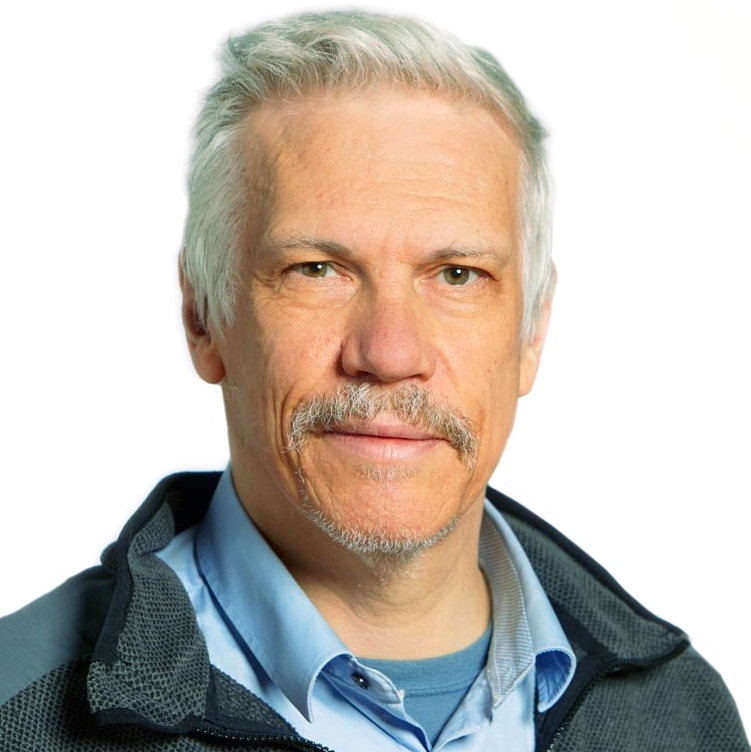}}]{Luc Van Gool}
is a full Professor for Computer Vision at INSAIT, Sofia University St.~Kliment Ohridski, and Professor Emeritus at ETH Z\"urich and the KU Leuven. His major research interests include 2D and 3D object recognition, texture analysis, range acquisition, stereo vision, robot vision, and optical flow. He received several best paper awards (e.g.\ David Marr Prize 1998, Best Paper CVPR 2007). He received the Koenderink Award in 2016 and the ``Distinguished Researcher'' nomination by the IEEE Computer Society in 2017. He was the holder of an ERC Advanced Grant. He has been a program committee member of several major computer vision conferences. He has led research on autonomous cars in the context of the Toyota TRACE labs at ETH and in Leuven, and has an extensive collaboration with Huawei on the topic of image and video enhancement.
\end{IEEEbiography}

\vfill

\newpage

\begin{IEEEbiography}[{\includegraphics[width=1in,clip,keepaspectratio]{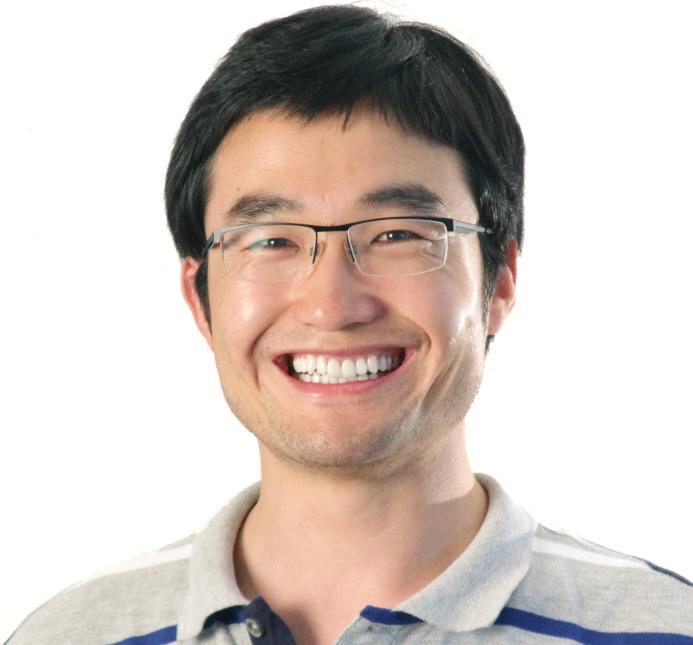}}]{Dengxin Dai}
has received his PhD degree from ETH Z\"urich, in 2016. He is Director of Research with Huawei Z\"urich Research Center. Previously, he was a group leader at MPI for Informatics and a senior researcher with ETH Z\"urich. He is a member of the ELLIS Society, an associate editor of the International Journal of Computer Vision, and has served as area chair of multiple vision conferences including CVPR, ECCV, and ICRA. He received the Golden Owl Award from ETH Z\"urich in 2021 for his exceptional teaching, and the German Pattern Recognition Award in 2022 for his outstanding contribution in the area of scalable and robust visual perception.

\end{IEEEbiography}

\end{document}

%% file: sections/0_abstract.tex
Level-5 driving automation requires a robust visual perception system that can parse input images under \emph{any} condition. However, existing driving datasets for dense semantic perception are either dominated by images captured under normal conditions or are small in scale. To address this, we introduce ACDC, the Adverse Conditions Dataset with Correspondences for training and testing methods for diverse semantic perception tasks on adverse visual conditions. ACDC consists of a large set of 8012 images, half of which (4006) are equally distributed between four common adverse conditions: fog, nighttime, rain, and snow. Each adverse-condition image comes with a high-quality pixel-level panoptic annotation, a corresponding image of the same scene under normal conditions, and a binary mask that distinguishes between intra-image regions of clear and uncertain semantic content. 1503 of the corresponding normal-condition images feature panoptic annotations, raising the total annotated images to 5509. ACDC supports the standard tasks of semantic segmentation, object detection, instance segmentation, and panoptic segmentation, as well as the newly introduced uncertainty-aware semantic segmentation. A detailed empirical study demonstrates the challenges that the adverse domains of ACDC pose to state-of-the-art supervised and unsupervised approaches and indicates the value of our dataset in steering future progress in the field. Our dataset and benchmark are publicly available at \url{https://acdc.vision.ee.ethz.ch}.

%% file: sections/1_introduction.tex
\begin{figure*}[!tb]
    \centering
    \begin{tikzpicture}
    \tikzstyle{every node}=[font=\footnotesize]
    \begin{axis}[
      ybar,
      ymode=log,
      width=\textwidth,
      height=4.5cm,
      xmin=0,
      xmax=26,
      ymin=5e5,
      ymax=3e10,
      ymajorgrids=true,
      ylabel={number of pixels},
      ytick={1e6,1e8,1e10},
      yticklabels={$10^6$,$10^8$,$10^{10}$,},
      xtick={1.5,5,8.5,13.5,18,21,24.5},
      minor xtick={3,7,10,17,19,23},
      xticklabels = {
        flat,
        construction,
        nature,
        vehicle,
        sky,
        object,
        human,
      },
      major x tick style = {opacity=0},
      minor x tick num = 1,
      xtick pos=left,
      every node near coord/.append style={
      anchor=west,
      rotate=90,
      font=\scriptsize,
      }
    ]
    
    \addplot[bar shift=0pt,draw=road,          fill opacity=0.9,fill=road!80!white           , nodes near coords=road                 ] plot coordinates{ ( 1,     2197870968  ) };
    \addplot[bar shift=0pt,draw=sidewalk,      fill opacity=0.8,fill=sidewalk!80!white       , nodes near coords=sidewalk             ] plot coordinates{ ( 2,     639279568   ) };

    \addplot[bar shift=0pt,draw=building,      fill opacity=0.8,fill=building!80!white       , nodes near coords=build.               ] plot coordinates{ ( 4,     1947561122  ) };
    \addplot[bar shift=0pt,draw=wall,          fill opacity=0.8,fill=wall!80!white           , nodes near coords=wall                 ] plot coordinates{ ( 5,     178501769  ) };
    \addplot[bar shift=0pt,draw=fence,         fill opacity=0.8,fill=fence!80!white          , nodes near coords=fence                ] plot coordinates{ ( 6,     167445508   ) };

    \addplot[bar shift=0pt,draw=vegetation,    fill opacity=0.8,fill=vegetation!80!white     , nodes near coords=veget.               ] plot coordinates{ ( 8,    1966261640   ) };
    \addplot[bar shift=0pt,draw=terrain,       fill opacity=0.8,fill=terrain!80!white        , nodes near coords=terrain              ] plot coordinates{ ( 9,    184160238    ) };

    \addplot[bar shift=0pt,draw=car,           fill opacity=0.8,fill=car!80!white            , nodes near coords=car           ] plot coordinates{ ( 11,    228846867   ) };
    \addplot[bar shift=0pt,draw=train,         fill opacity=0.8,fill=train!80!white          , nodes near coords=train         ] plot coordinates{ ( 12,    62226283        ) };
    \addplot[bar shift=0pt,draw=truck,         fill opacity=0.8,fill=truck!80!white          , nodes near coords=truck         ] plot coordinates{ ( 13,    29791566     ) };
    \addplot[bar shift=0pt,draw=bus,           fill opacity=0.8,fill=bus!80!white            , nodes near coords=bus           ] plot coordinates{ ( 14,    22978630     ) };
    \addplot[bar shift=0pt,draw=bicycle,       fill opacity=0.8,fill=bicycle!80!white        , nodes near coords=bicycle       ] plot coordinates{ ( 15,    10229698      ) };
    \addplot[bar shift=0pt,draw=motorcycle,    fill opacity=0.8,fill=motorcycle!80!white     , nodes near coords=motorcycle    ] plot coordinates{ ( 16,    4686652        ) };

    \addplot[bar shift=0pt,draw=sky,           fill opacity=0.8,fill=sky!80!white            , nodes near coords=sky                  ] plot coordinates{ ( 18,    2875662970 ) };

    \addplot[bar shift=0pt,draw=pole,          fill opacity=0.8,fill=pole!80!white           , nodes near coords=pole                 ] plot coordinates{ ( 20,    99905141   ) };
    \addplot[bar shift=0pt,draw=traffic sign,  fill opacity=0.8,fill=traffic sign!80!white   , nodes near coords=traffic sign         ] plot coordinates{ ( 21,    34913387   ) };
    \addplot[bar shift=0pt,draw=traffic light, fill opacity=0.8,fill=traffic light!80!white  , nodes near coords=traffic light        ] plot coordinates{ ( 22,    20034562    ) };

    \addplot[bar shift=0pt,draw=person,        fill opacity=0.8,fill=person!80!white         , nodes near coords=person        ] plot coordinates{ ( 24,    24005243     ) };
    \addplot[bar shift=0pt,draw=rider,         fill opacity=0.8,fill=rider!80!white          , nodes near coords=rider         ] plot coordinates{ ( 25,    3071170        ) };

    \end{axis}
    \end{tikzpicture}
    \vspace{-0.5cm}
    \caption{Number of finely annotated pixels per class in ACDC.}
    \label{fig:dataset:stats}
\end{figure*}

Most of the prominent large-scale image-based datasets for driving scene perception, including Cityscapes~\cite{Cityscapes}, Vistas~\cite{Mapillary} and KITTI~\cite{kitti}, are dominated by images captured under normal visual conditions, i.e., at daytime and in clear weather. Yet, vision applications such as automated driving impose a strict requirement on perception algorithms to maintain satisfactory performance in adverse domains. Although there have been efforts to include adverse visual domains in large-scale datasets, such as Oxford RobotCar~\cite{Oxford} and BDD100K~\cite{BDD100K}, these efforts focus either on localization/mapping tasks~\cite{Oxford,4Seasons} or on recognition tasks which \emph{do not involve dense pixel-level outputs}, such as object detection~\cite{BDD100K,Waymo:Open:Dataset,nuScenes}. For instance, while a notable 40\% of the object detection set of BDD100K pertains to nighttime, only 3\% of the images in its 10K semantic segmentation set, namely 345 images, are captured at nighttime~\cite{MGCDA_UIoU}. Thus, at the time when the conference version of this article appeared, there was no large adverse-condition pixel-level segmentation dataset available. In addition, the pixel-level annotation process for adverse-condition images is kept identical in~\cite{BDD100K,wilddash} to the normal-condition case, which leads to errors in the ground truth and renders it unreliable~\cite{MGCDA_UIoU}. The application of highly automated annotation pipelines with little human intervention, which has recently catalyzed the creation of large-scale pixel-level segmentation data~\cite{kirillov2023sam}, to adverse-condition images is thus especially prone to creating severe annotation errors, which is detrimental for model training, let alone rigorous evaluation, which requires highly accurate ground truth as reference. In contrast, seminal previous work~\cite{Cityscapes} has underlined the need for \emph{specialized} techniques and datasets for pixel-level semantic perception in adverse visual conditions, due to the inherent aleatory uncertainty in images captured in such conditions. These render entire image regions indiscernible even for humans.

ACDC constitutes precisely a response to this need for a large-scale labeled driving segmentation dataset specialized to adverse conditions, in terms of (i) size, (ii) domain adversity, and (iii) featured tasks. ACDC includes 5509 images with \emph{high-quality} pixel-level panoptic annotations. From this complete set of images, 4006 images are distributed equally among four common adverse conditions in real-world driving environments, namely fog, nighttime, rain, and snow, and the rest 1503 pertain to normal conditions, i.e.\ daytime and clear weather, thus granting ACDC a scale that slightly exceeds that of Cityscapes. The adverse-condition part of the dataset was deliberately recorded with the respective adverse conditions clearly present. Thus, a large domain shift from the normal clear-weather daytime conditions was achieved. Moreover, for each adverse-condition image, a corresponding normal-condition image of the same scene from approximately the same viewpoint is provided, intended for use by weakly supervised methods.

As to the tasks that our dataset supports, apart from standard dense semantic perception tasks such as semantic segmentation, object detection, instance segmentation and panoptic segmentation, we add the task of uncertainty-aware semantic segmentation. For the latter we introduce a specialized annotation protocol and a dedicated performance metric, termed average uncertainty-aware IoU (AUIoU). The key characteristic of uncertainty-aware semantic segmentation is the principled inclusion of image regions with indiscernible semantic content---\emph{invalid} regions---in annotation and evaluation. In particular, the annotation protocol for our adverse-condition images leverages privileged information in the form of the corresponding normal-condition images and the original adverse-condition videos, which enables us to \emph{reliably} assign legitimate semantic labels to invalid regions and to include them in the evaluation both for the aforementioned standard semantic perception tasks and for uncertainty-aware semantic segmentation. For the latter task, the separation of labeled pixels into invalid and valid is encoded in a binary mask. While both tasks require a hard semantic prediction, the uncertainty-aware task additionally expects a confidence map prediction. AUIoU is designed to take into account both the semantic and the confidence prediction and to reward predictions with low confidence on invalid pixels and high confidence on valid pixels. The requirement for an additional confidence prediction is relevant for safety-oriented applications, as it can help the downstream decision-making system avoid the fatal consequences of a low-confidence prediction being false, e.g.\ when a pedestrian is missed.

Compared to previous related datasets, ACDC thus boasts: (i) several different conditions, all evenly represented, instead of a single condition or a highly skewed distribution over conditions, (ii) a high representation of adversities in the set instead of a bias towards normal conditions, (iii) large-scale adverse-condition data instead of small-scale ones, (iv) reliable high-quality large-scale annotations in adverse conditions instead of unreliable ones~\cite{BDD100K} in this setting, (v) legitimate labels even for invalid regions which are indiscernible even for humans thanks to privileged correspondence-based information instead of mere exclusion of these regions from the ground truth. A quantitative comparison to other adverse-condition datasets across these axes is deferred to Sec.~\ref{sec:dataset:comparison}.

Apart from being a challenging benchmark for supervised semantic perception approaches, ACDC is a well-suited test bed for domain adaptation. A multitude of recent works have focused on unsupervised domain adaptation (UDA) for semantic segmentation~\cite{curriculum:domain:adaptation:17,cyCADA,synthetic:semantic:segmentation,FCNs:adaptation,adapt:structured:output:cvpr18,chen2018road,DCAN:adaptation,self:training:adaptation,advent:adaptation,category:adversaries:adaptation,bidirectional:learning:adaptation,patch:align:adaptation,crst:adaptation,fda:adaptation,textda:adaptation,sim:adaptation,mrnet:rectifying} and object detection~\cite{DomainAdaptiveFasterRCNN,chen2021scale,mic:adaptation,li2023sigma_plus_plus,hsu2020epm,li2022sigma}, but most of them are validated only on artificial synthetic-to-real or real-to-synthetic settings, using GTA5~\cite{playing:data} and SYNTHIA~\cite{Synthia:dataset} as source datasets and Cityscapes~\cite{Cityscapes} as the target dataset for semantic segmentation and object detection or Cityscapes as the source dataset and Foggy Cityscapes~\cite{SFSU_synthetic} as the target dataset for object detection. The \emph{real-world normal-to-adverse domain adaptation} scenario for semantic segmentation and object detection, which is much more relevant for real-world deployment of autonomous cars due to the difficulty of both acquiring and annotating adverse-condition data, has largely been overlooked. In particular, prior to the initial, conference version of ACDC~\cite{ACDC}, much fewer works had considered normal-to-adverse adaptation in their experiments~\cite{SFSU_synthetic,SynRealDataFogECCV18,daytime:2:nighttime,GCMA_UIoU:v1,MGCDA_UIoU,rainy:night:scene:understanding,input:adapters:adaptation} and whenever they did, they either restricted the target adverse domain to a single condition, e.g.\ nighttime~\cite{GCMA_UIoU:v1,MGCDA_UIoU,daytime:2:nighttime}, fog~\cite{SFSU_synthetic,SynRealDataFogECCV18}, or rain~\cite{rainy:night:scene:understanding}, or did not include a quantitative evaluation on the real target domain altogether~\cite{input:adapters:adaptation}. We attribute this fragmentation of normal-to-adverse adaptation works to the absence of a general large-scale dataset for semantic perception that evenly covers the majority of common adverse conditions and provides reliable ground truth for a sound evaluation in such challenging domains. ACDC answers exactly the need for such a dataset and has already served since its initial release as a test bed for unsupervised and weakly supervised domain adaptation~\cite{daformer:domain:adaptation,sepico:adaptation,hrda:domain:adaptation,sakaridis2023ciss,refign:domain:adaptation,mic:adaptation,CMA}. Experiments such as Cityscapes$\to$ACDC adaptation are straightforward thanks to the identical label sets of the two datasets, which facilitates validation of new domain adaptation approaches in the normal-to-adverse setting. We further introduce a novel domain adaptation setting from Cityscapes to the normal-condition part of ACDC, which isolates the sensor-level shift as the only difference in source and target domains given the geographical similarity of the two datasets, and establish a new UDA benchmark based on this setting.

Overall, we experiment with ACDC on all five semantic perception tasks it supports in four main directions: (i) unsupervised and weakly supervised normal-to-adverse and sensor-level domain adaptation, (ii) supervised semantic perception in adverse conditions, (iii) evaluation of models externally pre-trained on normal conditions, and (iv) evaluation of uncertainty-aware semantic segmentation baselines and oracles. Results show that access to ground-truth annotations under adverse conditions is indispensable for achieving high performance, as pre-trained models severely deteriorate under adverse conditions. Moreover, the real-world Cityscapes$\to$ACDC adaptation scenario stands for a challenging setting for all state-of-the-art UDA methods, which still trail fully supervised counterparts. This underlines the need for UDA methods that perform better when handling adverse target domains and highlights the importance of ACDC in steering future work in this direction. Finally, the uncertainty-aware annotations of ACDC create significant room for improvement over simple confidence prediction baselines and help promote future work on semantic segmentation that simultaneously models uncertainty.

An earlier version of this work has appeared in the International Conference on Computer Vision~\cite{ACDC}. Compared to the conference version, this paper makes the following additional contributions:
\begin{enumerate}
    \item A substantial amount of new annotations to the dataset, including (i) the upgrade of the initial 4006 semantic segmentation annotations of the adverse-condition images of ACDC to panoptic segmentation annotations and (ii) the annotation of 1503 normal-condition images, which were not annotated at all in the conference version, for panoptic segmentation.
    \item An extensive set of experimental comparisons on the newly supported tasks compared to the conference version, i.e.\ object detection, instance segmentation, and panoptic segmentation, covering diverse settings such as normal-to-adverse domain adaptation, sensor-level domain adaptation, supervised learning on adverse conditions, and evaluation of models externally pre-trained on normal conditions.
    \item An updated set of experimental comparisons on tasks and settings, such as normal-to-adverse domain adaptation for semantic segmentation and supervised semantic segmentation, which were already included in the conference version, taking into account respective recent state-of-the-art methods that have been presented since the publication of the conference version.
    \item Other enhanced and updated parts, such as (i) additional statistics and comparisons for dataset annotations based on the new format and the increased scale of the annotations, and (ii) an extended and updated overview of the related work, which covers the newly supported tasks and the latest advances in driving datasets for semantic perception and in adaptation of semantic segmentation.
\end{enumerate}

%% file: sections/2_related_works.tex
\PAR{Datasets for driving scene perception} include real-world and synthetic sets that support geometric and recognition tasks. KITTI~\cite{kitti} and Cityscapes~\cite{Cityscapes} pioneered this area with LiDAR and semantic image annotations, respectively. Subsequent datasets mostly aimed at increasing the scale~\cite{ApolloScape}, diversity~\cite{Mapillary} and number of tasks~\cite{BDD100K}. As high-quality pixel-level annotations proved hard to acquire~\cite{Cityscapes,Mapillary}, another line of work focused on creating synthetic sets at an even larger scale~\cite{Synthia:dataset,playing:data,playing:for:benchmarks,driving:in:the:matrix,SHIFT} and in which ground truth is automatically generated, as well as translating real datasets to adverse conditions such as fog or rain~\cite{SFSU_synthetic,SynRealDataFogECCV18,physics:based:rain:rendering}. Oxford Robotcar~\cite{Oxford} was the first real-world large-scale dataset in which adverse visual conditions such as nighttime, rain and snow were significantly represented, but it did not feature semantic annotations. While more recent large-scale sets~\cite{seeing:through:fog,CADC} that cover adverse conditions, such as Waymo Open~\cite{Waymo:Open:Dataset} and nuScenes~\cite{nuScenes}, include bounding boxes, they still lack dense pixel-level semantic annotations, which are vital for real-world autonomous agents~\cite{does:vision:matter:for:action} and are featured by ACDC. BDD100K~\cite{BDD100K} was the only exception to this rule prior to the publication of the conference version~\cite{ACDC} of this paper, with ca.\ 13\% of its 10000 pixel-level annotations pertaining to adverse conditions but containing severe errors~\cite{MGCDA_UIoU}, contrary to the high-quality reliable ground truth of ACDC. At the same time, only a small portion of each of the 1881 adverse-condition images in ADUULM~\cite{ADUULM} is annotated, whereas 92.5\% of the pixels in ACDC are annotated. On the other hand, several sets with small-scale pixel-level annotations covering adverse conditions~\cite{wilddash} had been presented, focusing on fog~\cite{SFSU_synthetic,CMAda:IJCV2020}, nighttime~\cite{daytime:2:nighttime,MGCDA_UIoU}, and rain~\cite{raincouver}. A notable case is Dark Zurich~\cite{MGCDA_UIoU}, with 201 fine pixel-level nighttime annotations and a dedicated annotation protocol and evaluation metric that handles regions with ambiguous content. The initial, conference version of ACDC improved both upon BDD100K, in terms of ground truth quality, and Dark Zurich, in terms of scale and condition diversity, featuring 5509 high-quality fine instance-level semantic annotations in which fog, night, rain, snow, and normal conditions are evenly represented. Since the initial publication of ACDC, a larger-scale version of WildDash, namely Wilddash2~\cite{Wilddash2}, has been released. The present extended version of ACDC exceeds the scale of Wilddash2 annotations (5509 vs.\ 5000). Moreover, in contrast both to Wilddash2 and to other recent dense semantic perception datasets with adverse conditions~\cite{alibeigi2023zenseact,mei2022waymo}, ACDC features cross-condition image-level correspondences with normal-condition reference images as well as a specialized annotation protocol which hinges exactly on the aforementioned correspondences to allow the assignment of legitimate, reliable semantic labels to indiscernible image regions which would otherwise be impossible to label.

\PAR{Semantic segmentation} has progressed rapidly over the last years, primarily through the design of convolutional neural networks. Based on fully convolutional architectures~\cite{FCNs:segmentation}, seminal works introduced atrous convolution~\cite{DeepLab,DeepLab:v2,dilated:convolution} and encoder-decoder structures with skip connections~\cite{UNet} to exploit context and improve localization, respectively. Balancing between global and local information was further addressed by parallel branches of different resolutions~\cite{refinenet,FRRN} and global pooling~\cite{pspnet}. Other works focused on real-time performance~\cite{BiSeNet}, leveraging different modalities such as depth~\cite{PADNet}, and defining neighborhood-based supervision~\cite{AAF} for segmentation. The current state of the art includes i.a.\ DeepLabv3+~\cite{DeepLab:v3+} and ANN~\cite{ANN:segmentation} with pyramid pooling modules, DANet~\cite{DANet} and CCNet~\cite{CCNet} with attention mechanisms, and HRNet~\cite{hrnet} and OCR~\cite{OCR} with high-resolution representations. While performance on the popular Cityscapes benchmark is increasingly saturating, we demonstrate that state-of-the-art methods achieve much lower performance on ACDC (see Sec.~\ref{sec:supervised:semseg}). Thus, ACDC provides a more challenging benchmark for semantic segmentation thanks to the adversity of its domains and is therefore able to foster further progress in the field.

\begin{figure*}
    \centering
    \includegraphics[clip,width=\textwidth,trim=2mm 123mm 40mm 3mm]{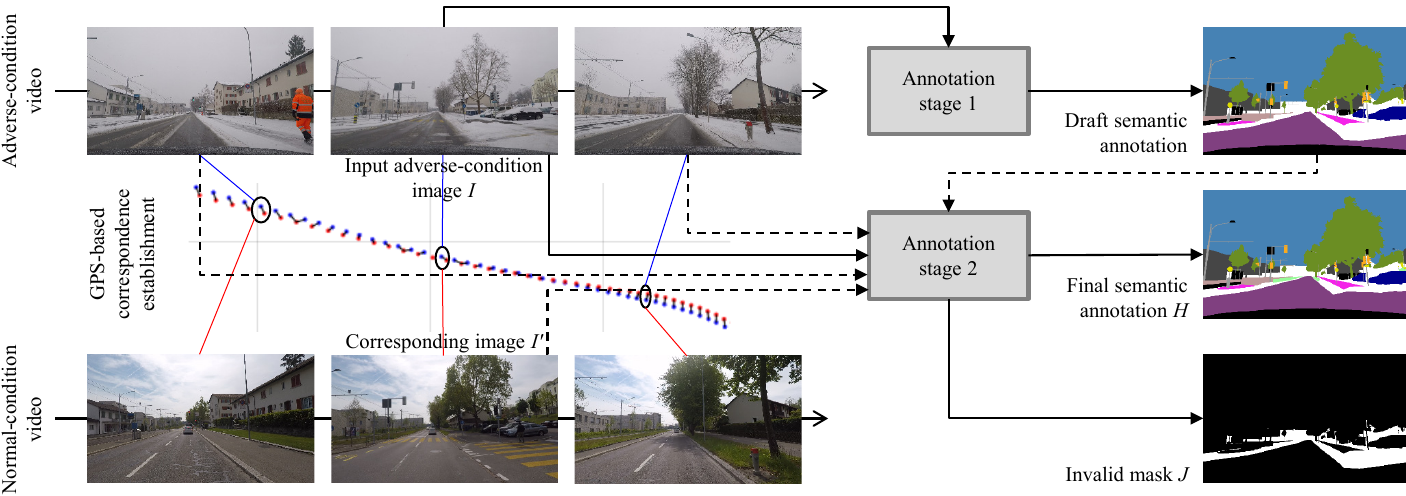}
    \vspace{-0.7cm}
    \caption{\textbf{Illustration of construction and semantic annotation pipeline for ACDC.} The color coding of the semantic classes matches Fig.~\ref{fig:dataset:stats}. All three annotation outputs pertain to the input adverse-condition image $I$. A white color in the draft semantic annotation from stage 1 and the final semantic annotation $H$ from stage 2 denotes unlabeled pixels. Dashed lines denote privileged information beyond the input adverse-condition image $I$ under annotation; this information is additionally leveraged in annotation stage 2.}
    \label{fig:annotation}
\end{figure*}

\PAR{Adaptation of semantic segmentation} networks to domains where full supervision is not available was launched shortly after the introduction of supervised approaches~\cite{FCNs:in:the:wild}. A major class of UDA works has employed adversarial domain adaptation to implicitly align the source and target domains at the level of pixels and/or features~\cite{cyCADA,adapt:structured:output:cvpr18,chen2018road,SynRealDataFogECCV18,synthetic:semantic:segmentation,FCNs:adaptation,advent:adaptation,category:adversaries:adaptation,sim:adaptation,patch:align:adaptation,decouplenet:adaptation}. Other approaches to UDA have relied on self-training with pseudo-labels in the target domain~\cite{self:training:adaptation,crst:adaptation,pixellevel:selflabeling:adaptation,simt:adaptation,diga:adaptation,IR2FRMM:adaptation} or have combined self-training with adversarial adaptation~\cite{bidirectional:learning:adaptation} or with pixel-level adaptation via explicit transforms from source to target~\cite{fda:adaptation,textda:adaptation}. However, all aforementioned approaches have been evaluated only on the artificial scenario of synthetic-to-real adaptation and overlook \emph{normal-to-adverse adaptation}, which is of higher practical importance for autonomous cars. ACDC has constituted the large-scale target-domain dataset which had previously been missing for such a normal-to-adverse experiment and steered the development of unsupervised and weakly supervised adaptation approaches~\cite{daformer:domain:adaptation,refign:domain:adaptation,hrda:domain:adaptation,sakaridis2023ciss,mic:adaptation,sepico:adaptation,CMA} that can cope with adverse target domains via the introduction of a competitive normal-to-adverse adaptation benchmark since the conference version of this paper~\cite{ACDC}. In the present extended version, we additionally introduce a sensor-level adaptation benchmark from Cityscapes to the newly annotated normal-condition split of ACDC. This benchmark does not involve a condition-level shift between the source and target domain, which both pertain to normal conditions, but it instead features a change in the camera sensor from Cityscapes to ACDC, which induces a sensor-related real-to-real low-level shift in the input images.

\PAR{Instance segmentation and panoptic segmentation} distinguish instances in the images compared to the semantic segmentation task. Instance segmentation aims to assign a segmentation mask for each object of interest in the image, while the panoptic segmentation task encompasses masks for both stuff and things classes as the combination of semantic segmentation and instance segmentation. Both of these tasks are applied in various real-world applications such as robotics, healthcare and geoscience. For instance segmentation, many works either follow the detect-then-segment pipeline~\cite{he2017maskrcnn} or explore the generation of a dynamic number of instance masks directly with clustering or dynamic kernels. The pioneering work of Mask R-CNN~\cite{he2017maskrcnn} generates instance masks based on object bounding box proposals. The following specialized instance segmentation works such as Cascaded Mask R-CNN, HTC and Detectors mainly focus on extracting better feature representations and generating more accurate object proposals. For panoptic segmentation, early works strive to combine semantic segmentation models and instance segmentation models to predict unified panoptic masks. Recent works begin to view the panoptic segmentation task in a unified perspective and formulate the stuff and things segmentation as a set prediction problem. The current state of the art for panoptic segmentation includes i.a.\ PanopticFPN and Panoptic-Deeplab with specialized instance prediction branches and semantic prediction branches, K-Net with dynamic kernels, and MaskFormer and Mask2Former with transformer architectures. Although current instance segmentation and panoptic segmentation models obtain impressive improvement on the popular Cityscapes benchmark, we show that these models generalize poorly to ACDC. The adverse images in ACDC pose a new challenge for instance and panoptic segmentation models and encourage further progress in the field.

\PAR{Adaptation of object detection} networks from a labeled source domain to another unlabeled target domain is also an active research field. Currently, there are mainly two categories of UDA for object detection methods: domain alignment and self-training. Domain alignment strives to bridge the domain gap by minimizing the domain discrepancy through style transfer, adversarial training and graph matching. Self-training relies on pseudo-labels to extract rich knowledge contained in the target domain and presents promising performance. Although all aforementioned strategies have been proven effective in alleviating the damage of domain shift, due to the lack of datasets with rich adverse conditions, these methods are mainly evaluated on synthetic-to-real, cross-city and cross-camera settings. The \emph{normal-to-adverse adaptation}, which is of high value for intelligent driving systems, is rarely examined. ACDC provides a large-scale dataset for adverse conditions and promotes the progress of unsupervised and weakly supervised object detection methods under adverse conditions.

%% file: sections/3_acdc_description.tex
We base the design of ACDC on the same general principles as seminal normal-condition datasets~\cite{Cityscapes} and adapt the collection and annotation process to fit better the adverse condition setting at hand. An overview of this process is shown in Fig.~\ref{fig:annotation}.

\subsection{Collection}
\label{sec:dataset:collection}

Our data collection is guided by the decision to record the same set of scenes both under adverse and normal conditions. We define the domain of \emph{normal} conditions as the combination of daytime and clear weather, i.e.\ good visibility and no precipitation or snow cover on the ground. While the focus of ACDC is on adverse conditions, the acquisition of the corresponding normal-condition images is vital both for the subsequent annotation step and to support weakly supervised methods, as the same scene can be much easier to parse in normal conditions, both for humans and machines.

Thus, we recorded several days of video in Switzerland by driving around in a car, primarily in urban areas but also on highways and in rural regions. In order to have a clear domain separation between different adverse conditions, we use the following criterion for the adverse-condition recordings: each recording takes place under only one type of adversity from a set of four types, i.e., fog, nighttime, rain, and snow. For example, our foggy recordings are performed at daytime and without rain or snow. For snow, both snowfall and snow cover on the ground are admissible. Moreover, we keep for further processing only the parts of the adverse-condition recordings that correspond to an intense presence of the respective condition, so as to maximize the domain shift from normal conditions as well as domain adversity.

We record with a 1080p GoPro Hero 5 camera, mounted in front of the windshield at nighttime and in normal conditions and behind the windshield in fog, rain, and snow. The camera records 8-bit RGB frames at a rate of 30 Hz.

\subsection{Correspondence Establishment}
\label{sec:dataset:correspondences}

 Our camera also provides GPS readings, which allow us to establish \emph{image-level correspondences} between adverse-condition and normal-condition recordings. In particular, for each adverse-condition recording, we perform a normal-condition recording along exactly the same route. We then use the sequences of GPS measurements of the two recordings to perform a global dynamic-programming-based matching of the adverse GPS sequence to the normal one, where the objective is defined by the Euclidean distances of matched pairs of GPS samples. Our global matching handles routes with loops better than simple nearest neighbors. Each adverse-condition frame is then matched to a normal-condition frame based on the corresponding matched samples of the GPS sequences, as shown in the left part of Fig.~\ref{fig:annotation}.
 
 \subsection{Dataset Splits}
 \label{sec:dataset:splits}
 
ACDC is split into four sets corresponding to the examined conditions. We manually selected 1000 foggy, 1006 nighttime, 1000 rainy and 1000 snowy images from the recordings for dense pixel-level semantic annotation, for a total of 4006 adverse-condition images. The selection process aimed at maximizing the complexity and diversity of captured scenes. Within each recording, any pair of selected images is at least 20 s or 50 m apart (whatever comes first).

The dataset is also split into training, validation, and test sets. We apply a global geographical split across all conditions, so that there is zero overlap between the three sets, even for different conditions. Given the abundance of training data from normal-condition datasets~\cite{Cityscapes,Mapillary,BDD100K} that allow to pre-train semantic segmentation models, we opt for a split with a greater test set size than usual. This aims at providing a highly challenging benchmark for semantic segmentation, both in terms of scale and domain adversity. In particular, we split the set of each adverse condition into 400 training, 100 validation and 500 test images, except the nighttime set with 106 validation images. This results in a total of 1600 training and 406 validation images with public annotations and 2000 test images with annotations withheld for benchmarking purposes, as per standard practice~\cite{Cityscapes}.

In the present extended version of ACDC, we newly provide annotations for a subset of the 4006 corresponding normal-condition reference images of the dataset, all of which were not annotated in the initial conference version. More specifically, for the reference normal-condition sets corresponding to each of the four adverse conditions, we annotate 50\% of the training splits (corresponding to 200 images each) and of the validation splits (corresponding to 50 images for the fog, rain, and snow reference validation splits and 53 images for the nighttime reference validation split) and 25\% of the test splits (corresponding to 125 images each). This results in a total of 800 training and 203 validation normal-condition reference images with public annotations and 500 test normal-condition reference images with annotations withheld for benchmarking purposes, as explained above. Added to the 4006 adverse-condition annotations of the initial conference version, these 1503 new normal-condition annotations raise the overall scale of annotations of the extended version of ACDC to 5509 pixel-level annotations.
 
\subsection{Semantic Annotation}
\label{sec:dataset:annotation}

\begin{table*}[!tb]
    \caption{\textbf{Comparison of ACDC against prior adverse-condition semantic segmentation datasets.} ``Adverse annot.'': total annotated adverse-condition images, ``Fog''/``Night''/``Rain''/``Snow'': annotated foggy/nighttime/rainy/snowy images, ``Inv.\ regions'': can invalid regions get legitimate labels?, ``Corr.\ normal'': are corresponding normal-condition images available?, ``Inv.\ masks'': are invalid masks available?}
    \label{table:datasets:comparison}
    \centering
    \small
    \setlength\tabcolsep{2pt}
    \begin{tabular*}{\linewidth}{l @{\extracolsep{\fill}} ccccccccccc}
    \toprule
    Dataset & Adverse annot. & Fog & Night & Rain & Snow & Classes & Reliable GT & Fine GT & Inv.\ regions & Corr.\ normal & Inv.\ masks\\
    \midrule
    Foggy Driving~\cite{SFSU_synthetic} & 101 & 101 & 0 & 0 & 0 & 19 & \yes & \yes & \no & \no & \no\\
    Foggy Zurich~\cite{CMAda:IJCV2020} & 40 & 40 & 0 & 0 & 0 & 19 & \yes & \yes & \no & \no & \no\\
    Nighttime Driving~\cite{daytime:2:nighttime} & 50 & 0 & 50 & 0 & 0 & 19 & \yes & \no & \no & \no & \no\\
    Dark Zurich~\cite{MGCDA_UIoU} & 201 & 0 & 201 & 0 & 0 & 19 & \yes & \yes & \yes & \yes & \yes\\
    Raincouver~\cite{raincouver} & 326 & 0 & 95 & 326 & 0 & 3 & \yes & \no & \no & \no & \no\\
    WildDash~\cite{wilddash} & 226 & 10 & 13 & 13 & 26 & 19 & \yes & \yes & \no & \no & \no\\
    BDD100K~\cite{BDD100K} & 1346 & 23 & 345 & 213 & 765 & 19 & \no & \yes & \no & \no & \no\\
    ACDC & \best{4006} & \best{1000} & \best{1006} & \best{1000} & \best{1000} & 19 & \yes & \yes & \yes & \yes & \yes\\
    \bottomrule
    \end{tabular*}
\end{table*}

Images captured under adverse conditions contain invalid regions, i.e. regions with indiscernible semantic content, which generally co-exist with valid regions in the same image. We take this into account for creating annotations of ACDC and design a specialized annotation protocol, which leverages privileged information from the corresponding normal-condition images and the original adverse-condition videos and allows (i) the reliable assignment of semantic labels to invalid regions and (ii) the creation of a binary mask that distinguishes valid from invalid regions.

Our annotation protocol for the 4006 adverse-condition images consists of two cascaded annotation stages. At stage~1, a semantic labeling draft is manually produced from the adverse-condition image $I$, in which pixels that cannot be unquestionably assigned to a single semantic class are left unlabeled. At stage~2, the corresponding normal-condition image $I^{\prime}$ and the adverse-condition video from which $I$ was extracted are used to augment and finalize the annotation. In particular, the annotator can assign a legitimate label to pixels that were left unlabeled in stage~1 and correct pixels that were incorrectly labeled in stage~1. Annotation during stage~2 using the corresponding normal-condition image and the neighboring frames of the adverse-condition video does not assume or require an exact match of the 6D camera pose between the annotated image and the auxiliary images or frames. In particular, the transfer of labels from the auxiliary images to the image under annotation is done through manual inspection by the annotators rather than automatically. We have explicitly instructed the annotators to be highly conservative in this manual transfer given potential mismatch of pose and dynamic image content, which is why 7.5\% of the pixels in the annotated images are left unlabeled even after stage~2. More specifically, pixels that remain unclear in stage~2 are left unlabeled and are not used for training or evaluation.

The final annotation outputs are twofold: (i) the final semantic annotation $H$ after stage~2, and (ii) a binary invalid mask $J$, where pixels whose label changed from stage~1 to stage~2 are set to 1 (invalid) and pixels with the same semantic label for both stages are set to 0 (valid). $J$ enables the introduction of the new task of uncertainty-aware semantic segmentation, which we detail in Sec.~\ref{sec:uncertain}.

The 4006 fine pixel-level annotations of ACDC were created by a professional team of annotators to ensure high-quality ground truth. Annotators were asked to be conservative in labeling pixels in both stages, so as to minimize errors. Both the initial draft from stage~1 and the final annotation from stage~2 passed through quality control (QC). QC was performed via manual inspection of each annotation by a second person (B), different from the one who created it (A). A and B were given incentives to minimize errors and maximize identified errors, respectively. A third independent person (C) verified whether each error identified by B was factual. The total average time required for annotating a single adverse-condition image was 3.3 h, with 25\% of it spent on QC. The semantic annotation of the 1503 normal-condition images is conducted in the standard way by using only the input normal-condition image.

The class specifications of ACDC are directly inherited from Cityscapes. In particular, we annotate the 19 evaluation classes of Cityscapes, which include the most common and traffic-related objects in driving scenes. Objects that belong to classes outside this set receive a fall-back label and are not used for training or evaluation. This choice of classes provides full compatibility of ACDC to Cityscapes and other normal-condition datasets for semantic segmentation~\cite{Mapillary,BDD100K}. Detailed annotation statistics are presented in Fig.~\ref{fig:dataset:stats}. An example of our two-stage annotation protocol is shown in Fig.~\ref{fig:annotation} for a snowy image. Note the assignment of a region in the lower right part of the image that is unlabeled at stage~1 (top right) to the \emph{road} label at stage~2 (middle right), thanks to the clear view from the normal-condition image.

\input{figures/instances_stat}

\begin{table}[!tb]
  \caption{Absolute and average number of instances in adverse conditions for ACDC, BDD100K, and DAWN on the respective training and validation datasets. ACDC (all) includes all training, validation and testing set.}
  \label{table:uda:inst_stats}
  \centering
  \setlength\tabcolsep{2pt}
  \small
  \begin{tabular*}{\linewidth}{l @{\extracolsep{\fill}} cccc}
  \toprule
   & \#humans$[10^3]$ & \#vehicles$[10^3]$ & \#h/image & \#v/image\\
  \midrule
  ACDC & 2.5 & 10.4 & 1.2 & 5.2\\
  ACDC (all) & 7.9 & 20.7 & 1.9 & 5.1\\
  BDD100K~\cite{BDD100K} & 0.7 & 6.6 & 1.1 & 10.2 \\
  DAWN~\cite{dawn2020dataset} & 0.4 & 7.4 & 0.4 & 7.4\\  
  \bottomrule
  \end{tabular*}
\end{table}

\subsection{Instance Annotation}
\label{sec:dataset:instance_annotation}

Besides the pixel-level semantic annotations, in the present extended version we also create dense instance-level annotations for countable objects, including vehicles and humans, to support higher-level semantic perception tasks such as instance segmentation and panoptic segmentation. To fully utilize the semantic annotations and ensure the consistency between different annotations, we develop a protocol to generate instance-level annotations based on semantic masks.

Our instance annotation protocol contains two steps. In the first step, we convert the pixel-level semantic masks to polygon representations. In the second step, we ask the annotators to merge/split polygons to form polygon annotation for each instance. The polygon annotations are finally transformed to standard COCO format for instance-level tasks. The instance-level annotations were also combined with prior semantic annotations to construct panoptic annotations for panoptic segmentation.

\begin{table*}[tb]
  \caption{\textbf{Comparison of state-of-the-art domain-adaptive semantic segmentation methods on Cityscapes$\to$ACDC adaptation.} Cityscapes serves as the source domain and the entire adverse-condition part of ACDC including all four adverse conditions serves as the target domain. The first, second, and third groups of rows present unsupervised DeepLabv2-based~\cite{DeepLab:v2}, weakly supervised, and unsupervised SegFormer-based~\cite{segformer} methods, respectively. The performance of the respective models trained on Cityscapes (Source model) and of the oracle models trained on ACDC with all 1600 labels (Oracle) is also reported in all cases, while for the DeepLabv2 case, we additionally report the performance of the partial oracle models trained on ACDC with 100 labels (Oracle-100), and 200 labels (Oracle-200).}
  \label{table:uda:seg_all:adverse}
  \centering
  \setlength\tabcolsep{2pt}
  \footnotesize
  \begin{tabular*}{\linewidth}{l @{\extracolsep{\fill}} cccccccccccccccccccc}
  \toprule
  Method & \ver{road} & \ver{sidew.} & \ver{build.} & \ver{wall} & \ver{fence} & \ver{pole} & \ver{light} & \ver{sign} & \ver{veget.} & \ver{terrain} & \ver{sky} & \ver{person} & \ver{rider} & \ver{car} & \ver{truck} & \ver{bus} & \ver{train} & \ver{motorc.} & \ver{bicycle} & mIoU\\
  \midrule
  Source model~\cite{DeepLab:v2} & 71.9 & 26.2 & 51.1 & 18.8 & 22.5 & 19.7 & 33.0 & 27.7 & 67.9 & 28.6 & 44.2 & 43.1 & 22.1 & 71.2 & 29.8 & 33.3 & 48.4 & 26.2 & 35.8 &     38.0\\
  AdaptSegNet~\cite{adapt:structured:output:cvpr18} & 69.4 & 34.0 & 52.8 & 13.5 & 18.0 & 4.3 & 14.9 & 9.7 & 64.0 & 23.1 & 38.2 & 38.6 & 20.1 & 59.3 & 35.6 & 30.6 & 53.9 & 19.8 & 33.9 &    33.4\\
  ADVENT~\cite{advent:adaptation} & 72.9 & 14.3 & 40.5 & 16.6 & 21.2 & 9.3 & 17.4 & 21.2 & 63.8 & 23.8 & 18.3 & 32.6 & 19.5 & 69.5 & 36.2 & 34.5 & 46.2 & 26.9 & 36.1 &     32.7\\
  BDL~\cite{bidirectional:learning:adaptation} & 56.0 & 32.5 & 68.1 & 20.1 & 17.4 & 15.8 & 30.2 & 28.7 & 59.9 & 25.3 & 37.7 & 28.7 & 25.5 & 70.2 & 39.6 & 40.5 & 52.7 & 29.2 & 38.4 &      37.7\\
  CLAN~\cite{category:adversaries:adaptation} & 79.1 & 29.5 & 45.9 & 18.1 & 21.3 & 22.1 & 35.3 & 40.7 & 67.4 & 29.4 & 32.8 & 42.7 & 18.5 & 73.6 & 42.0 & 31.6 & 55.7 & 25.4 & 30.7 &      39.0\\
  CRST~\cite{crst:adaptation} & 51.7 & 24.4 & 67.8 & 13.3 & 9.7 & 30.2 & 38.2 & 34.1 & 58.0 & 25.2 & 76.8 & 39.9 & 17.1 & 65.4 & 3.7 & 6.6 & 39.6 & 11.8 & 8.6 &     32.8\\
  FDA~\cite{fda:adaptation} & 73.2 & 34.7 & 59.0 & 24.8 & 29.5 & 28.6 & 43.3 & 44.9 & 70.1 & 28.2 & 54.7 & 47.0 & 28.5 & 74.6 & 44.8 & 52.3 & 63.3 & 28.3 & 39.5 &    45.7\\
  SIM~\cite{sim:adaptation} & 53.8 & 6.8 & 75.5 & 11.6 & 22.3 & 11.7 & 23.4 & 25.7 & 66.1 & 8.3 & 80.6 & 41.8 & 24.8 & 49.7 & 38.6 & 21.0 & 41.8 & 25.1 & 29.6 &    34.6\\
  MRNet~\cite{mrnet:rectifying} & 72.2 & 8.2 & 36.4 & 13.7 & 18.5 & 20.4 & 38.7 & 45.4 & 70.2 & 35.7 & 5.0 & 47.8 & 19.1 & 73.6 & 42.1 & 36.0 & 47.4 & 17.7 & 37.4 &    36.1\\
  DACS~\cite{dacs:domain:adaptation} & 58.5 & 34.7 & 76.4 & 20.9 & 22.6 & 31.7 & 32.7 & 46.8 & 58.7 & 39.0 & 36.3 & 43.7 & 20.5 & 72.3 & 39.6 & 34.8 & 51.1 & 24.6 & 38.2 & 41.2 \\
  CISS~\cite{sakaridis2023ciss} & 70.5 & 36.7 & 67.0 & 29.4 & 30.2 & 31.6 & 45.6 & 48.9 & 70.4 & 24.7 & 65.5 & 48.2 & 31.1 & 76.6 & 45.7 & 47.0 & 62.8 & 26.8 & 38.9 & 47.2 \\
  Oracle-100 & 84.4 & 54.8 & 76.4 & 19.3 & 28.9 & 29.5 & 36.5 & 42.6 & 74.2 & 40.3 & 87.7 & 42.5 & 16.5 & 74.9 & 36.5 & 28.6 & 55.9 & 27.3 & 38.6 & 47.1\\
  Oracle-200 & 86.2 & 55.0 & 77.9 & 21.7 & 30.9 & 30.0 & 37.6 & 42.5 & 76.8 & 45.8 & 90.2 & 45.4 & 19.1 & 75.8 & 38.5 & 38.0 & 64.2 & 21.6 & 39.5 & 49.3\\
  Oracle & 88.0 & 62.3 & 80.8 & 37.0 & 35.1 & 33.9 & 49.8 & 49.5 & 80.1 & 50.7 & 92.5 & 51.1 & 26.5 & 79.9 & 49.0 & 41.1 & 72.2 & 26.5 & 44.2 &    55.3\\
  \midrule
  Source model~\cite{refinenet} & 66.3 & 28.9 & 67.6 & 19.2 & 25.9 & 36.7 & 50.0 & 47.5 & 69.4 & 28.8 & 83.0 & 42.1 & 17.7 & 72.6 & 30.9 & 31.6 & 48.9 & 26.1 & 36.7 &     43.7\\
  MGCDA~\cite{MGCDA_UIoU} & 73.4 & 28.7 & 69.9 & 19.3 & 26.3 & 36.8 & 53.0 & 53.3 & 75.4 & 32.0 & 84.6 & 51.0 & 26.1 & 77.6 & 43.2 & 45.9 & 53.9 & 32.7 & 41.5 & 48.7\\
  Oracle & 92.5 & 71.2 & 86.2 & 39.0 & 44.0 & 53.2 & 68.8 & 66.0 & 85.1 & 59.3 & 94.9 & 65.2 & 38.5 & 85.8 & 53.8 & 59.7 & 76.2 & 47.5 & 54.5 &         65.3\\
  \midrule
  Source model~\cite{daformer:domain:adaptation}  & 80.5 & 37.4 & 80.5 & 34.7 & 30.4 & 43.7 & 57.9 & 54.2 & 79.0 & 51.6 & 87.6 & 57.4 & 34.0 & 81.5 & 51.9 & 59.1 & 70.4 & 37.5 & 49.3 & 56.8  \\
  DAFormer~\cite{daformer:domain:adaptation} & 58.4 & 51.3 & 84.0 & 42.7 & 35.1 & 50.7 & 30.0 & 57.0 & 74.8 & 52.8 & 51.3 & 58.2 & 32.6 & 82.7 & 58.3 & 54.9 & 82.4 & 44.1 & 50.7 & 55.4\\
  SePiCo~\cite{sepico:adaptation} & 61.3 & 48.6 & 84.9 & 39.6 & 40.3 & 54.2 & 48.9 & 60.6 & 74.8 & 54.3 & 57.2 & 65.2 & 38.3 & 84.8 & 66.2 & 60.4 & 85.5 & 44.5 & 53.1 & 59.1\\
  HRDA~\cite{hrda:domain:adaptation} & 88.3 & 57.9 & 88.1 & 55.2 & 36.7 & 56.3 & 62.9 & 65.3 & 74.2 & 57.7 & 85.9 & 68.8 & 45.6 & 88.5 & 76.4 & 82.4 & 87.7 & 52.7 & 60.4 & 68.0 \\
  CISS~\cite{sakaridis2023ciss} & 92.0 & 69.6 & 89.2 & 57.2 & 40.5 & 55.8 & 67.1 & 67.3 & 75.2 & 59.7 & 86.4 & 70.0 & 47.5 & 88.9 & 73.1 & 77.5 & 87.0 & 55.6 & 61.7 & 69.6 \\
  MIC~\cite{mic:adaptation} & 90.8 & 67.1 & 89.2 & 54.5 & 40.5 & 57.2 & 62.0 & 68.4 & 76.3 & 61.8 & 87.0 & 71.3 & 49.4 & 89.7 & 75.7 & 86.8 & 89.1 & 56.9 & 63.0 & 70.4 \\
  Oracle  & 93.2 & 74.2 & 89.5 & 54.5 & 47.4 & 57.0 & 68.9 & 66.9 & 88.5 & 66.0 & 96.2 & 64.2 & 30.6 & 85.8 & 59.6 & 64.7 & 86.3 & 39.8 & 54.3 & 67.8 \\
  \bottomrule
  \end{tabular*}
\end{table*}

\begin{table}[tb]
  \caption{\textbf{Comparison of state-of-the-art unsupervised domain adaptation methods on Cityscapes$\to$ACDC adaptation for individual conditions.} We train a separate model on each condition-specific subset of ACDC and evaluate each model on the condition it has been trained for. Performance of the model trained only on the source domain (Source model) and of oracles with access to the target-domain labels for each condition (Oracle) is also reported.}
  \label{table:uda:seg_individual}
  \centering
  \setlength\tabcolsep{2pt}
  \small
  \begin{tabular*}{\linewidth}{l @{\extracolsep{\fill}} cccc}
  \toprule
  Method & Fog & Night & Rain & Snow\\
  \midrule
  Source model & 33.5 & 30.1 & 44.5 & 40.2\\
  \midrule
  AdaptSegNet~\cite{adapt:structured:output:cvpr18} & 31.8 & 29.7 & 49.0 & 35.3\\
  ADVENT~\cite{advent:adaptation} & 32.9 & 31.7 & 44.3 & 32.1\\
  BDL~\cite{bidirectional:learning:adaptation} & 37.7 & 33.8 & 49.7 & 36.4\\
  CLAN~\cite{category:adversaries:adaptation} & 39.0 & 31.6 & 44.0 & 37.7\\
  FDA~\cite{fda:adaptation} & 39.5 & 37.1 & 53.3 & 46.9\\
  SIM~\cite{sim:adaptation} & 36.6 & 28.0 & 44.5 & 33.3\\
  MRNet~\cite{mrnet:rectifying} & 38.8 & 27.9 & 45.4 & 38.7\\
  \midrule
  Oracle & 52.2 & 45.4 & 57.6 & 56.8\\
  \bottomrule
  \end{tabular*}
\end{table}

The total 4006 adverse-condition images and 1503 normal-condition reference images are annotated at the instance level by a professional team of annotators. 
Each final instance-level annotation underwent a QC process. 
Specifically, QC was carried out through manual inspection by two individuals: the original annotator and an independent reviewer who had not previously seen the condition, in order to minimize bias.
The total time required for annotating a single image given the initial semantic mask was 0.6 h on average.

To be compatible with semantic annotations, for instance annotations, we only create instance-level masks for traffic participant classes such as humans and vehicles. We present the detailed distribution of instances in ACDC in Fig.~\ref{fig:dataset:inst_stats}. The dataset follows a long-tail distribution across countable classes in terms of instance counts. The \emph{vehicle} category is dominated by the \emph{car} class, while the \emph{person} class emerges as the predominant class for the \emph{human} category. This skewed distribution reflects the real world, where cars often outnumber other vehicle types heavily, and pedestrians are more commonly encountered compared to riders.

\subsection{Comparison to Related Datasets}
\label{sec:dataset:comparison}

To the best of our knowledge, ACDC constitutes the largest adverse-condition driving dataset for dense semantic perception to date. In Table~\ref{table:datasets:comparison}, we compare ACDC to prior datasets that also address semantic segmentation under adverse conditions. Most of these datasets focus on a single condition and are of small scale. WildDash covers a wider variety of adverse conditions but also has a small scale. BDD100K includes 10000 images with semantic segmentation annotations. We inspected these images manually to identify those that pertain to fog, night, rain, and snow. We found that only 1346/10000 images pertain to any of these four conditions. By contrast, ACDC is primarily composed of these four common adverse conditions. Notably, it contains one order of magnitude more annotated images than any other competing dataset for each of fog, night and rain. At the same time, our specialized annotation protocol using corresponding normal-condition images ensures \emph{reliable} annotations even for invalid regions contrary to BDD100K, making ACDC a high-quality dataset for training and evaluation for adverse conditions. 
We also provide a comparison on the number of annotated instances  with existing datasets covering adverse conditions in Table~\ref{table:uda:inst_stats}. Only a small portion of BDD100K provides instance-level annotations in adverse conditions including fog, night, rain and snow. The DAWN dataset consists of 1000 images in adverse weather conditions from real traffic environments. ACDC presents clear advantages in the total number of annotated humans and vehicles in challenging adverse conditions compared to both DAWN and BDD100K, thereby offering a broader spectrum of diverse scenes under challenging adverse conditions.

\begin{table*}[tb]
  \caption{\textbf{Comparison of state-of-the-art domain-adaptive object detection methods on Cityscapes$\to$ACDC adaptation.} Cityscapes serves as the source domain and the entire adverse-condition part of ACDC including all four adverse conditions serves as the target domain. The first and second groups of rows present unsupervised and weakly supervised methods, respectively. All unsupervised methods share the same network architecture. The performance of the respective models trained on Cityscapes (Source model) and of the oracle models trained on ACDC with all 1600 labels (Oracle) is also reported.}
  \label{table:uda:det_all}
  \centering
  \setlength\tabcolsep{2pt}
  \footnotesize
  \begin{tabular*}{\linewidth}{l @{\extracolsep{\fill}} cccccccccc}
  \toprule
  Method & \ver{person} & \ver{rider} & \ver{car} & \ver{truck} & \ver{bus} & \ver{train} & \ver{motorc.} & \ver{bicycle} & AP$_{50}^{box}$& AP$^{box}$\\
  \midrule
  Source model (Faster R-CNN)~\cite{faster:rcnn} & 22.8 & 12.2 & 51.9 & 20.0 & 19.6 & 16.0 & 13.4 & 10.4 & 20.8 & 10.3    \\
  DA-Faster~\cite{DomainAdaptiveFasterRCNN} & 28.0 & 13.6 & 57.0 & 13.1 & 13.3 & 10.6 & 8.2 & 14.2 & 19.8 & 9.2    \\
  SADA~\cite{chen2021scale} & 34.2 & 12.2 & 61.8 & 11.0 & 5.4 & 7.3 & 9.6 & 15.8 & 19.7 & 9.4     \\
  MIC (SADA)~\cite{mic:adaptation} & 40.0 & 23.0 & 67.2 & 13.5 & 8.2 & 12.3 & 20.5 & 22.9 & 25.9 & 12.1     \\
  FRCNN-SIGMA++~\cite{li2023sigma_plus_plus} & 26.4 & 19.5 & 13.6 & 16.5 & 16.6 & 57.8 & 22.7 & 20.5 & 24.2 & 11.9 \\ 
  Oracle & 28.7 & 17.3 & 61.8 & 29.8& 14.8 & 36.1 & 19.8 & 13.1 & 27.7 & 13.1    \\
  \midrule
  Source model (FCOS)~\cite{tian2021fcos} & 28.4 & 10.9 & 53.8 & 18.9 & 17.4 & 13.5 & 13.2 & 10.8 & 20.9 & 10.7    \\
  EPM~\cite{hsu2020epm}   & 30.8 & 11.5 & 56.0 & 16.7 & 19.6 & 15.6 & 16.3 & 9.9 & 22.0 & 11.2    \\
  SIGMA~\cite{li2022sigma} & 31.5 & 9.8 & 59.7 & 17.5 & 10.1 & 14.1 & 19.3 & 17.0 & 22.4 & 9.5    \\
  Oracle & 40.5 & 21.7 & 67.5 & 29.5 & 15.7 & 37.5 & 18.5 & 14.1 & 30.6 & 15.7    \\
  \bottomrule
  \end{tabular*}
\end{table*}

\begin{table}[tb]
  \caption{\textbf{Comparison of state-of-the-art unsupervised domain-adaptive object detection methods on Cityscapes$\to$ACDC adaptation for individual conditions.} We train a separate model on each condition-specific subset of ACDC and evaluate each model on the condition it has been trained for. Performance of the model trained only on the source domain (Source model) and of oracles with access to the target domain labels for each condition (Oracle) is also reported in $AP^{box}_{50}$.}
  \label{table:uda:det_individual}
  \centering
  \setlength\tabcolsep{2pt}
  \small
  \begin{tabular*}{\linewidth}{l @{\extracolsep{\fill}} cccc}
  \toprule
  Method & Fog & Night & Rain & Snow\\
  \midrule
  Source model~\cite{faster:rcnn}  & 19.7 & 14.4 & 23.9 & 29.2 \\
  DA-Faster~\cite{DomainAdaptiveFasterRCNN}     & 17.3 & 11.6  & 21.7 & 29.9 \\
  SADA~\cite{chen2021scale}          & 19.5 & 17.9 & 24.0 & 28.2 \\
  MIC (SADA)~\cite{mic:adaptation}     & 24.8 & 18.4 & 26.1 & 31.5 \\
  FRCNN-SIGMA++~\cite{li2023sigma_plus_plus} & 23.2 & 23.2 & 27.4 & 33.8\\
  Oracle        & 28.9 & 27.9 & 35.9 & 41.9 \\
  \midrule
  Source model~\cite{tian2021fcos} & 22.0 & 14.4 & 22.6 & 28.4 \\
  EPM~\cite{hsu2020epm}          & 22.3 & 15.7 & 21.9 & 25.8 \\
  SIGMA~\cite{li2022sigma}        & 25.4 & 18.5 & 24.4 & 19.9 \\
  Oracle       & 28.6 & 28.7 & 36.2 & 39.2 \\
  \bottomrule
  \end{tabular*}
\end{table}

%% file: figures/instances_stat.tex
\begin{figure}[!tb]
    \centering
    \begin{tikzpicture}
    \tikzstyle{every node}=[font=\footnotesize]
    \begin{axis}[
      ybar,
      ymode=log,
      width=\columnwidth,
      height=4.5cm,
      xmin=0,
      xmax=10,
      ymin=80,
      ymax=80000,
      ymajorgrids=true,
      ylabel={number of instances},
      ytick={200,2000,20000},
      yticklabels={$200$,$2000$,$20000$,},
      xtick={3,8.5},
      minor xtick={7},
      xticklabels = {
        vehicle,
        human,
      },
      major x tick style = {opacity=0},
      minor x tick num = 1,
      xtick pos=left,
      every node near coord/.append style={
      anchor=west,
      rotate=90,
      font=\scriptsize,
      }
    ]

    \addplot[bar shift=0pt,draw=car,           fill opacity=0.8,fill=car!80!white            , nodes near coords=car           ] plot coordinates{ ( 1,    16523   ) };
    \addplot[bar shift=0pt,draw=train,         fill opacity=0.8,fill=train!80!white          , nodes near coords=train         ] plot coordinates{ ( 2,    711        ) };
    \addplot[bar shift=0pt,draw=truck,         fill opacity=0.8,fill=truck!80!white          , nodes near coords=truck         ] plot coordinates{ ( 3,    1062     ) };
    \addplot[bar shift=0pt,draw=bus,           fill opacity=0.8,fill=bus!80!white            , nodes near coords=bus           ] plot coordinates{ ( 4,    401     ) };
    \addplot[bar shift=0pt,draw=bicycle,       fill opacity=0.8,fill=bicycle!80!white        , nodes near coords=bicycle       ] plot coordinates{ ( 5,    1319      ) };
    \addplot[bar shift=0pt,draw=motorcycle,    fill opacity=0.8,fill=motorcycle!80!white     , nodes near coords=motorcycle    ] plot coordinates{ ( 6,    669        ) };

    \addplot[bar shift=0pt,draw=person,        fill opacity=0.8,fill=person!80!white         , nodes near coords=person        ] plot coordinates{ ( 8,    7304     ) };
    \addplot[bar shift=0pt,draw=rider,         fill opacity=0.8,fill=rider!80!white          , nodes near coords=rider         ] plot coordinates{ ( 9,    598        ) };

    \end{axis}
    \end{tikzpicture}
    \caption{Number of instances per class in ACDC.}
    \label{fig:dataset:inst_stats}
\end{figure}

%% file: sections/4_adapt2adverse.tex
ACDC supports various semantic perception tasks, including semantic segmentation, object detection, instance segmentation, and panoptic segmentation. In this section, we experiment on our dataset with domain adaptation methods for semantic segmentation and object detection.

\subsection{Domain-Adaptive Semantic Segmentation}

We present a new benchmark for UDA of semantic segmentation: Cityscapes$\to$ACDC. We select fourteen representative state-of-the-art UDA methods, train them with their default configurations for adaptation from Cityscapes to the entire ACDC and present the results in Table~\ref{table:uda:seg_all:adverse}. Ten of these methods are trained with the earlier DeepLabv2-based architecture~\cite{DeepLab:v2}, while five of these methods are trained with the more modern SegFormer backbone~\cite{segformer}. While most of the DeepLabv2-based methods have previously achieved significant performance gains in the popular synthetic-to-real adaptation setting, we observe that most of them do not improve upon the source-domain baseline in our normal-to-adverse setting. The best-performing DeepLabv2-based methods are CISS and FDA, which are respectively based on non-adversarial, feature-level and pixel-level adaptation strategies with an explicit Fourier prior. Only CISS slightly outperforms the model that is supervised with only 100 target-domain labels. With regard to the more recent, SegFormer-based methods, most of them manage to deliver substantial performance improvements on the target, adverse-condition domain of ACDC compared to the already strong source model. In fact, MIC, CISS, and HRDA even prove capable of \emph{surpassing} the performance of the \emph{oracle} model, which has exactly the same architecture as the source model but also access to target-domain labels during training. This is an encouraging finding for the domain adaptation community, as it corroborates the benefit of introducing informed inductive biases to learned models via proper losses or architectural modules in order to improve their generalization to unlabeled data over merely feeding the models with more labeled data.

The image-level correspondences of ACDC between adverse and normal conditions act as weak supervision. We thus additionally experiment with MGCDA, a weakly supervised method that exploits such correspondences. MGCDA outperforms FDA but is still inferior to its fully supervised counterpart.

In addition, we train state-of-the-art UDA methods to adapt from Cityscapes to individual conditions of ACDC in Table~\ref{table:uda:seg_individual}. The increased uniformity of the target domains in this setting results in larger performance gains overall compared to Table~\ref{table:uda:seg_all:adverse}. However, night and snow prove particularly challenging for most methods and only FDA brings a performance gain on snow.

\begin{table*}[tb]
  \caption{\textbf{Comparison of state-of-the-art domain-adaptive semantic segmentation methods on Cityscapes$\to$ACDC-Reference adaptation.} Cityscapes serves as the source domain and ACDC-Reference serves as the target domain. The first and second groups of rows present DeepLabv2-based~\cite{DeepLab:v2} and SegFormer-based~\cite{segformer} unsupervised methods, respectively. The performance of the respective models trained on Cityscapes (Source model) and of the oracle models trained on ACDC-Reference with all its 800 training labels (Oracle) is reported.}
  \label{table:uda:seg_all:normal}
  \centering
  \setlength\tabcolsep{2pt}
  \footnotesize
  \begin{tabular*}{\linewidth}{l @{\extracolsep{\fill}} cccccccccccccccccccc}
  \toprule
  Method & \ver{road} & \ver{sidew.} & \ver{build.} & \ver{wall} & \ver{fence} & \ver{pole} & \ver{light} & \ver{sign} & \ver{veget.} & \ver{terrain} & \ver{sky} & \ver{person} & \ver{rider} & \ver{car} & \ver{truck} & \ver{bus} & \ver{train} & \ver{motorc.} & \ver{bicycle} & mIoU\\
  \midrule
  Source model~\cite{DeepLab:v2} & 84.6 & 48.2 & 75.4 & 26.1 & 30.5 & 32.7 & 30.1 & 40.0 & 82.8 & 56.7 & 84.2 & 47.3 & 45.8 & 79.2 & 28.1 & 45.4 & 59.0 & 32.8 & 52.7 & 51.7    \\
  AdaptSegNet~\cite{adapt:structured:output:cvpr18} & 88.8 & 58.8 & 80.6 & 23.9 & 33.0 & 22.9 & 48.4 & 39.7 & 85.8 & 62.8 & 94.2 & 50.6 & 46.2 & 61.1 & 40.5 & 44.4 & 53.8 & 40.5 & 58.1 & 54.4    \\
  ADVENT~\cite{advent:adaptation} & 84.9 & 54.3 & 83.1 & 26.3 & 28.5 & 23.7 & 37.5 & 35.5 & 85.6 & 62.7 & 95.8 & 53.5 & 50.2 & 50.0 & 48.0 & 51.4 & 63.0 & 43.9 & 59.2 & 54.6    \\
  BDL~\cite{bidirectional:learning:adaptation} & 89.4 & 61.5 & 80.1 & 25.4 & 28.4 & 22.0 & 46.8 & 40.8 & 85.8 & 63.3 & 94.3 & 48.2 & 48.7 & 76.8 & 43.6 & 44.3 & 59.6 & 46.8 & 58.9 & 56.1    \\
  CLAN~\cite{category:adversaries:adaptation} & 87.9 & 49.6 & 78.5 & 27.5 & 29.2 & 35.9 & 53.7 & 49.5 & 86.3 & 66.6 & 90.2 & 56.0 & 41.9 & 80.5 & 33.4 & 46.9 & 61.2 & 47.7 & 55.3 & 56.7    \\
  FDA~\cite{fda:adaptation} & 91.8 & 66.8 & 83.3 & 33.5 & 33.8 & 37.6 & 59.7 & 55.0 & 86.4 & 61.1 & 93.5 & 57.3 & 51.9 & 81.5 & 37.4 & 65.6 & 61.3 & 45.4 & 57.5 & 61.1    \\
  SIM~\cite{sim:adaptation} & 86.7 & 50.6 & 81.4 & 13.0 & 28.4 & 23.9 & 48.0 & 35.7 & 85.5 & 64.5 & 91.3 & 51.0 & 50.9 & 72.6 & 43.0 & 42.9 & 53.1 & 32.8 & 44.5 & 52.6    \\
  MRNet~\cite{mrnet:rectifying} & 90.1 & 59.4 & 83.5 & 31.3 & 30.1 & 40.6 & 59.3 & 53.7 & 88.4 & 67.0 & 95.6 & 58.1 & 55.1 & 84.4 & 53.5 & 44.6 & 64.8 & 58.2 & 64.2 & 62.2    \\
  CISS~\cite{sakaridis2023ciss} & 92.8 & 69.2 & 84.6 & 34.3 & 34.4 & 42.4 & 59.5 & 57.4 & 87.1 & 61.8 & 94.7 & 59.9 & 54.4 & 82.7 & 48.0 & 66.0 & 65.7 & 51.5 & 61.0 & 63.5    \\
  Oracle  & 94.1 & 74.7 & 86.5 & 44.3 & 36.2 & 39.4 & 58.3 & 53.2 & 87.4 & 68.8 & 96.1 & 56.3 & 44.7 & 83.1 & 42.8 & 52.5 & 67.5 & 45.8 & 56.7 & 62.5    \\
  \midrule
  Source model~\cite{daformer:domain:adaptation}  & 89.1 & 58.1 & 89.6 & 44.1 & 38.1 & 54.3 & 68.6 & 63.4 & 91.2 & 72.6 & 97.6 & 66.0 & 56.5 & 87.8 & 50.1 & 74.7 & 77.9 & 55.6 & 64.0 & 68.4    \\
  DAFormer~\cite{daformer:domain:adaptation}  & 92.4 & 69.9 & 90.6 & 63.4 & 36.5 & 55.2 & 70.2 & 55.9 & 91.1 & 70.1 & 97.6 & 66.5 & 57.7 & 83.5 & 53.5 & 74.3 & 79.7 & 58.8 & 62.1 & 69.9    \\
  HRDA~\cite{hrda:domain:adaptation}  & 88.8 & 53.6  & 92.0  &  66.4  &  36.2  &  58.2 & 76.6 & 60.7 & 91.0 & 73.0 & 97.3 & 76.3 & 69.1 & 91.8 & 70.2 & 95.0 & 89.3 & 68.3 & 75.3 & 75.2 \\
  MIC~\cite{mic:adaptation} & 90.9 & 62.8 & 92.1 & 65.5 & 41.9 & 61.9 & 76.7 & 71.1 & 88.7 & 75.0 & 94.2 & 76.3 & 69.5 & 92.6 & 72.4 & 94.7 & 90.1 & 70.5 & 75.6 & 77.0    \\
  CISS~\cite{sakaridis2023ciss} & 95.8 & 80.3 & 92.6 & 69.0 & 38.4 & 60.8 & 76.9 & 68.5 & 92.0 & 74.3 & 98.0 & 76.9 & 70.6 & 92.9 & 71.6 & 91.9 & 88.5 & 69.7 & 75.5 & 78.1 \\
  Oracle  & 96.0 & 80.9 & 91.3 & 63.1 & 45.8 & 59.0 & 72.3 & 66.2 & 91.6 & 75.2 & 98.0 & 67.8 & 58.3 & 86.5 & 56.3 & 65.7 & 80.6 & 53.6 & 67.6 & 72.4    \\
  \bottomrule
  \end{tabular*}
\end{table*}

\begin{table*}[tb]
  \caption{\textbf{Comparison of state-of-the-art domain-adaptive object detection methods on Cityscapes$\to$ACDC-Reference adaptation.} Cityscapes serves as the source domain and ACDC-Reference serves as the target domain. The first and second groups of rows present two-stage domain-adaptive detection and one-stage domain-adaptive detection methods, respectively. All methods share the same ResNet-50 backbone. The performance of the respective models trained on Cityscapes (Source model) and of the oracle models trained on ACDC-Reference with all its 800 training labels (Oracle) is also reported.}
  \label{table:uda:det_ref_all}
  \centering
  \setlength\tabcolsep{2pt}
  \footnotesize
  \begin{tabular*}{\linewidth}{l @{\extracolsep{\fill}} cccccccccc}
  \toprule
  Method & \ver{person} & \ver{rider} & \ver{car} & \ver{truck} & \ver{bus} & \ver{train} & \ver{motorc.} & \ver{bicycle} & AP$_{50}^{box}$& AP$^{box}$\\
  \midrule
  Source model (Faster R-CNN)~\cite{faster:rcnn} & 22.1 & 32.2 & 45.4 & 16.4 & 19.4 & 20.8 & 26.7 & 24.3 & 25.9 & 12.6 \\
  DA-Faster~\cite{DomainAdaptiveFasterRCNN} & 21.9 & 34.8 & 46.7 & 13.6 & 17.5 & 18.7 & 26.3 & 27.6 & 25.9 & 12.1 \\
  SADA~\cite{chen2021scale} & 38.0 & 40.9 & 56.3 & 3.5 & 6.7 & 1.7 & 25.8 & 29.6 & 25.3 & 12.0 \\
  MIC (SADA)~\cite{mic:adaptation} & 35.1 & 37.9 & 56.1 & 8.9 & 10.5 & 10.5 & 29.3 & 31.4 & 27.5 & 12.6\\
  FRCNN-SIGMA++~\cite{li2023sigma_plus_plus} & 21.9 & 31.2 & 44.8 & 18.2 & 15.8 & 21.8 & 27.6 & 26.9 & 26.0 & 12.6 \\
  Oracle & 24.3 & 34.6 & 49.0 & 31.6 & 20.5 & 27.9 & 34.5 & 25.0 & 30.9 & 15.4 \\
  \midrule
  Source model (FCOS)~\cite{tian2021fcos} & 30.6 & 28.3 & 50.6 & 19.8 & 21.5 & 12.6 & 25.4 & 21.8 & 26.3 & 13.3 \\
  EPM~\cite{hsu2020epm}   & 32.3 & 28.7 & 52.2 & 16.8 & 19.7 & 12.4 & 29.2 & 19.9 & 26.4 & 13.4 \\
  SIGMA~\cite{li2022sigma} & 31.5 & 31.2 & 53.6 & 18.7 & 17.3 & 16.9 & 28.6 & 26.8 & 28.1 & 14.0 \\
  Oracle & 32.7 & 56.8 & 25.5 & 32.6 & 29.2 & 32.6 & 23.3 & 24.6 & 32.2 & 15.9\\
  \bottomrule
  \end{tabular*}
\end{table*}

\subsection{Domain-Adaptive Object Detection}

We establish a new benchmark for UDA of object detection: Cityscapes$\to$ACDC. We select seven representative UDA methods for detection, and perform adaptation from Cityscapes to the entire adverse-condition part of the ACDC training set including all four adverse conditions, with the default configuration designed for Cityscapes to Foggy Cityscapes adaptation. The Cityscapes$\to$ACDC adaptation results are reported in Table~\ref{table:uda:det_all}. As different UDA methods are built on either one-stage or two-stage detection frameworks, we report the results in two groups: two-stage UDA detection methods share the same Faster R-CNN detection architecture and one-stage UDA detection methods share the same FCOS detection framework. For two-stage detection methods, we report the performance of the adversarial-training-based UDA methods DA-Faster, SADA and MIC (SADA) and the graph-matching-based method FRCNN-SIGMA++. For one-stage detection methods, we present the results of the adversarial-learning-based method EPM and the graph-matching-based method SIGMA. Following the previous works in cross-domain object detection, we report $AP^{box}_{50}$ for each category by default. We also provide overall COCO $AP^{box}$ for reference. As most of UDA object detection works benchmark their method on Cityscapes to Foggy Cityscapes for normal-to-adverse adaptation, for comparison we adopt the same configurations to perform adaptation from Cityscapes to ACDC. We expect to present the difference between real and synthetic adverse data and the importance of realistic adverse-condition images in ACDC. For a fair comparison, we utilize the validation set to pick the best model and report its performance on the test set as described in ~\cite{li2022sigma}.

From Table~\ref{table:uda:det_all} we observe that the configuration designed for Cityscapes$\to$Foggy Cityscapes may not be applicable on Cityscapes$\to$ACDC and ACDC demonstrates a challenging benchmark for normal-to-adverse adaptation. Several adversarial-based UDA methods bring losses in performance compared to the source-only model. Other methods presenting limited improvement still have an obvious gap compared to the oracle model. MIC (SADA) exhibits the largest improvement among two-stage object detectors and SIGMA obtains the best performance among one-stage object detectors. Although these models obtain improvement for Cityscapes$\to$Foggy Cityscapes task, the performance drop on Cityscapes$\to$ACDC indicates that the synthetic Foggy Cityscapes is still different from real-world adverse conditions and that ACDC poses a new challenge to existing UDA methods and enables a more realistic setting for domain-adaptive detection.

\begin{table*}[tb]
  \caption{\textbf{Comparison of state-of-the-art supervised semantic segmentation methods on ACDC.} Training and evaluation are performed using the training and test sets of the entire adverse-condition part of ACDC including all four adverse conditions, respectively.}
  \label{table:supervised:all}
  \centering
  \setlength\tabcolsep{2pt}
  \footnotesize
  \begin{tabular*}{\linewidth}{l @{\extracolsep{\fill}} cccccccccccccccccccc}
  \toprule
  Method & \ver{road} & \ver{sidew.} & \ver{build.} & \ver{wall} & \ver{fence} & \ver{pole} & \ver{light} & \ver{sign} & \ver{veget.} & \ver{terrain} & \ver{sky} & \ver{person} & \ver{rider} & \ver{car} & \ver{truck} & \ver{bus} & \ver{train} & \ver{motorc.} & \ver{bicycle} & mIoU\\
  \midrule
  RefineNet~\cite{refinenet} & 92.5 & 71.2 & 86.2 & 39.0 & 44.0 & 53.2 & 68.8 & 66.0 & 85.1 & 59.3 & 94.9 & 65.2 & 38.5 & 85.8 & 53.8 & 59.7 & 76.2 & 47.5 & 54.5 &         65.3\\
  DeepLabv2~\cite{DeepLab:v2} & 88.0 & 62.3 & 80.8 & 37.0 & 35.1 & 33.9 & 49.8 & 49.5 & 80.1 & 50.7 & 92.5 & 51.1 & 26.5 & 79.9 & 49.0 & 41.1 & 72.2 & 26.5 & 44.2 &    55.3\\
  DeepLabv3+~\cite{DeepLab:v3+} & 93.4 & 74.8 & 89.2 & 53.0 & 49.0 & 58.7 & 71.1 & 67.4 & 87.8 & 62.7 & 95.9 & 69.7 & 36.0 & 88.1 & 67.7 & 71.8 & 85.1 & 48.0 & 59.8 &    70.0\\
  HRNet~\cite{hrnet} & 95.3 & 79.9 & 90.7 & 53.7 & 57.4 & 65.9 & 78.4 & 75.9 & 88.8 & 68.6 & 96.1 & 75.5 & 54.0 & 91.2 & 68.2 & 76.2 & 85.4 & 58.4 & 65.1 & 75.0\\
  Mask2Former~\cite{cheng2021mask2former} & 96.2 & 83.9 & 91.9 & 62.0 & 59.7 & 70.4 & 80.4 & 79.0 & 90.4 & 73.0 & 96.7 & 78.2 & 50.8 & 91.3 & 74.9 & 74.3 & 92.9 & 57.0 & 66.1 & 77.3 \\
  ViT-Adapter~\cite{chen2022vitadapter} & 96.4 & 84.6 & 92.2 & 68.0 & 63.7 & 69.8 & 80.5 & 80.0 & 90.2 & 72.6 & 96.4 & 79.0 & 48.8 & 92.0 & 83.1 & 68.7 & 92.3 & 63.8 & 68.1 & 78.4 \\
  \bottomrule
  \end{tabular*}
\end{table*}

In addition, we also train state-of-the-art UDA methods to adapt from Cityscapes to individual conditions of ACDC in Table~\ref{table:uda:det_individual}. The uniformity of target domain in this setting enables a larger performance gain compared to a target domain with mixed conditions in Table~\ref{table:uda:det_all}. We observe that in some conditions, the UDA method presents even worse performance than the source-only model. This is because the adapted model from the final epoch is not the optimal model for the new domain, which also reflects the importance of hyperparameters on different UDA tasks. Moreover, although the adapted models obtain some performance improvement in a certain condition, the gap between the adapted model and the oracle model is still obvious, indicating the difficulty of ACDC for existing UDA object detection methods.

\subsection{Analysis for Normal-to-Adverse Adaptation}
\label{sec:n2a_adapt:analysis}

In this section, we have investigated cross-domain semantic segmentation and object detection methods for adapting from normal conditions to adverse ones, a scenario characterized by a large domain gap.
Table~\ref{table:uda:seg_all:adverse} shows that more recent transformer-based methods using SegFormer-type backbones~\cite{segformer} exhibit greater robustness than older CNN-based methods based on DeepLabv2~\cite{DeepLab:v2} under large domain shifts, consistently achieving higher performance.
Furthermore, among various domain adaptation strategies, self-training proves beneficial across both semantic segmentation and object detection when the domain gap is large. Moreover, approaches involving input-level adaptation, such as FDA~\cite{fda:adaptation} and CISS~\cite{sakaridis2023ciss}, benefit semantic segmentation performance consistently across backbones and conditions.
From a weather-specific perspective, we find that nighttime conditions present the hardest challenge across the examined tasks, whereas rain is the easiest setting for semantic segmentation and snow is the easiest one for object detection.
Since object detection primarily focuses on foreground objects, our experimental results suggest that different categories experience distinct domain shifts even within the same scene.
This observation highlights the potential of category-aware or instance-aware adaptation strategies to further enhance the robustness of perception models under adverse conditions.

%% file: sections/5_adapt2normal.tex
ACDC also contains 4006 reference images captured in normal conditions, i.e.\ daytime and clear weather, to which we will refer in the following as ACDC-Reference. As detailed in Sec.~\ref{sec:dataset:splits}, 1503 of these images have been newly annotated in the present extended version. Thus, we also provide here two new benchmarks for sensor-level adaptation on semantic segmentation and object detection. For this type of adaptation, we use the Cityscapes training set, which is characterized by normal conditions, as the source domain and all ACDC-Reference training images as the target domain. The performance of the adaptation from the camera sensor of Cityscapes to the respective sensor of ACDC is evaluated on the annotated part of the test split of the ACDC-Reference subset, which comprises 500 images. 

\begin{table}[tb]
  \caption{\textbf{Comparison of condition experts vs.\ uber models on the different conditions of ACDC for semantic segmentation.} The first group of rows presents condition-specific expert models trained on a single condition, while the second group presents uber models trained on all adverse conditions. Note that the performance on all conditions is \emph{not} an average of the respective performances on individual conditions.}
  \label{table:supervised:experts_vs_ubers}
  \centering
  \setlength\tabcolsep{2pt}
  \small
  \begin{tabular*}{\linewidth}{l @{\extracolsep{\fill}} ccccc}
  \toprule
  Method & Fog & Night & Rain & Snow & All\\
  \midrule
  RefineNet~\cite{refinenet} & 63.6 & 52.2 & 66.4 & 62.5 & 62.8\\
  DeepLabv2~\cite{DeepLab:v2} & 52.2 & 45.4 & 57.6 & 56.8 & 54.9\\
  DeepLabv3+~\cite{DeepLab:v3+} & 68.7 & 59.2 & 73.5 & 70.5 & 69.6\\
  HRNet~\cite{hrnet} & 70.8 & 63.2 & 72.7 & 70.2 & 70.9\\
  \midrule
  RefineNet~\cite{refinenet} & 65.7 & 55.5 & 68.7 & 65.9 & 65.3\\
  DeepLabv2~\cite{DeepLab:v2} & 54.5 & 45.3 & 59.3 & 57.1 & 55.3\\
  DeepLabv3+~\cite{DeepLab:v3+} & 69.1 & 60.9 & 74.1 & 69.6 & 70.0\\
  HRNet~\cite{hrnet} & 74.7 & 65.3 & 77.7 & 76.3 & 75.0\\
  \bottomrule
  \end{tabular*}
\end{table}

\begin{figure*}
  \centering
  \subfloat{\includegraphics[width=0.198\textwidth]{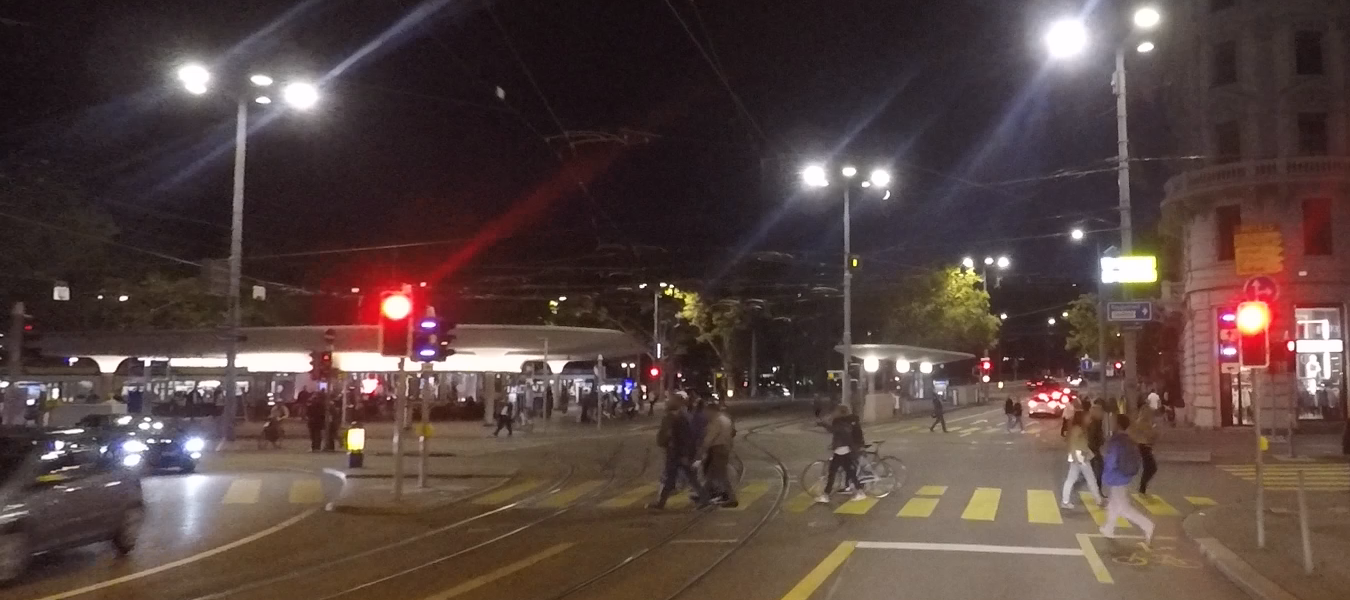}}
  \hfil
  \subfloat{\includegraphics[width=0.198\textwidth]{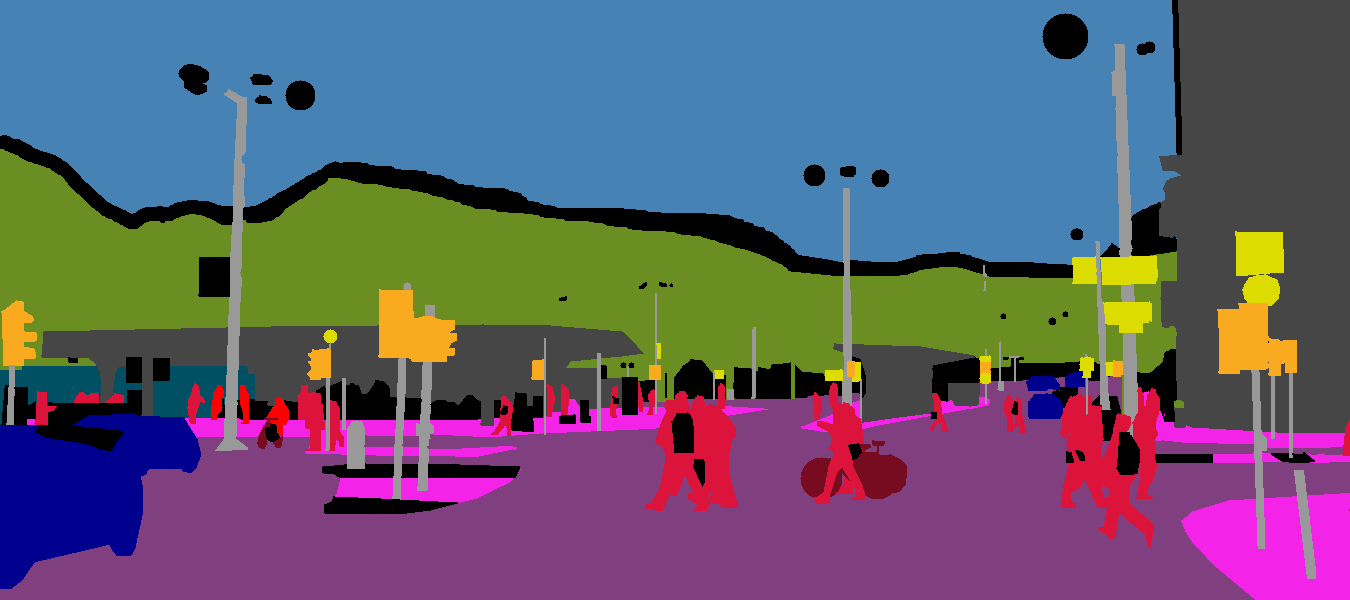}}
  \hfil
  \subfloat{\includegraphics[width=0.198\textwidth]{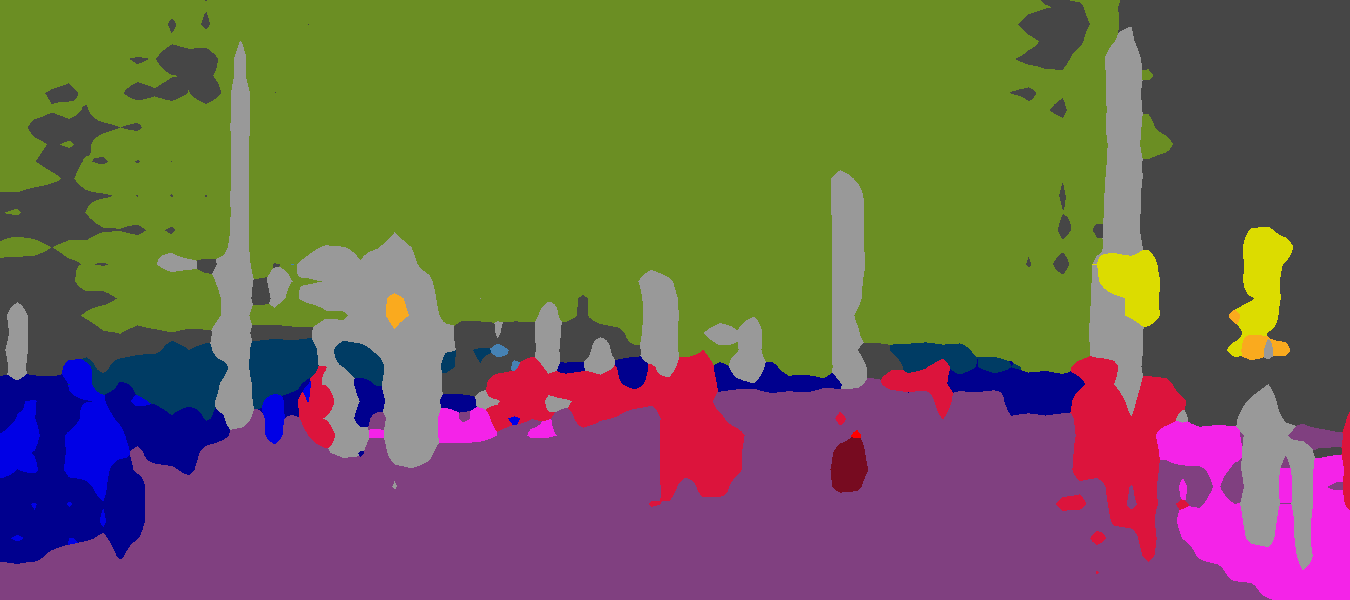}}
  \hfil
  \subfloat{\includegraphics[width=0.198\textwidth]{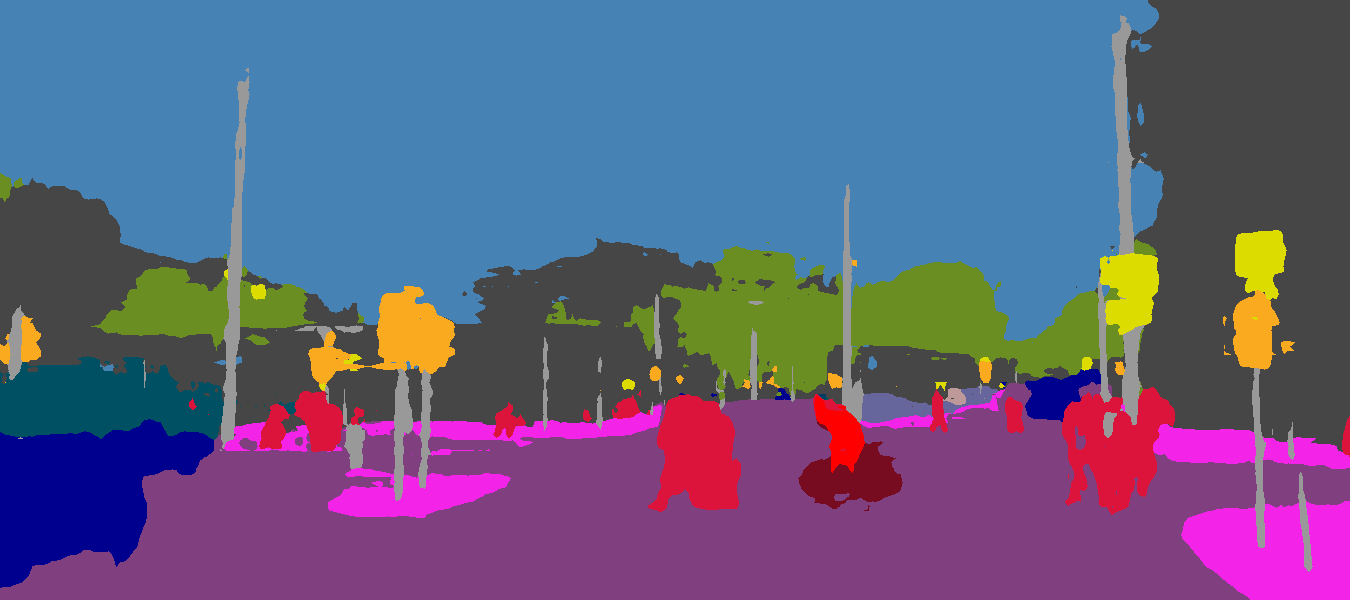}}
  \hfil
  \subfloat{\includegraphics[width=0.198\textwidth]{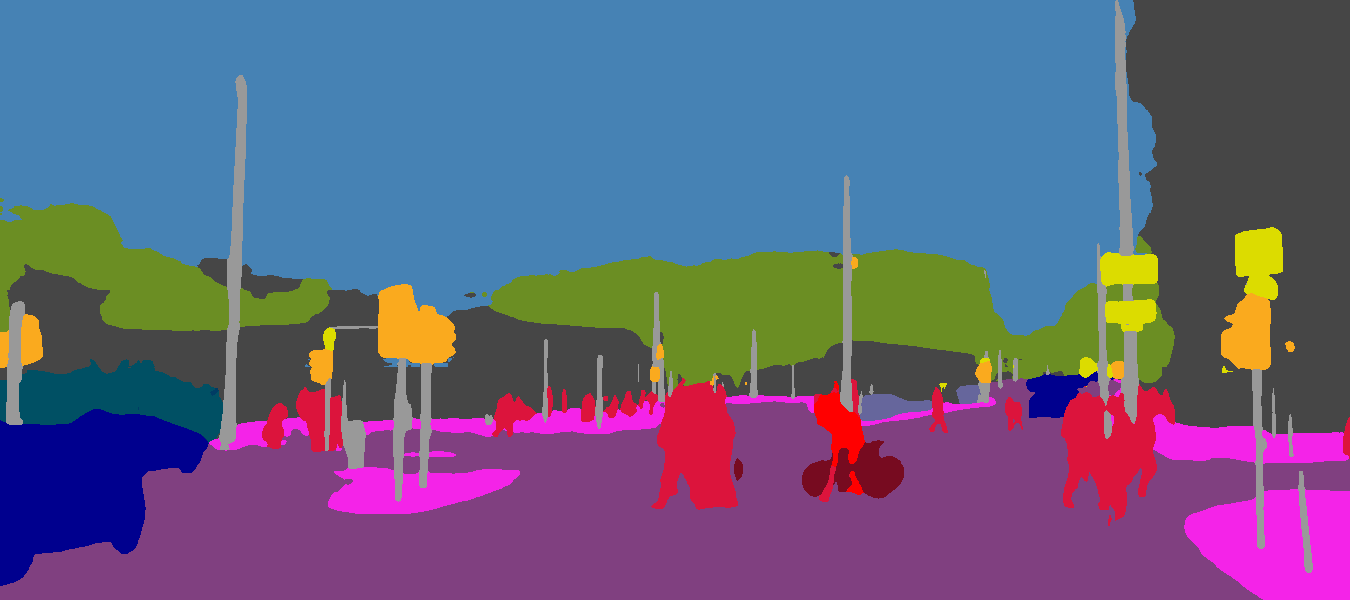}}
  \\
  \vspace{-0.35cm}
  \subfloat{\includegraphics[width=0.198\textwidth]{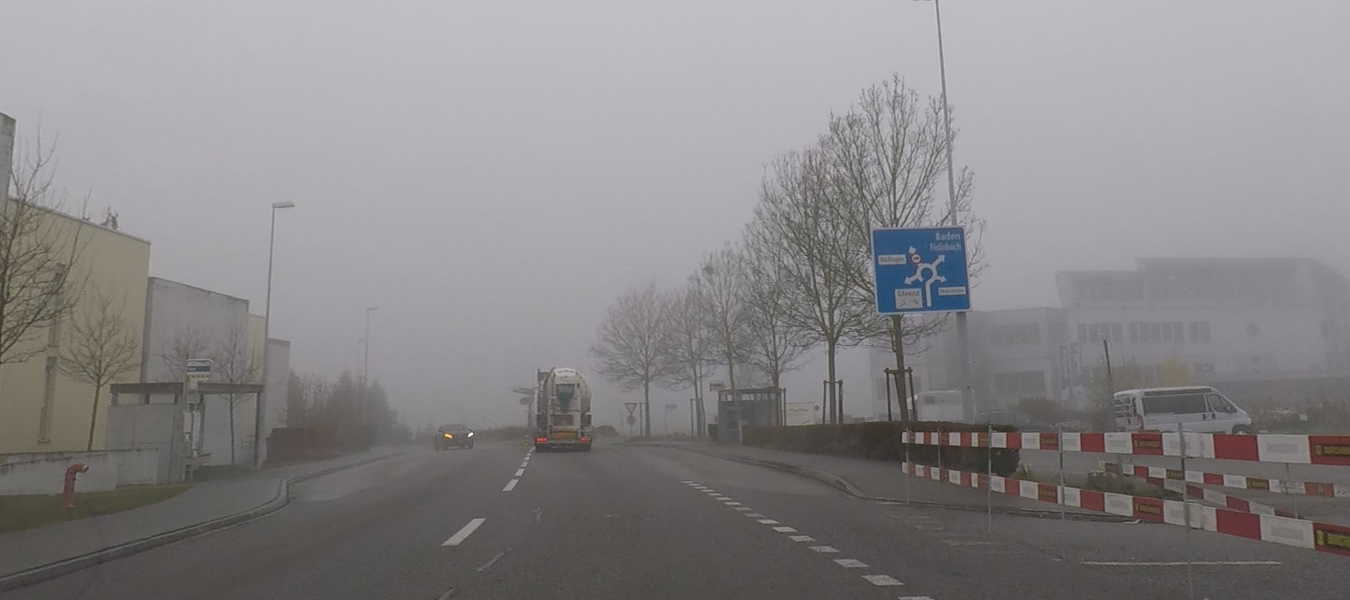}}
  \hfil
  \subfloat{\includegraphics[width=0.198\textwidth]{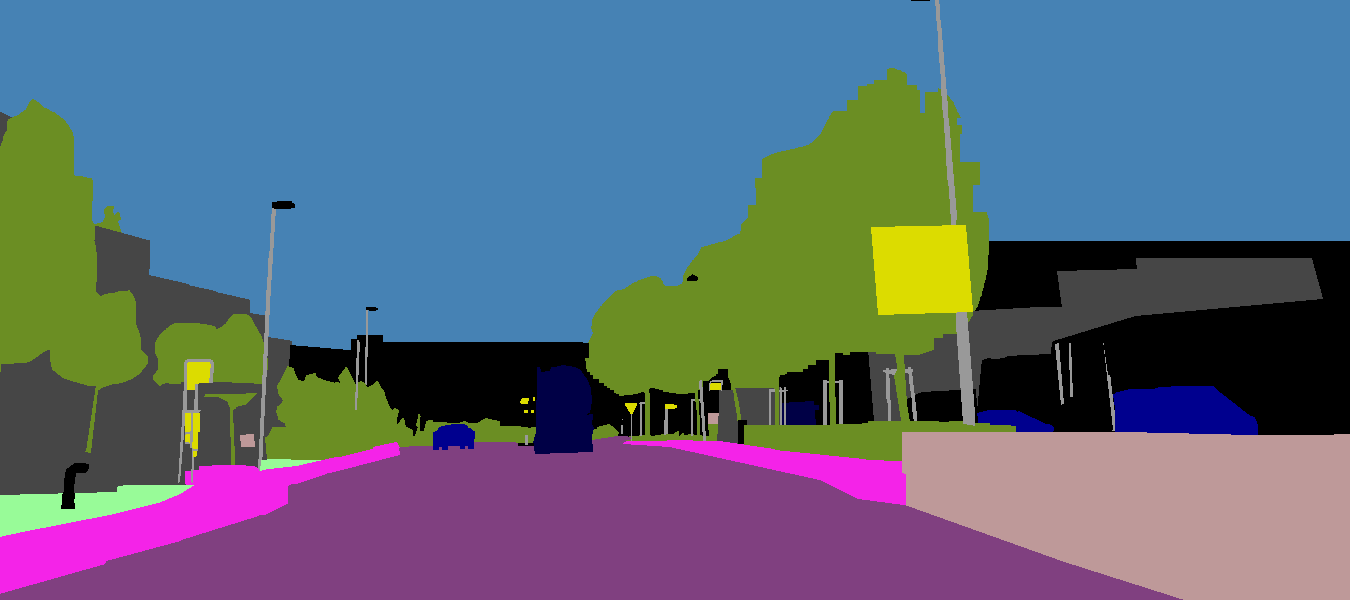}}
  \hfil
  \subfloat{\includegraphics[width=0.198\textwidth]{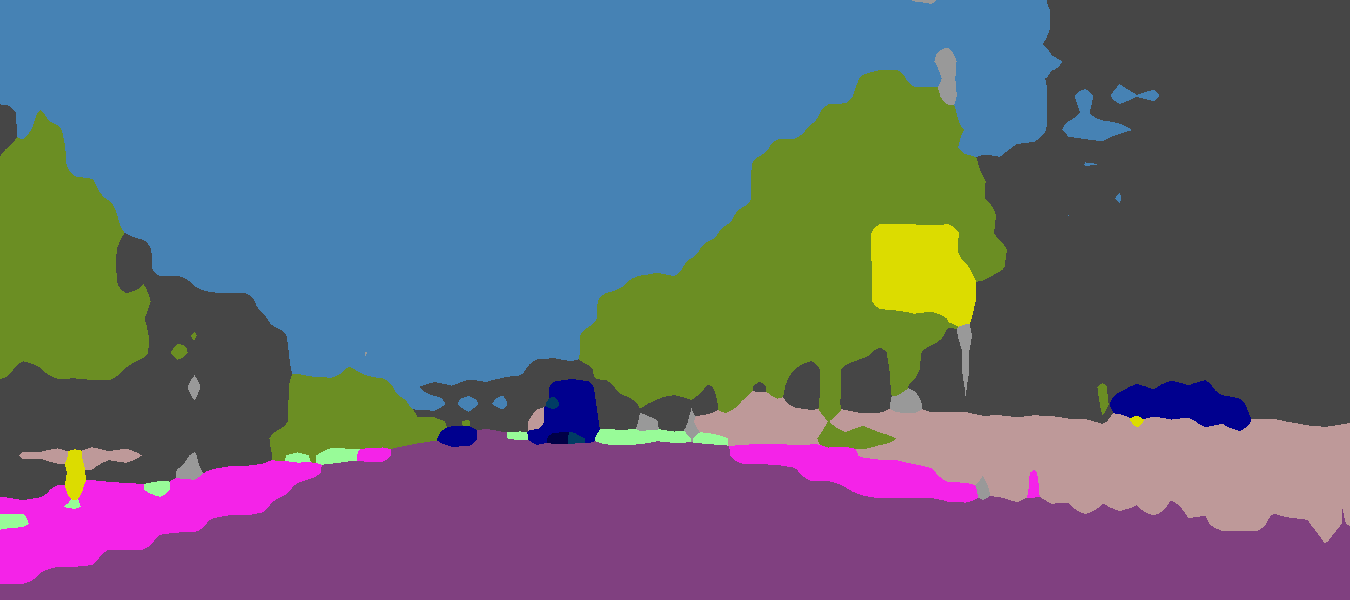}}
  \hfil
  \subfloat{\includegraphics[width=0.198\textwidth]{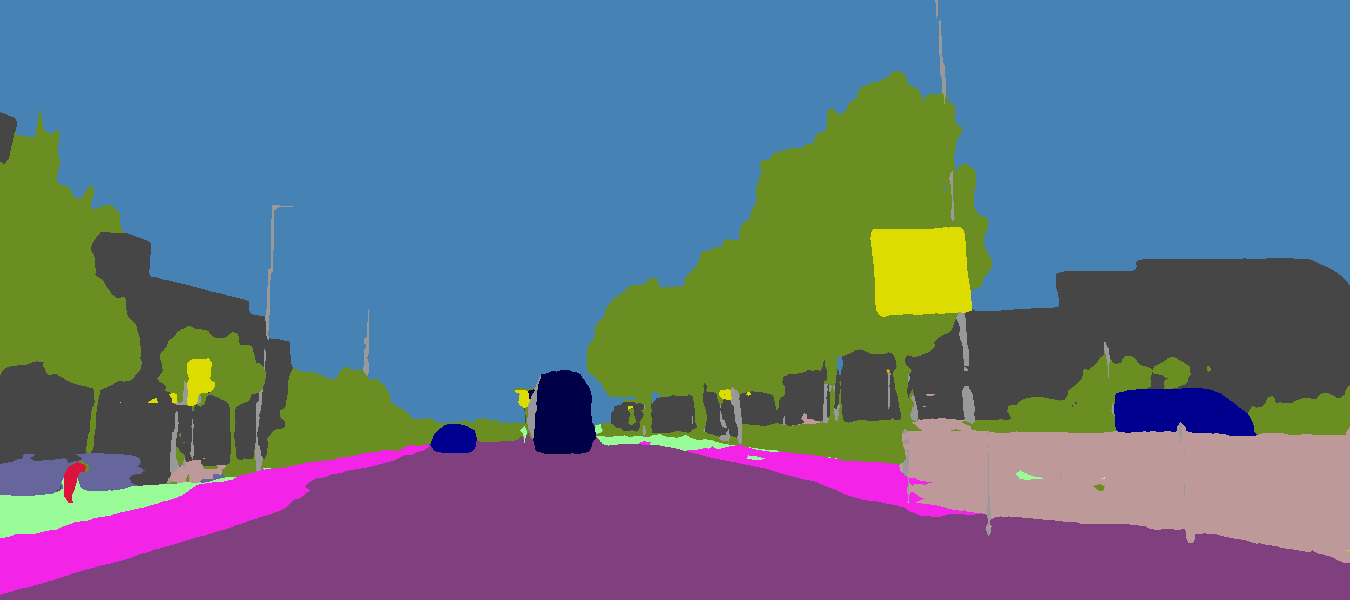}}
  \hfil
  \subfloat{\includegraphics[width=0.198\textwidth]{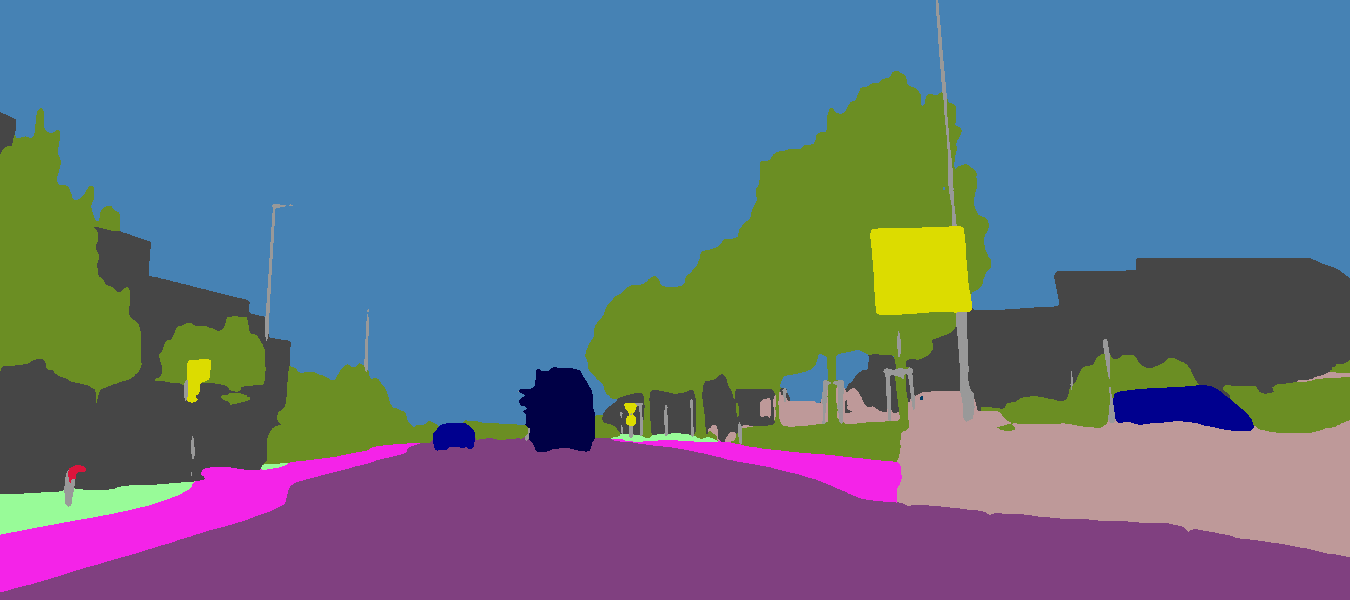}}
  \\
  \vspace{-0.35cm}
  \subfloat{\includegraphics[width=0.198\textwidth]{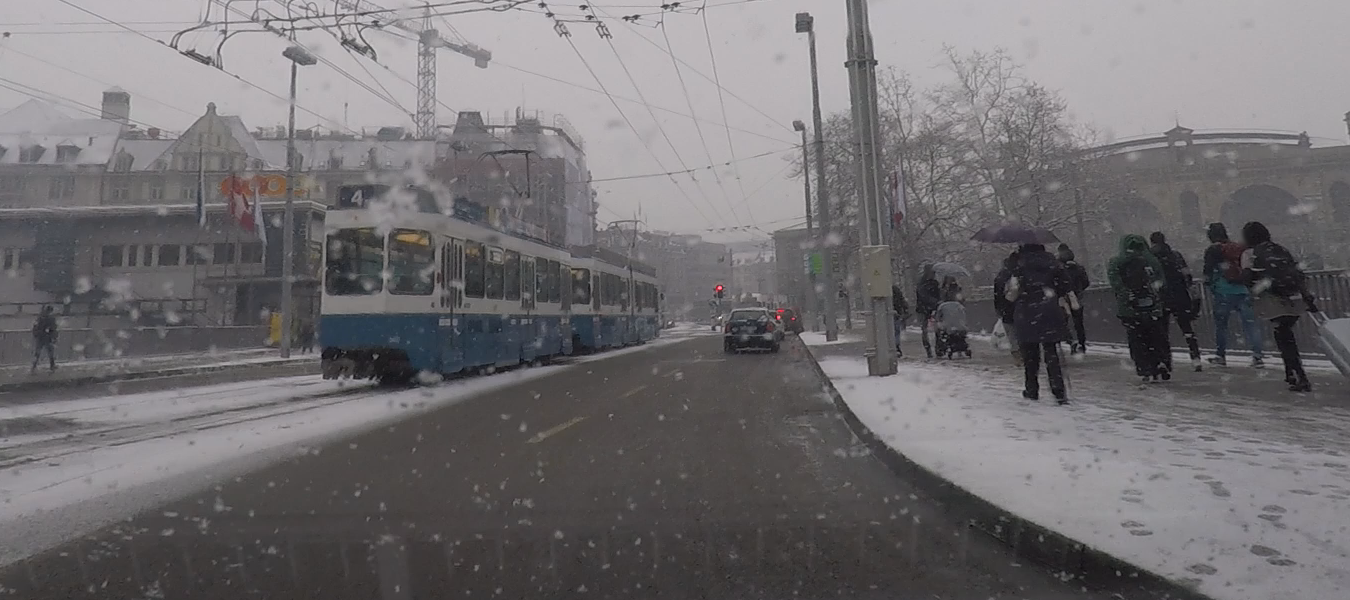}}
  \hfil
  \subfloat{\includegraphics[width=0.198\textwidth]{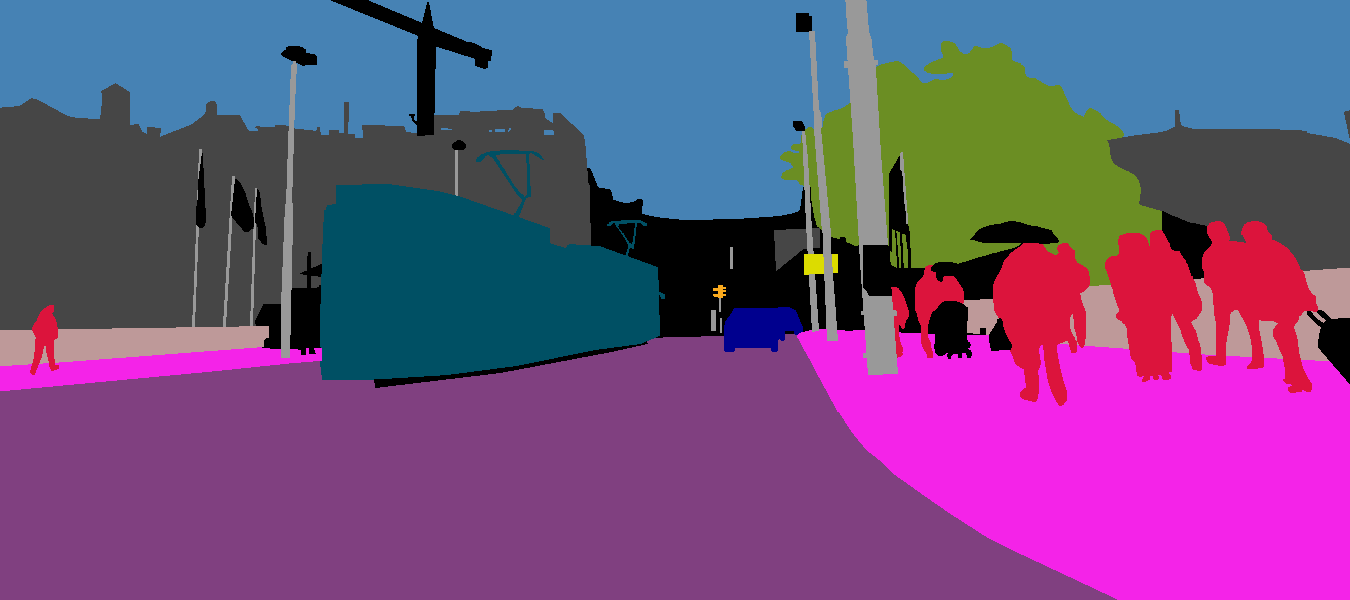}}
  \hfil
  \subfloat{\includegraphics[width=0.198\textwidth]{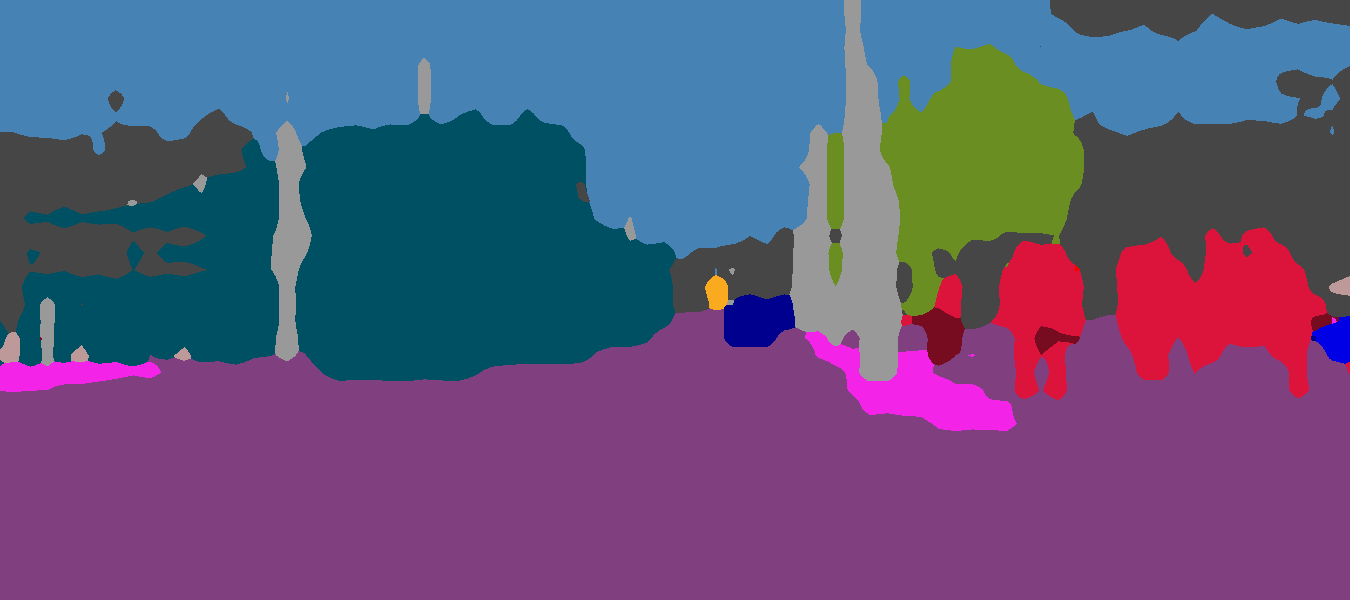}}
  \hfil
  \subfloat{\includegraphics[width=0.198\textwidth]{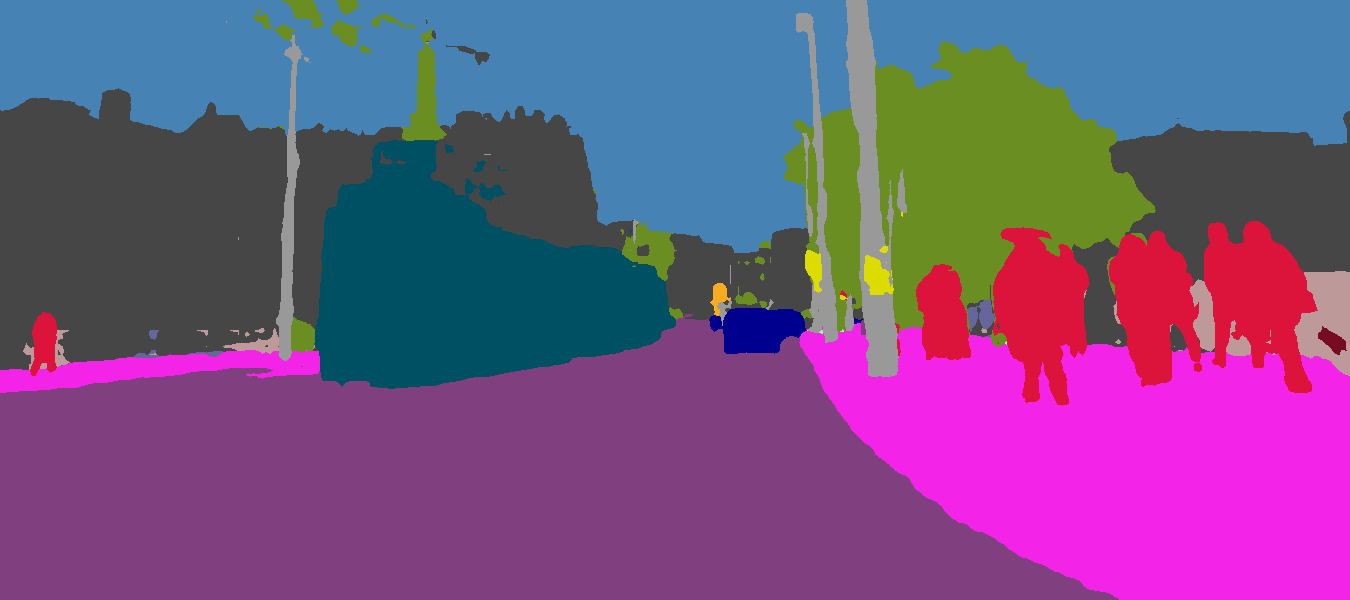}}
  \hfil
  \subfloat{\includegraphics[width=0.198\textwidth]{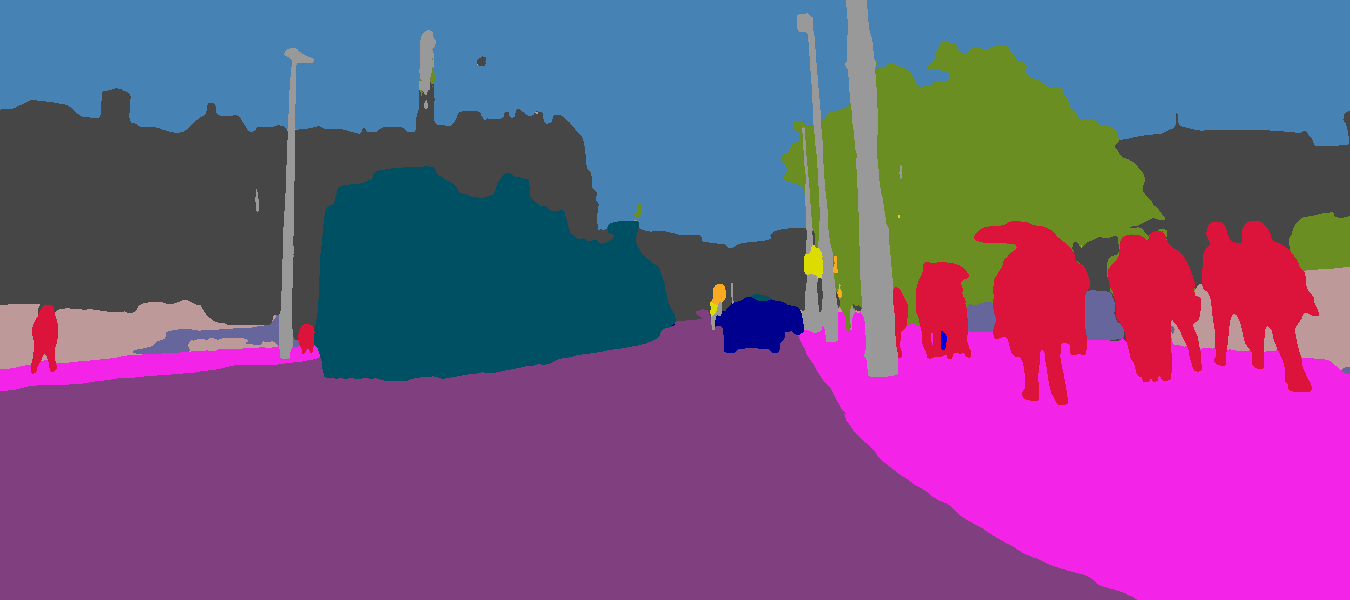}}
  \caption{\textbf{Qualitative results of selected semantic segmentation methods on ACDC.} From left to right: image, ground-truth annotation, FDA~\cite{fda:adaptation}, DeepLabv3+~\cite{DeepLab:v3+}, and HRNet~\cite{hrnet}. The color coding of the semantic classes matches Fig.~\ref{fig:dataset:stats}.}
  \label{fig:semseg}
\end{figure*}

\subsection{Domain-Adaptive Semantic Segmentation}

We introduce a new benchmark for sensor-level real-to-real UDA of semantic segmentation: Cityscapes$\to$ACDC-Reference. We select eleven representative state-of-the-art UDA methods, train them with their default configurations for adaptation from Cityscapes to the ACDC-Reference subset and present the results in Table~\ref{table:uda:seg_all:normal}. Eight of these methods are trained with the earlier DeepLabv2-based architecture~\cite{DeepLab:v2}, while four of them are trained with the more recent SegFormer-based architecture~\cite{segformer} (CISS~\cite{sakaridis2023ciss} is trained with both).

Among the DeepLabv2-based methods, CISS, MRNet and FDA excel on the target-domain test set of ACDC-Reference, as they match (MRNet and FDA) or even exceed (CISS) the mean IoU performance of the oracle model which is trained on labeled images from ACDC-Reference. Thus, we conclude that when focusing on this earlier architecture, the domain gap which is caused by the different sensor characteristics between Cityscapes and ACDC is possible to be closed by state-of-the-art UDA methods. Considering the more recent, SegFormer-based methods, the difference in performance between the source model and the oracle model becomes only slight, namely 4.0\% in mean IoU. The three top-performing methods, namely CISS, MIC and HRDA, significantly surpass the performance of the oracle model, indicating that in this domain adaptation setting, the inductive biases which are introduced to the models by the respective aforementioned domain adaptation and generalization strategies boost the target-domain performance even more than access to in-domain training data does, which only the oracle model enjoys.

\begin{table*}[tb]
  \caption{\textbf{Comparison of class-level performance of DeepLabv3+ condition experts on the various conditions of ACDC.} A different model is trained on each individual condition and then evaluated on this condition.}
  \label{table:supervised:individual:deeplabv3+}
  \centering
  \setlength\tabcolsep{2pt}
  \footnotesize
  \begin{tabular*}{\linewidth}{l @{\extracolsep{\fill}} cccccccccccccccccccc}
  \toprule
  Condition & \ver{road} & \ver{sidew.} & \ver{build.} & \ver{wall} & \ver{fence} & \ver{pole} & \ver{light} & \ver{sign} & \ver{veget.} & \ver{terrain} & \ver{sky} & \ver{person} & \ver{rider} & \ver{car} & \ver{truck} & \ver{bus} & \ver{train} & \ver{motorc.} & \ver{bicycle} & mIoU\\
  \midrule
  Fog & 93.8 & 77.4 & 88.8 & 51.0 & 43.3 & 54.2 & 68.2 & 71.7 & 87.7 & 74.6 & 98.2 & 53.5 & 32.1 & 83.8 & 69.3 & 84.4 & 85.3 & 47.2 & 40.1 &     68.7\\
  Night & 94.7 & 75.9 & 85.0 & 48.4 & 38.6 & 52.2 & 55.8 & 54.4 & 76.1 & 30.3 & 84.2 & 67.4 & 41.1 & 85.0 & 8.3 & 62.3 & 80.6 & 35.6 & 49.8 &     59.2\\
  Rain & 92.8 & 77.4 & 93.9 & 67.3 & 58.1 & 64.1 & 74.4 & 75.9 & 94.2 & 50.8 & 98.6 & 70.8 & 33.4 & 90.4 & 67.7 & 79.2 & 86.8 & 54.6 & 66.1 &   73.5\\
  Snow & 91.9 & 70.9 & 90.1 & 48.9 & 52.0 & 62.2 & 79.2 & 74.5 & 92.0 & 47.0 & 97.6 & 78.2 & 35.9 & 90.4 & 61.7 & 64.3 & 89.2 & 43.9 & 69.4 &   70.5\\
  \bottomrule
  \end{tabular*}
\end{table*}

\begin{table}[tb]
  \caption{\textbf{Cross-evaluation of DeepLabv3+ condition experts on the various adverse conditions of ACDC.} Each model is trained on an individual condition and evaluated on each condition separately. Performance of the Cityscapes pre-trained model is also reported.}
  \label{table:supervised:individual:deeplabv3+:cross}
  \centering
  \setlength\tabcolsep{1.5pt}
  \small
  \begin{tabular}{lcccc}
  \toprule
  Train/Eval & Fog & Night & Rain & Snow\\
  \midrule
  Normal & 45.7 & 25.0 & 50.0 & 42.0 \\
  \midrule
  Fog   & 68.7 & 40.7 & 63.5 & 59.1 \\
  Night & 58.5 & 59.2 & 55.6 & 49.6 \\
  Rain  & 65.2 & 46.0 & 73.5 & 63.5 \\
  Snow  & 59.2 & 38.0 & 69.3 & 70.5 \\
  \bottomrule
  \end{tabular}
\end{table}

\subsection{Domain-Adaptive Object Detection}

We present the sensor-level object detection adaptation results in Table~\ref{table:uda:det_ref_all}. According to the results, although both Cityscapes and ACDC-Reference contain images captured in normal conditions, there still exists a domain gap between Cityscapes and ACDC-Reference. If we take the performance gap between the source-only model and the oracle model as an indicator of the domain gap, Cityscapes has a smaller domain gap to ACDC-Reference set compared to the adverse-condition part of ACDC.

We observe that on common categories such as person and car, state-of-the-art UDA methods for detection obtain equal or better performance compared to the respective oracle models. However, for the less frequent categories such as truck and bus, even if the domain gap is small, there is still an obvious performance gap. This indicates that how to effectively mine the knowledge from these rare categories remains a pressing research question for the area of domain-adaptive object detection.

\subsection{Analysis for Sensor-Level Adaptation}
\label{sec:n2n_adapt:analysis}

This section has examined adaptation from Cityscapes to ACDC-Reference. Since both datasets contain urban images captured under clear weather, the domain gap is considerably smaller than in the normal-to-adverse adaptation setting.
Consistent with the previous analysis in Sec.~\ref{sec:n2a_adapt:analysis}, self-training–based methods as well as input-level adaptation methods---two sets which are not mutually exclusive here---achieve the best performance either on semantic segmentation or object detection under a small domain gap.
Interestingly, under this less complex domain shift, these types of approaches even surpass the oracle model in the semantic segmentation task, whereas domain-adaptive object detection methods fail to do so against their competing oracles.
We attribute this difference to the disparity in supervision density between the two tasks: semantic segmentation provides dense pixel-level labels, while object detection offers only sparse bounding-box annotations.
This finding underscores the importance of developing task-specific adaptation strategies that account for the supervision granularity of different perception tasks.

%% file: sections/6_supervised.tex
In this section, we benchmark several supervised methods for different central dense semantic perception tasks, including semantic segmentation, instance segmentation, and panoptic segmentation, on ACDC.

\input{figures/panoptic_vis_res}

\subsection{Semantic Segmentation}
\label{sec:supervised:semseg}

\begin{table*}[tb]
  \caption{\textbf{Comparison of state-of-the-art supervised instance segmentation methods on ACDC.} Training and evaluation are performed using the training and test sets of the entire adverse-condition part of ACDC including all four adverse conditions, respectively.}
  \label{table:supervised:instance_all}
  \centering
  \setlength\tabcolsep{2pt}
  \footnotesize
  \begin{tabular*}{\linewidth}{l @{\extracolsep{\fill}} ccccccc}
  \toprule
  Method & AP$^{box}$ & AP$_{50}^{box}$ & AP$_{75}^{box}$ & AP$^{mask}$ & AP$_{50}^{mask}$ & AP$_{75}^{mask}$\\
  \midrule
  Mask R-CNN~\cite{he2017maskrcnn}  & 22.7 & 43.4 & 20.9 & 20.7 & 39.7 & 19.0 \\
  Cascaded Mask R-CNN~\cite{cai2019cascadercnn} & 24.4 & 42.1 & 24.1 & 21.3 & 39.5 & 20.4  \\
  HTC~\cite{chen2019htc}         & 26.0 & 45.2 & 26.0 & 23.0 & 41.9 & 22.0 \\
  Detectors~\cite{qiao2020detectors} & 26.6 & 44.3 & 27.0 & 23.5 & 41.8 & 22.3  \\
  \bottomrule
  \end{tabular*}
\end{table*}

We use ACDC to train six state-of-the-art supervised semantic segmentation methods and report their performance in Table~\ref{table:supervised:all}. Qualitative results are shown in Fig.~\ref{fig:semseg} for two supervised methods and one UDA method. We draw the following conclusions: (1) full supervision in adverse conditions is more valuable than designing a better architecture trained solely on normal conditions, as even an earlier method~\cite{DeepLab:v2} performs better with full supervision than the top-performing externally pre-trained model (cf.\ Table~\ref{table:pretrained_seg}). (2) ACDC is a challenging benchmark for supervised methods due to its hard visual domains; even the very recent ViT-Adapter~\cite{chen2022vitadapter} scores only 78.4\% mIoU on the test set, which is 6.8\% lower than its respective performance of 85.2\% on Cityscapes~\cite{hrnet}. (3) The rankings of the supervised and the pre-trained models do not correlate well, as can be seen from the results in Tables~\ref{table:supervised:all} and \ref{table:pretrained_seg}.

The last point suggests that state-of-the-art networks such as HRNet have enough capacity to overfit to datasets such as Cityscapes, which would explain the low performance of the Cityscapes pre-trained HRNet model on ACDC. We test this hypothesis by training HRNet \emph{jointly on Cityscapes and ACDC}; our expectation is that the jointly trained model will at least match the performance of the individually trained models on each dataset. This is confirmed, as the jointly trained model gets 81.2\% mIoU on Cityscapes and 74.8\% on ACDC, beating and being on a par with the respective individually trained models. Thus, even if ACDC is not of very large scale, it helps to efficiently regularize segmentation models for normal conditions as well.

Table~\ref{table:supervised:experts_vs_ubers} compares models trained on a single adverse condition, termed condition experts, against models trained on the entire training set, termed uber models. Each condition expert is evaluated on the condition it has been trained on. The uber models generally beat the respective condition experts across different conditions and segmentation networks. This hints that the capacity of these networks is large enough to discover discriminative representations for all conditions simultaneously. We also evaluate ensembles of condition experts against uber models on the complete test set (``All''), where the ensemble uses the expert corresponding to the condition of the input image for prediction. Again, the uber models outperform the ensembles of experts for all examined methods. Moreover, all methods perform worst at nighttime, indicating that the nighttime set of ACDC represents a harder domain than the other sets.

We focus on the widely used DeepLabv3+ network~\cite{DeepLab:v3+} for a detailed study of class-level performance across different conditions and compare the performance of the four condition experts in Table~\ref{table:supervised:individual:deeplabv3+}. We make the following observations: (1) the lowest performance for \emph{road} and \emph{sidewalk} occurs in snow, which can be attributed to confusion between the two classes due to similar appearance in the presence of snow cover. (2) Classes that usually appear dark or are not well-lit at nighttime, e.g., \emph{building}, \emph{vegetation}, \emph{traffic sign}, and \emph{sky}, are harder to segment at nighttime. (3) Performance on classes with instances of small size, such as \emph{person}, \emph{rider}, and \emph{bicycle}, is lowest on fog, probably due to the combined effect of contrast reduction and low resolution for instances of these classes that are far from the camera.

We also evaluate in Table~\ref{table:supervised:individual:deeplabv3+:cross} the four DeepLabv3+ condition experts on conditions that are not encountered at training. Excluding nighttime, the results are close to symmetric with respect to training versus evaluation condition; e.g., training on fog and testing on snow results in a similar performance to training on snow and testing on fog. In contrast, performance of the night expert on other conditions is much higher than performance of other experts at night, implying that representations learned from the nighttime domain can generalize better to the other adverse conditions than vice versa.

\subsection{Instance Segmentation}

We use ACDC to train four popular supervised instance segmentation methods and report their performance in Table ~\ref{table:supervised:instance_all}. Detectors obtain the best performance in object detection with 26.6\% $AP^{box}$ and instance segmentation with 23.5\% $AP^{mask}$ simultaneously. Although HTC presents better $AP_{50}^{box}$ and $AP_{50}^{mask}$ than Detectors, Detectors exhibit a better capability of localization and achieve better performance at higher IoU thresholds, namely on $AP_{75}^{box}$ and $AP_{75}^{mask}$.

In Table~\ref{table:supervised:instance_experts_vs_ubers}, we compare the instance segmentation performance of condition experts versus uber models, similarly to Table~\ref{table:supervised:experts_vs_ubers}. Generally, across different instance segmentation architectures, uber models outperform condition experts, which are only optimized on a single condition. Night is still the most challenging condition for all methods. We also observe that the performance gap between the uber model and the condition expert model is the highest in fog, which indicates that images exhibiting different appearance shifts from the one induced by fog can still benefit a model's robustness to fog a lot, even though such images are characterized by a significant domain gap to the target condition of fog.

\begin{table}[!tb]
  \caption{\textbf{Comparison of condition experts vs.\ uber models on the different conditions of ACDC for instance segmentation.} The first group of rows presents condition-specific expert models trained on a single condition, while the second group presents uber models trained on all adverse conditions. For each condition we report the performance in $AP^{mask}$ separately. Note that the performance on all conditions is \emph{not} an average of the respective performances on individual conditions.}
  \label{table:supervised:instance_experts_vs_ubers}
  \centering
  \setlength\tabcolsep{2pt}
  \small
  \begin{tabular*}{\linewidth}{l @{\extracolsep{\fill}} ccccc}
  \toprule
  Method & Fog & Night & Rain & Snow & All\\
  \midrule
  Mask R-CNN~\cite{he2017maskrcnn}          & 15.6 & 10.7 & 21.3 & 20.8 & 16.8\\
  Cascaded MRCNN~\cite{cai2019cascadercnn}      & 16.2 & 11.5 & 21.2 & 22.3 & 17.9\\
  HTC~\cite{chen2019htc}                 & 17.3 & 12.4 & 22.3 & 23.4 & 18.8\\
  Detectors~\cite{qiao2020detectors}           & 17.4 & 13.1 & 23.3 & 23.4 & 19.0\\
  \midrule
  Mask R-CNN~\cite{he2017maskrcnn}          & 24.4 & 14.2 & 21.6 & 27.4 & 20.7 \\
  Cascaded MRCNN~\cite{cai2019cascadercnn}      & 24.3 & 13.9 & 22.5 & 28.2 & 21.3 \\
  HTC~\cite{chen2019htc}                 & 26.0 & 15.4 & 23.2 & 30.2 & 23.0 \\
  Detectors~\cite{qiao2020detectors}           & 25.3 & 16.5 & 24.9 & 29.8 & 23.5 \\
  \bottomrule
  \end{tabular*}
\end{table}

\subsection{Panoptic Segmentation}
We use ACDC to train four popular supervised panoptic segmentation methods and report their performance in Table ~\ref{table:supervised:panoptic_all}. Qualitative results are shown in Fig.~\ref{fig:panseg}. Mask2Former obtains the best $PQ$ and $PQ^{stuff}$ performance among these methods, while a simple PanopticFPN obtains the best $PQ^{things}$. We also present a comparison between condition experts and uber models for panoptic segmentation in Table~\ref{table:supervised:panoptic_experts_vs_ubers}. Mask2Former exhibits advantages in most conditions. At the same time, uber models outperform most condition experts by a large margin. Interestingly, we observe that unlike supervised semantic segmentation and instance segmentation, in rain, the uber models are roughly on a par with the rain experts across the four examined architectures. This indicates that the domain shifts in other conditions with respect to rain provide limited help in distinguishing the categories in rain.

\begin{table}[!tb]
  \caption{\textbf{Comparison of state-of-the-art supervised panoptic segmentation methods on ACDC.} Training and evaluation are performed using the training and test sets of the entire adverse-condition part of ACDC including all four adverse conditions, respectively.}
  \label{table:supervised:panoptic_all}
  \centering
  \setlength\tabcolsep{2pt}
  \small
  \begin{tabular*}{\linewidth}{l @{\extracolsep{\fill}} ccccc}
  \toprule
  Method & PQ & PQ$^{things}$ & PQ$^{stuff}$ & SQ & RQ\\
  \midrule
  PanopticFPN~\cite{Kirillov_2019_CVPR_panoptic_fpn} & 43.9 & 36.4 & 49.3 & 77.6 & 54.6 \\
  K-Net~\cite{NEURIPS2021_knet} & 47.2 & 30.5 & 59.4 & 77.8 & 58.8 \\
  Panoptic-Deeplab~\cite{cheng2020panoptic_deeplab} & 49.4 & 35.5  & 59.5 & 79.7 & 60.1 \\
  Mask2Former~\cite{cheng2021mask2former} & 49.8 & 33.9 & 61.3 & 80.0 & 60.7 \\
  \bottomrule
  \end{tabular*}
\end{table}

\begin{table}[!tb]
  \caption{\textbf{Comparison of condition experts vs.\ uber models on the different conditions of ACDC for panoptic segmentation.} The first group of rows presents condition-specific expert models trained on a single condition, while the second group presents uber models trained on all adverse conditions. For each case we report the performance in $PQ$. Note that the performance on all conditions is \emph{not} an average of the respective performances on individual conditions.}
  \label{table:supervised:panoptic_experts_vs_ubers}
  \centering
  \setlength\tabcolsep{2pt}
  \small
  \begin{tabular*}{\linewidth}{l @{\extracolsep{\fill}} ccccc}
  \toprule
  Method & Fog & Night & Rain & Snow & All\\
  \midrule
  PanopticFPN~\cite{Kirillov_2019_CVPR_panoptic_fpn}      & 38.4 & 29.8 & 46.7 & 44.8 & 41.3 \\
  K-Net~\cite{NEURIPS2021_knet}            & 37.9 & 30.7 & 48.5 & 48.0 & 43.3 \\
  Panoptic-Deeplab~\cite{cheng2020panoptic_deeplab} & 42.4 & 34.1 & 52.7 & 51.6 & 46.7 \\
  Mask2Former~\cite{cheng2021mask2former}      & 44.9 & 34.0 & 53.0 & 52.5 & 47.1 \\
  \midrule
  PanopticFPN~\cite{Kirillov_2019_CVPR_panoptic_fpn}      & 43.9 & 32.6 & 43.9 & 49.1 & 44.3 \\
  K-Net~\cite{NEURIPS2021_knet}            & 47.8 & 33.4 & 47.1 & 53.2 & 45.6 \\
  Panoptic-Deeplab~\cite{cheng2020panoptic_deeplab} & 49.1 & 37.2 & 53.1 & 55.1 & 49.4 \\
  Mask2Former~\cite{cheng2021mask2former}      & 52.9 & 39.4 & 54.2 & 58.6 & 51.1 \\
  \bottomrule
  \end{tabular*}
\end{table}

\subsection{Analysis for Supervised Learning on Adverse Conditions}
\label{sec:supervised:analysis}

Comparing the results from Sec.~\ref{sec:n2a_adapt} and the present section, we deduce that simply leveraging direct supervision on adverse-condition data is more effective for semantic segmentation in adverse conditions than substantially upgrading the network backbone and architecture. Moreover, we have demonstrated that training a generalist semantic, instance, or panoptic segmentation model jointly on all normal- and adverse-condition data generally benefits performance across all domains, both the individual adverse-condition ones and the normal-condition one, compared to training separate condition specialists, with the exception of panoptic segmentation and rain. This observation holds despite the fact that the adverse-condition data are relatively small. Across semantic segmentation, instance segmentation, and panoptic segmentation, our results show that transformer-based methods consistently exhibit stronger generalization than CNN-based methods when trained on the same data.
For CNN-based architectures, high-resolution feature representations enhance robustness across diverse domains, while effective multi-scale feature fusion is critical for strong in-domain performance.
With carefully designed mechanisms, CNN-based methods can achieve performance close to that of transformer-based models in various perception tasks, narrowing the performance gap despite their inherent differences in architecture. Regarding performance on individual classes across different conditions, we have examined the semantic segmentation setting in class-specific detail in Table~\ref{table:supervised:individual:deeplabv3+} and found a non-negligible correlation between the intensity of the effect of each adversity on different classes and the respective class-level performance. Finally, we have evidenced in Table~\ref{table:supervised:individual:deeplabv3+:cross} for the task of semantic segmentation that specialist models trained on harder conditions generalize better to different yet easier conditions than vice versa.

%% file: figures/panoptic_vis_res.tex
\begin{figure*}
  \centering
  \subfloat{\includegraphics[width=0.198\textwidth]{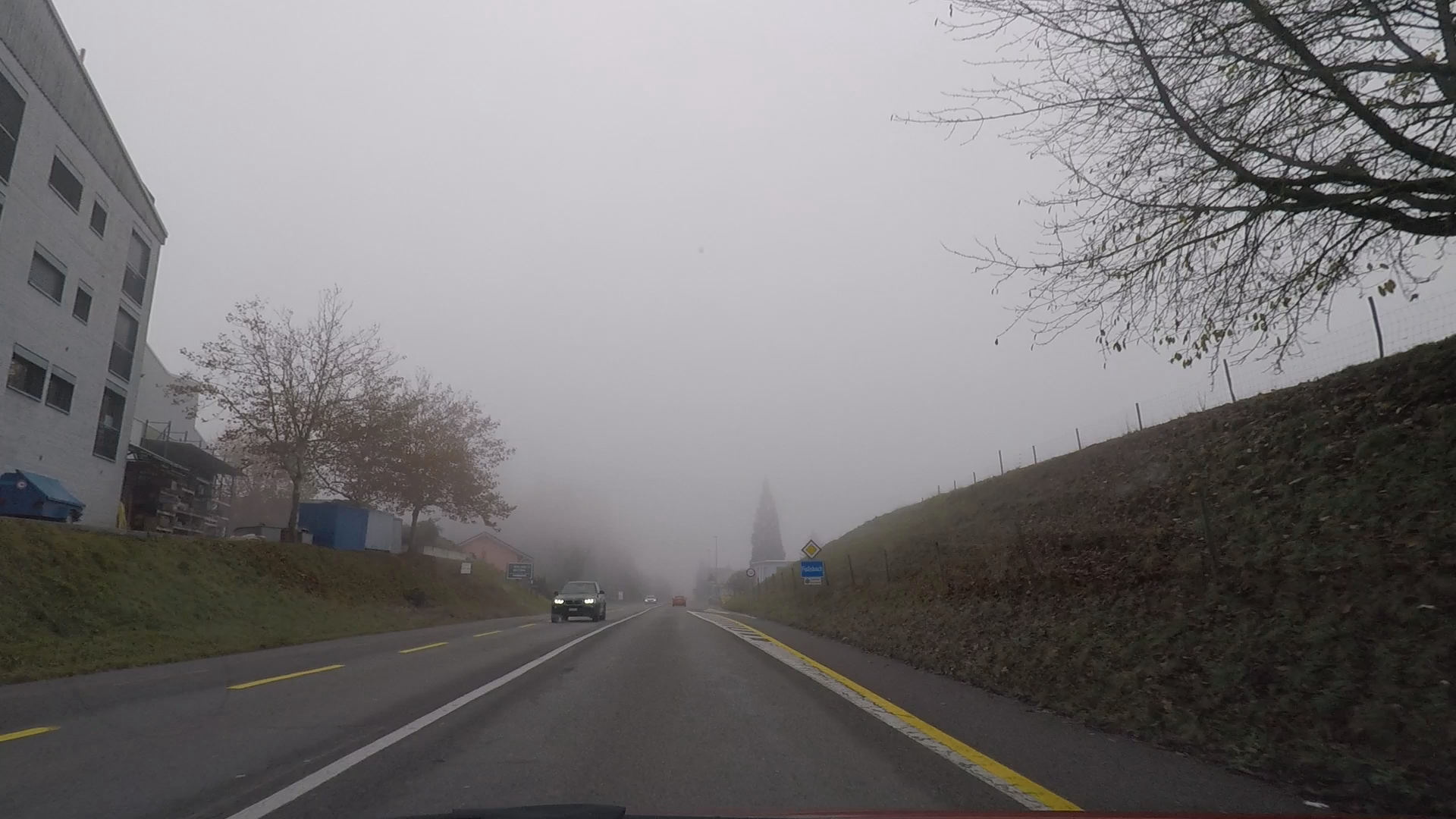}}
  \hfil
  \subfloat{\includegraphics[width=0.198\textwidth]{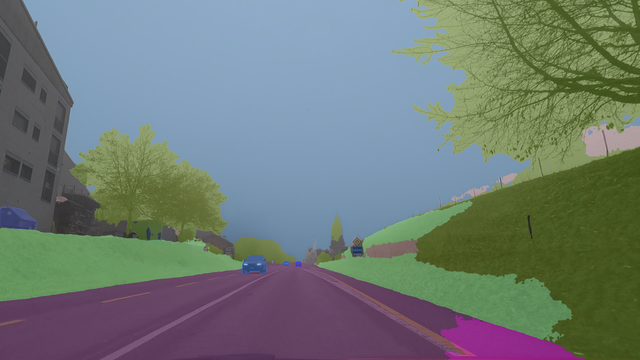}}
  \hfil
  \subfloat{\includegraphics[width=0.198\textwidth]{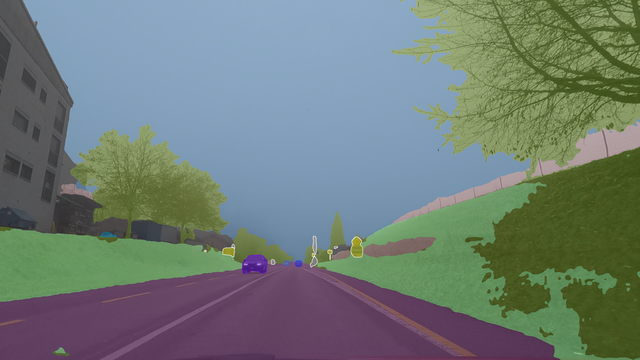}}
  \hfil
  \subfloat{\includegraphics[width=0.198\textwidth]{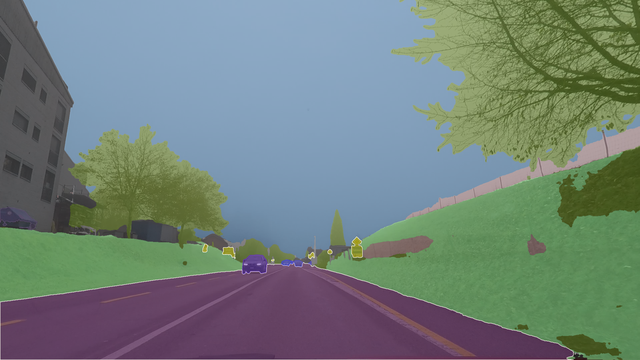}}
  \hfil
  \subfloat{\includegraphics[width=0.198\textwidth]{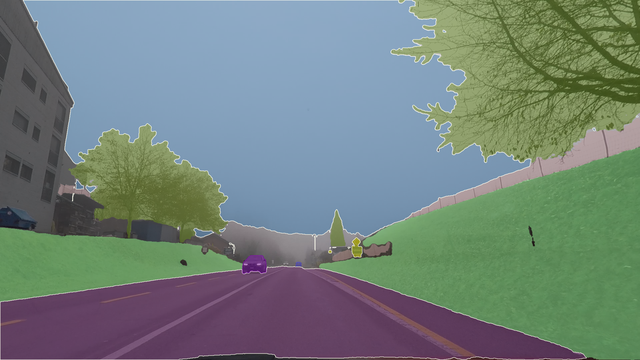}}
  \\
  \vspace{-0.35cm}
  \subfloat{\includegraphics[width=0.198\textwidth]{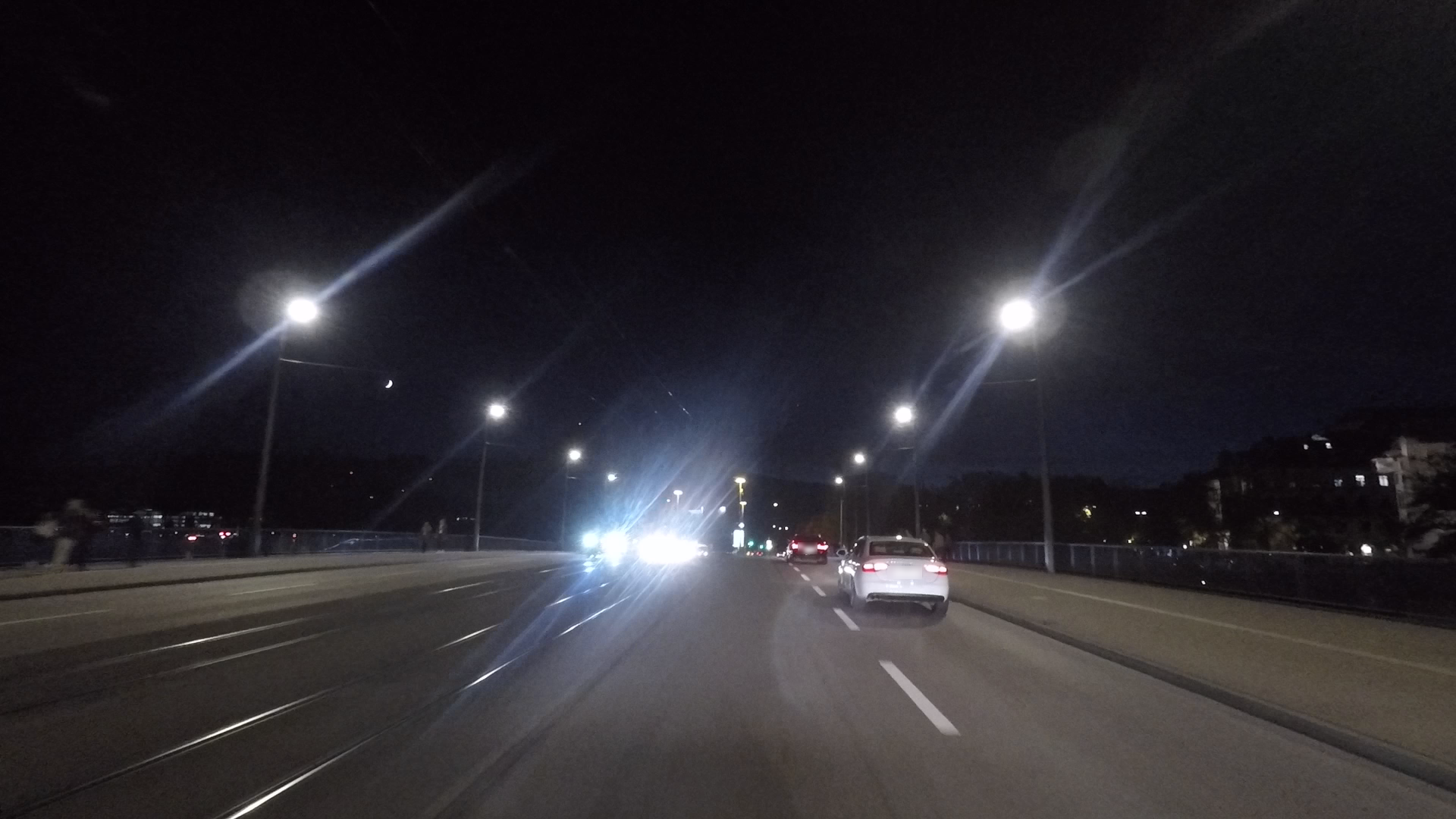}}
  \hfil
  \subfloat{\includegraphics[width=0.198\textwidth]{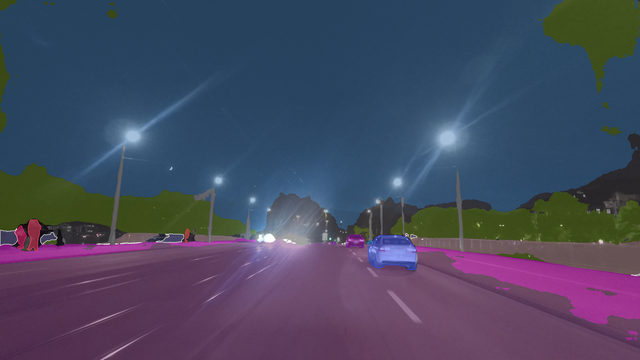}}
  \hfil
  \subfloat{\includegraphics[width=0.198\textwidth]{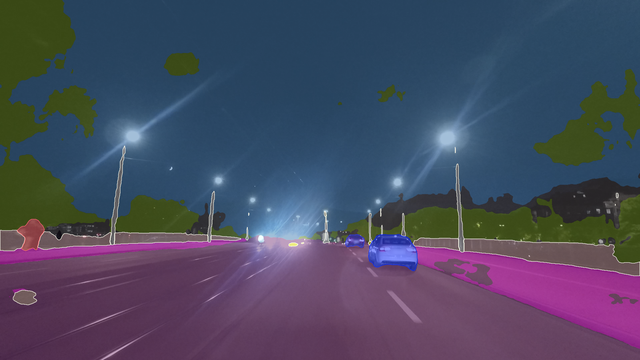}}
  \hfil
  \subfloat{\includegraphics[width=0.198\textwidth]{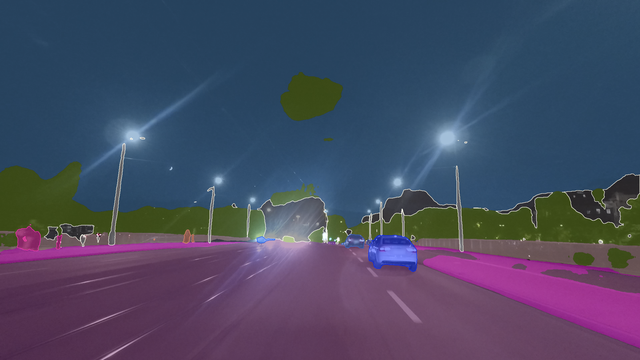}}
  \hfil
  \subfloat{\includegraphics[width=0.198\textwidth]{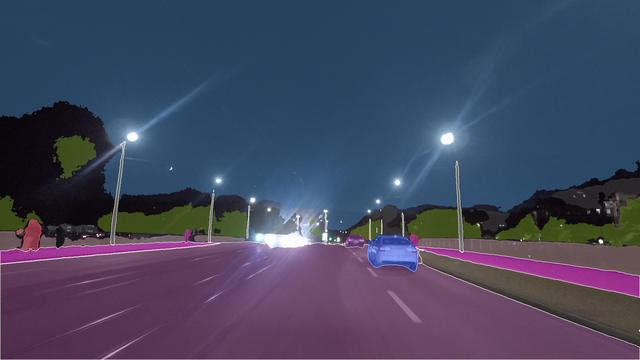}}
  \\
  \vspace{-0.35cm}
  \subfloat{\includegraphics[width=0.198\textwidth]{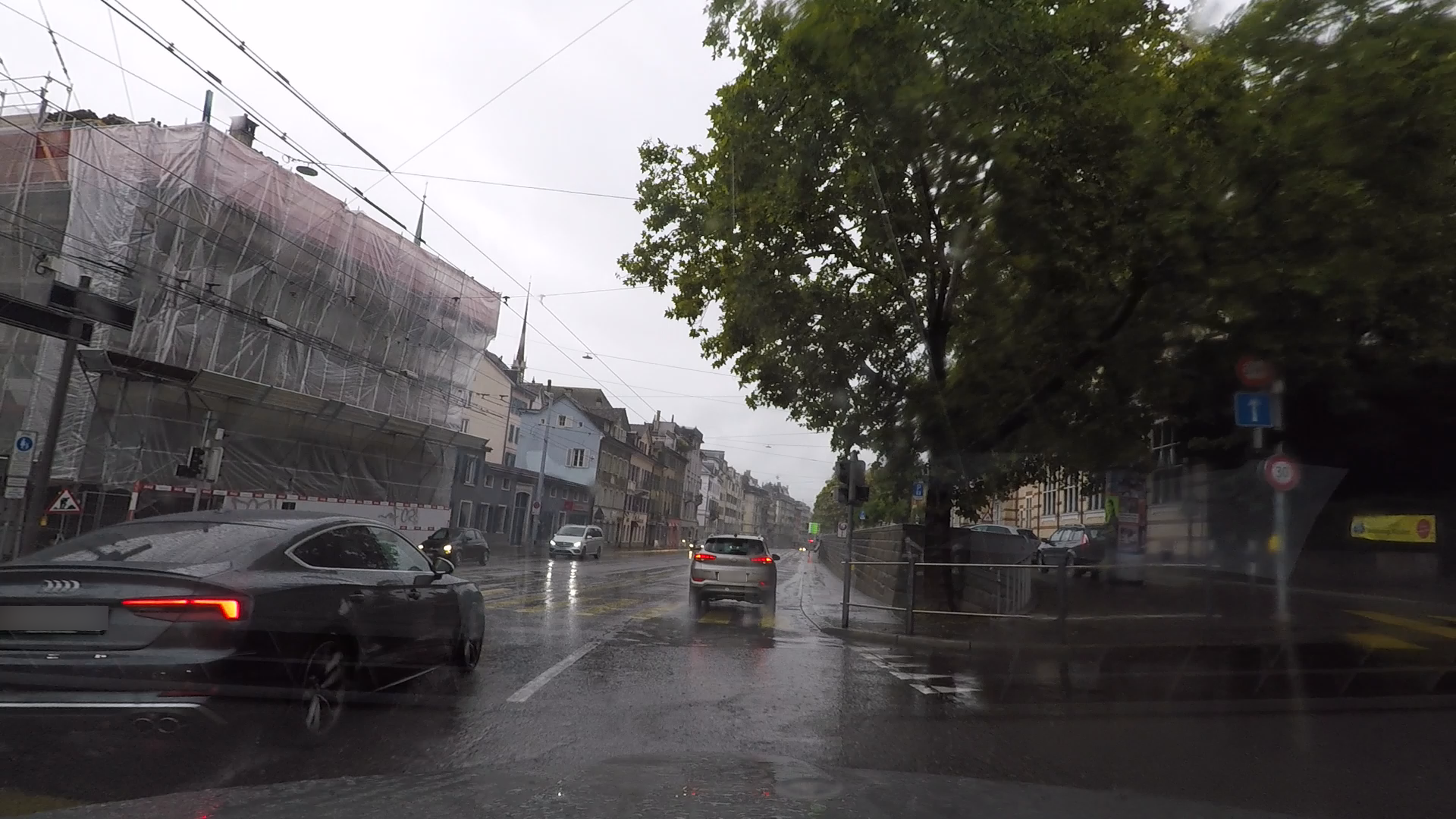}}
  \hfil
  \subfloat{\includegraphics[width=0.198\textwidth]{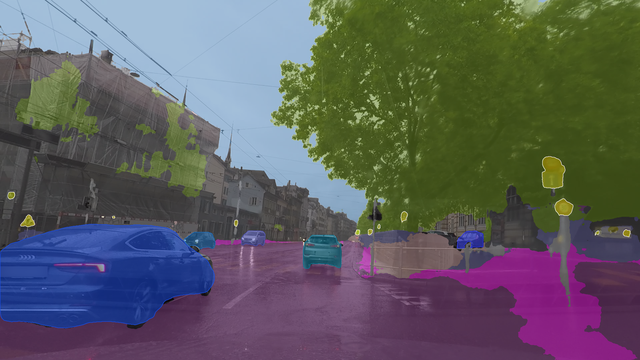}}
  \hfil
  \subfloat{\includegraphics[width=0.198\textwidth]{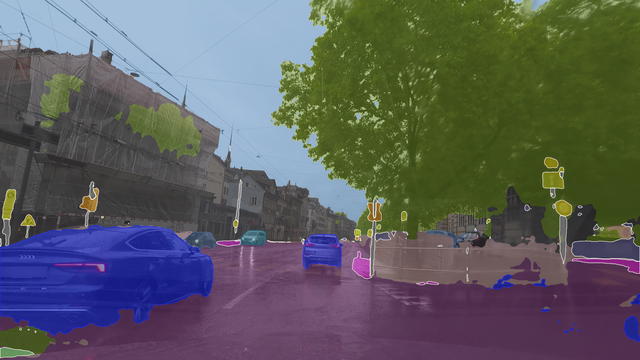}}
  \hfil
  \subfloat{\includegraphics[width=0.198\textwidth]{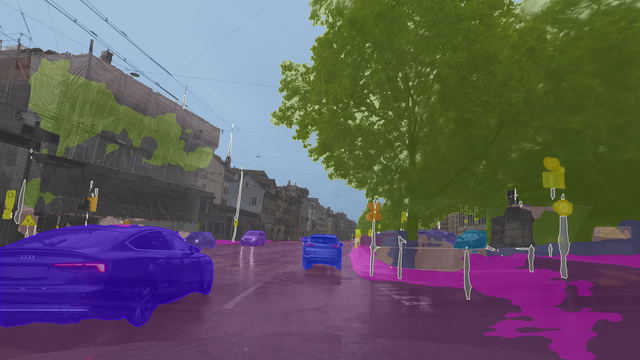}}
  \hfil
  \subfloat{\includegraphics[width=0.198\textwidth]{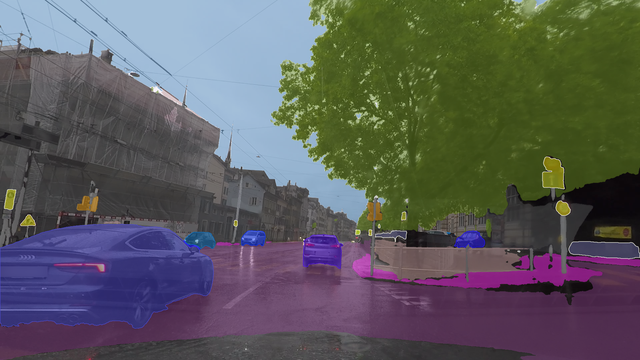}}
  \\
  \vspace{-0.35cm}
  \subfloat{\includegraphics[width=0.198\textwidth]{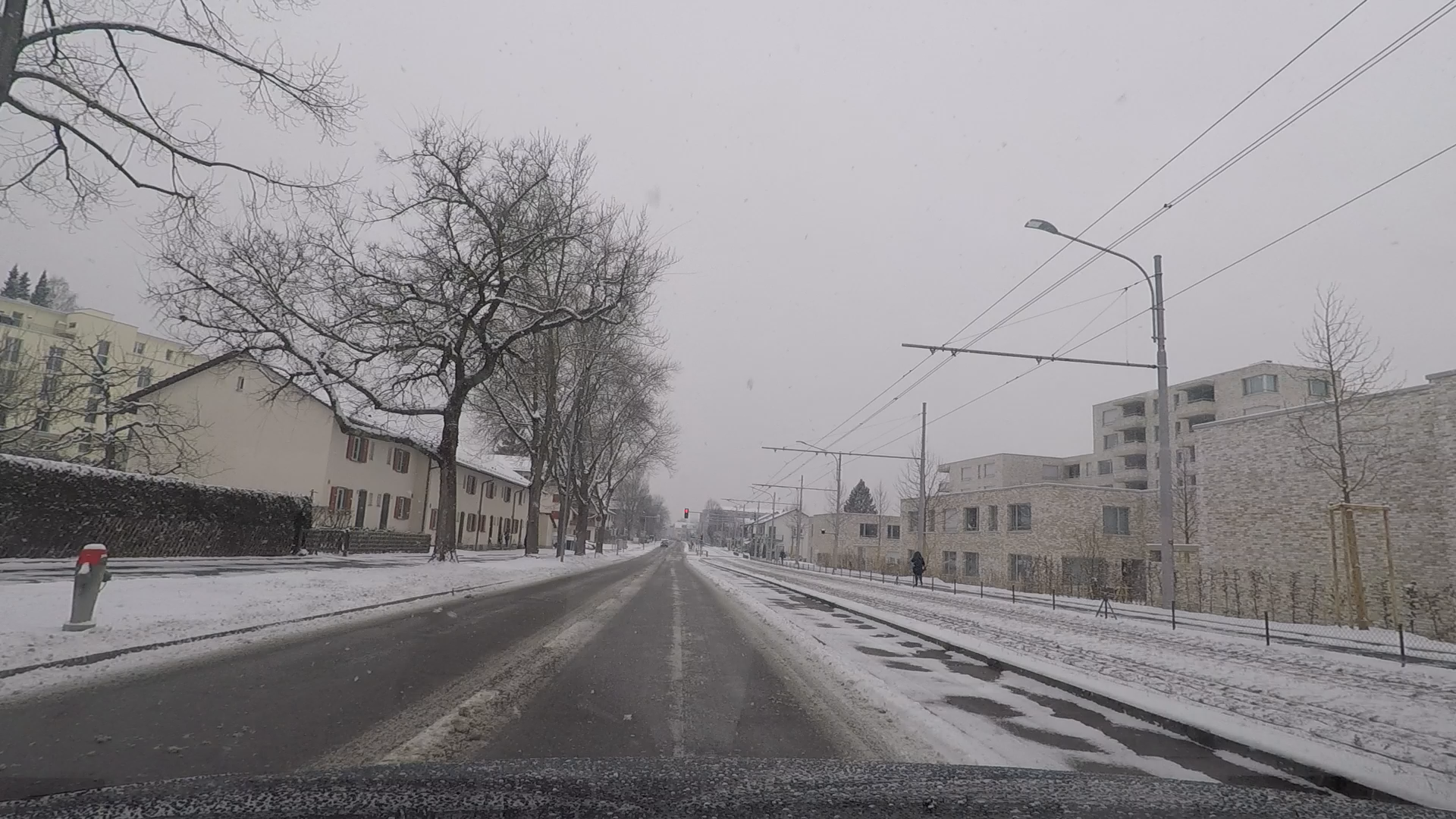}}
  \hfil
  \subfloat{\includegraphics[width=0.198\textwidth]{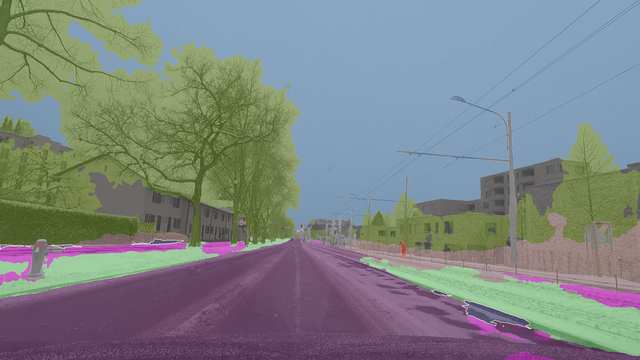}}
  \hfil
  \subfloat{\includegraphics[width=0.198\textwidth]{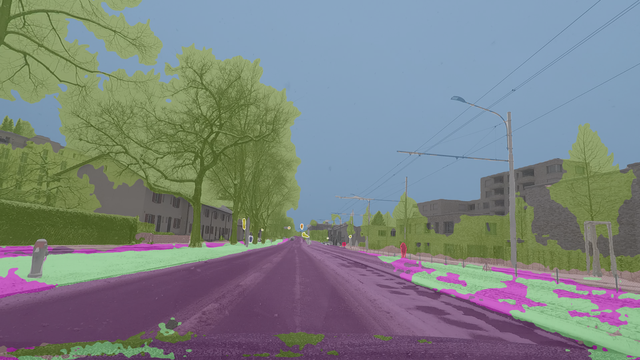}}
  \hfil
  \subfloat{\includegraphics[width=0.198\textwidth]{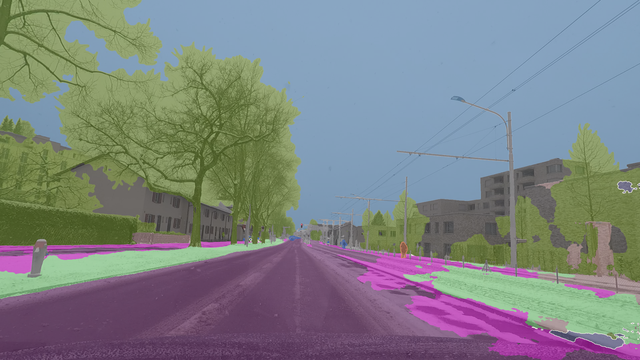}}
  \hfil
  \subfloat{\includegraphics[width=0.198\textwidth]{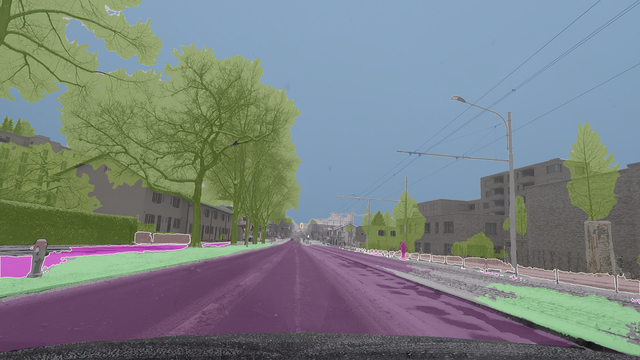}}
  \caption{\textbf{Qualitative results of panoptic segmentation methods on ACDC.} From left to right: image, Panoptic FPN~\cite{Kirillov_2019_CVPR_panoptic_fpn}, K-Net~\cite{NEURIPS2021_knet}, Panoptic-Deeplab~\cite{cheng2020panoptic_deeplab}, and Mask2Former~\cite{cheng2021mask2former}. The color coding of the semantic classes matches Fig.~\ref{fig:dataset:stats}.}
  \label{fig:panseg}
\end{figure*}

%% file: sections/7_external.tex
In this section, we evaluate on ACDC models that have been pre-trained on external datasets, for various semantic perception tasks.

\subsection{Semantic Segmentation}

\begin{table}[!tb]
  \caption{\textbf{Comparison of externally pre-trained semantic segmentation models on ACDC for individual conditions and jointly for all adverse conditions.} The three groups of rows present models pre-trained on normal, foggy, and nighttime conditions respectively. CS: Cityscapes~\cite{Cityscapes}, FC: Foggy Cityscapes~\cite{SFSU_synthetic}, FC-DBF: Foggy Cityscapes-DBF~\cite{SynRealDataFogECCV18}, FZ: Foggy Zurich~\cite{SynRealDataFogECCV18}, ND: Nighttime Driving~\cite{daytime:2:nighttime}, DZ: Dark Zurich~\cite{MGCDA_UIoU}.}
  \label{table:pretrained_seg}
  \centering
  \setlength\tabcolsep{2pt}
  \small
  \begin{tabular*}{\linewidth}{l @{\extracolsep{\fill}} cccccc}
  \toprule
  Method & Trained on & Fog & Night & Rain & Snow & All\\
  \midrule
  RefineNet~\cite{refinenet} & CS & 46.4 & 29.0 & 52.6 & 43.3 & 43.7\\
  DeepLabv2~\cite{DeepLab:v2} & CS & 33.5 & 30.1 & 44.5 & 40.2 & 38.0\\
  DeepLabv3+~\cite{DeepLab:v3+} & CS & 45.7 & 25.0 & 50.0 & 42.0 & 41.6\\
  DANet~\cite{DANet} & CS & 34.7 & 19.1 & 41.5 & 33.3 & 33.1\\
  HRNet~\cite{hrnet} & CS & 38.4 & 20.6 & 44.8 & 35.1 & 35.3\\
  \midrule
  SFSU~\cite{SFSU_synthetic} & FC & 45.6 & 29.5 & 51.6 & 41.4 & 42.9\\
  CMAda~\cite{SynRealDataFogECCV18} & FC-DBF+FZ & 51.2 & 32.0 & 53.4 & 47.6 & 47.1\\
  \midrule
  DMAda~\cite{daytime:2:nighttime} & ND & 50.7 & 32.7 & 54.9 & 48.9 & 47.9\\
  GCMA~\cite{GCMA_UIoU:v1} & CS+DZ & 52.4 & 42.9 & 58.0 & 53.8 & 53.4\\
  MGCDA~\cite{MGCDA_UIoU} & CS+DZ & 45.9 & 40.8 & 54.2 & 50.5 & 48.9\\
  DANNet~\cite{DANNet} & CS+DZ & -- & 47.6 & -- & -- & --\\
  \bottomrule
  \end{tabular*}
\end{table}

In Table~\ref{table:pretrained_seg}, we use ACDC to evaluate semantic segmentation models which have been pre-trained on external datasets. For models pre-trained on Cityscapes, the performance drop is larger on the nighttime set, implying that the domain shift from the normal-condition domain is larger for this set. Methods that specialize on fog or nighttime generally perform better on that condition compared to models pre-trained on Cityscapes. Moreover, most of these specialized methods also improve the performance on conditions other than the one encountered at training time.

\subsection{Instance Segmentation}

In Table~\ref{table:pretrained_inst}, we evaluate various models pre-trained on Cityscapes for instance segmentation. All these instance segmentation models exhibit rather low performance on ACDC, indicating that there is a large domain gap between Cityscapes and ACDC. Night is the most challenging condition for all these pre-trained models. Although HTC brings improvement in the supervised setting compared to Mask R-CNN and Cascaded Mask R-CNN as we have discussed in Table~\ref{table:supervised:instance_all}, it deteriorates the out-of-distribution performance in the present case of external pre-training. As HTC enhances the correlation of mask branches in different stages with interleaved execution, apparently the domain-specific bias is also strengthened and a worse out-of-distribution robustness is thus induced. As Detectors are built on top of HTC, they also exhibit a limited out-of-distribution performance on ACDC.

\begin{table}[!tb]
  \caption{\textbf{Comparison of externally pre-trained instance segmentation models on ACDC for individual conditions and jointly for all adverse conditions.} The two groups of rows present performance in $AP^{box}$ and $AP^{mask}$ respectively. CS: Cityscapes~\cite{Cityscapes}.}
  \label{table:pretrained_inst}
  \centering
  \setlength\tabcolsep{1pt}
  \small
  \begin{tabular*}{\linewidth}{l @{\extracolsep{\fill}} cccccc}
  \toprule
  Method & Trained on & Fog & Night & Rain & Snow & All\\
  \midrule
  Mask R-CNN~\cite{he2017maskrcnn}          & CS & 11.1 & 4.8 & 12.8 & 14.3 & 10.2 \\
  Cascaded Mask R-CNN~\cite{cai2019cascadercnn} & CS & 11.1 & 5.3 & 13.3 & 15.5 & 10.6  \\
  HTC~\cite{chen2019htc}                 & CS & 7.6  & 2.6 & 9.2  & 9.2  & 6.5 \\
  Detectors~\cite{qiao2020detectors}           & CS & 12.1 & 3.8 & 12.9 & 12.7 & 9.4 \\
  \midrule
  Mask R-CNN~\cite{he2017maskrcnn}          & CS & 9.8  & 3.6 & 11.0 & 12.9 & 8.8 \\
  Cascaded Mask R-CNN~\cite{cai2019cascadercnn} & CS & 9.8  & 3.9 & 11.8 & 12.5 & 9.0  \\
  HTC~\cite{chen2019htc}                 & CS & 7.0  & 2.3 & 8.4  & 8.3  & 5.9 \\
  Detectors~\cite{qiao2020detectors}           & CS & 10.1 & 2.6 & 10.9 & 10.1 & 7.8 \\
  \bottomrule
  \end{tabular*}
\end{table}

\subsection{Panoptic Segmentation}

In Table~\ref{table:pretrained_pan}, we report the performance of Cityscapes-pre-trained panoptic segmentation models on ACDC adverse-condition images. We also observe a performance drop caused by the domain shift between Cityscapes and ACDC on state-of-the-art panoptic segmentation models. Moreover, the transformer-based Mask2Former outperforms the convolutional methods PanopticFPN and Panoptic-Deeplab. Interestingly, we also find that both mask-based methods, including K-Net and Mask2Former, present a substantially better generalization ability than per-pixel classification based methods. This indicates the mask-based methods are less affected by the domain shift in adverse conditions. However, even though the transformer-based Mask2Former model pre-trained on normal conditions shows impressive robustness to adverse conditions, it still performs much worse than the Mask2Former model which has been specifically trained on adverse conditions. By introducing the panoptically annotated extension of ACDC, we hope that the latter will help to close this performance gap by fostering the development of both normal-to-adverse panoptic segmentation adaptation methods and better-generalizing supervised panoptic segmentation methods.

\begin{table}[!tb]
  \caption{\textbf{Comparison of externally pre-trained panoptic segmentation models on ACDC for individual conditions and jointly for all adverse conditions.} We report the performance in $PQ$ for different conditions. CS: Cityscapes~\cite{Cityscapes}.}
  \label{table:pretrained_pan}
  \centering
  \setlength\tabcolsep{2pt}
  \small
  \begin{tabular*}{\linewidth}{l @{\extracolsep{\fill}} cccccc}
  \toprule
  Method & Trained on & Fog & Night & Rain & Snow & All\\
  \midrule
  PanopticFPN~\cite{Kirillov_2019_CVPR_panoptic_fpn} & CS & 15.9 & 4.0 & 18.6 & 13.1 & 13.0 \\
  K-Net~\cite{NEURIPS2021_knet} & CS & 17.3 & 6.0 & 23.0 & 18.7 & 16.7 \\
  Panoptic-Deeplab~\cite{cheng2020panoptic_deeplab} & CS & 6.5 & 1.6 & 8.3 & 1.6 & 4.7 \\
  Mask2Former~\cite{cheng2021mask2former} & CS & 42.7 & 19.9 & 41.4 & 42.0 & 37.7 \\
  \bottomrule
  \end{tabular*}
\end{table}

\subsection{Analysis for Externally Pre-trained Models}
\label{sec:external:analysis}

Consistent with the observations in Sec.~\ref{sec:supervised:analysis}, transformer-based methods demonstrate stronger generalization than CNN-based methods.
In Sec.~\ref{sec:supervised:analysis}, we noted that certain CNN-based models can achieve substantial in-domain performance gains through carefully designed feature aggregation mechanisms.
However, in the present externally pre-trained setting, such designs may reduce out-of-distribution robustness.
For example, HTC underperforms its base model, Cascade Mask R-CNN, in Table~\ref{table:pretrained_inst}, and Panoptic-DeepLab falls behind its base PanopticFPN in Table~\ref{table:pretrained_pan} on out-of-distribution evaluations.

%% file: sections/8_uncertainty.tex
\begin{table}[!tb]
  \caption{\textbf{Uncertainty-aware semantic segmentation baseline results using AUIoU.} Supervised methods for standard semantic segmentation are trained and evaluated either separately on each condition or jointly on all adverse conditions for semantic label prediction. Confidence prediction baselines: globally constant and equal to $100\%$ (Constant $100\%$), max-softmax network outputs (Max-Softmax), ground-truth invalid masks (GT).}
  \label{table:uiou:baselines}
  \centering
  \setlength\tabcolsep{2pt}
  \small
  \begin{tabular*}{\linewidth}{l @{\extracolsep{\fill}} lccccc}
  \toprule
  Method & Confidence & Fog & Night & Rain & Snow & All\\
  \midrule
  RefineNet~\cite{refinenet} & Constant $100\%$ & 63.6 & 52.2 & 66.4 & 62.5 & 65.3\\
  RefineNet~\cite{refinenet} & Max-Softmax & 60.6 & 51.4 & 62.5 & 59.9 & 62.5\\
  RefineNet~\cite{refinenet} & GT & 67.9 & 61.1 & 67.9 & 64.0 & 68.8\\
  \midrule
  DeepLabv2~\cite{DeepLab:v2} & Constant $100\%$ & 52.2 & 45.4 & 57.6 & 56.8 & 55.3\\
  DeepLabv2~\cite{DeepLab:v2} & Max-Softmax & 51.9 & 45.9 & 56.0 & 56.8 & 54.7\\
  DeepLabv2~\cite{DeepLab:v2} & GT & 56.7 & 54.7 & 59.1 & 58.4 & 58.9\\
  \midrule
  DeepLabv3+~\cite{DeepLab:v3+} & Constant $100\%$ & 68.7 & 59.2 & 73.5 & 70.5 & 70.0\\
  DeepLabv3+~\cite{DeepLab:v3+} & Max-Softmax & 66.4 & 59.1 & 70.6 & 67.9 & 67.8\\
  DeepLabv3+~\cite{DeepLab:v3+} & GT & 73.1 & 67.1 & 75.0 & 72.0 & 73.3\\
  \bottomrule
  \end{tabular*}
\end{table}

Existing uncertainty-based works on semantic segmentation~\cite{simultaneous:segmentation:outliers,uncertainty:bayesian} are evaluated only with IoU, which does not assess the predicted confidence. In contrast, for uncertainty-aware semantic segmentation, algorithms are required to output both a hard semantic prediction $\hat{H}$ and a confidence map $C$ with values in the range $[0,\,1]$. The average UIoU (AUIoU) metric is computed by thresholding $C$ at multiple values across the range $[0,\,1]$, calculating the UIoU~\cite{MGCDA_UIoU} for each threshold and averaging the results. A pixel $p$ with confidence value below the threshold at hand is treated as invalid and contributes positively if $J(p) = 1$ (true invalid) and negatively if $J(p) = 0$ (false invalid).

\subsection{Baselines and Oracles}
\label{sec:uncertainsemseg:baselines}

We present the results of straightforward baselines for uncertainty-aware semantic segmentation that are based on methods for standard semantic segmentation in Table~\ref{table:uiou:baselines}. We first evaluate three such methods using constant confidence maps equal to 1, not modeling confidence. In this case, AUIoU reduces to IoU. Any sensible method that models confidence should improve upon this baseline. Using the max-softmax scores of these methods as confidence maps generally yields inferior results to globally constant confidence, as softmax is not a good proxy for confidence. An upper bound for the performance of the examined methods is obtained by using a confidence oracle. More specifically, we use the binary complement of the ground-truth invalid mask $J$ as the confidence prediction. This raises AUIoU performance significantly across all conditions compared to the globally constant confidence baseline. The performance gap between the oracle and the baseline is largest for night, indicating that explicitly modeling uncertainty has the potential to improve performance especially in the nighttime domain. We have also trained \cite{simultaneous:segmentation:outliers} on ACDC, using the GT invalid masks for training its outlier detection part. The learned confidence by \cite{simultaneous:segmentation:outliers} leads to lower test set AUIoU (52.0\%) than constant confidence (53.0\%), indicating that a better modeling of uncertainty is needed in future approaches.

%% file: sections/9_conclusion.tex
In this paper, we have presented ACDC, a large-scale dataset and benchmark suite for robust semantic driving scene perception. Our dataset covers adverse visual domains that are common in driving scenarios and features high-quality pixel-level panoptic annotations which also include visually degraded image regions, while the present extended version also includes normal-condition annotations, completing the condition span of the dataset. Our annotations support a wide range of five dense semantic perception tasks: standard and uncertainty-aware semantic segmentation, object detection, instance segmentation, and panoptic segmentation.

We have evaluated several state-of-the-art approaches on our benchmark, both in the supervised and the unsupervised setting. The conclusions from this evaluation show the importance of ACDC in steering future progress in the field: (i) ACDC provides a challenging, real-world target domain for unsupervised domain adaptation approaches to various semantic perception tasks both in the normal-to-adverse adaptation setting and in the sensor-level adaptation setting, (ii) ACDC is a hard benchmark for supervised semantic perception, as state-of-the-art methods generally score much lower on it than on standard normal-condition benchmarks such as Cityscapes, (iii) ACDC can be used jointly with existing normal-condition datasets for training in order to regularize models better and improve their performance both under normal and adverse conditions.

%% file: appendices.tex
\appendices

\section{Training Details}
\label{supp:sec:training}

We provide the detailed training configurations for the various methods for semantic segmentation, object detection, instance segmentation and panoptic segmentation that have been used in Sec.~\ref{sec:n2a_adapt}, \ref{sec:n2n_adapt}, and \ref{sec:supervised} and for the method in~\cite{simultaneous:segmentation:outliers} for uncertainty-aware semantic segmentation that has been used in Sec.~\ref{sec:uncertain}.

\begin{table}[!tb]
  \caption{\textbf{Training details for UDA semantic segmentation methods in Cityscapes$\to$ACDC adaptation.} ``SSL rounds'': number of training rounds that include supervision from pseudo-labels; if not relevant for a method, -- is reported. ``Training iterations'': number of SGD iterations for each training round (number of epochs for each training round is alternatively reported).}
  \label{table:supp:training:uda:all}
  \centering
  \setlength\tabcolsep{2pt}
  \small
  \begin{tabular}{lcc}
  \toprule
  Method & SSL rounds & Training iterations\\
  \midrule
  AdaptSegNet & -- & 95k\\
  ADVENT & -- & 80k\\
  BDL & 0 & 80k\\
  CLAN & -- & 90k\\
  CRST & 3 & 2 epochs\\
  FDA & 1 & 80k\\
  SIM & 1 & 80k\\
  MRNet & 1 & 50k\\
  \bottomrule
  \end{tabular}
\end{table}

\begin{table}[!tb]
  \caption{\textbf{Training details for UDA semantic segmentation methods in Cityscapes$\to$ACDC adaptation for individual conditions.} ``SSL rounds'': number of training rounds that include supervision from pseudo-labels; if not relevant for a method, -- is reported. ``Training iterations'': number of SGD iterations for each training round.}
  \label{table:supp:training:uda:individual}
  \centering
  \setlength\tabcolsep{2pt}
  \small
  \begin{tabular}{lcc}
  \toprule
  Method & SSL rounds & Training iterations\\
  \midrule
  AdaptSegNet & -- & 40k\\
  ADVENT & -- & 40k\\
  BDL & 0 & 40k\\
  CLAN & -- & 40k\\
  FDA & 1 & 40k\\
  SIM & 1 & 40k\\
  MRNet & 1 & 40k\\
  \bottomrule
  \end{tabular}
\end{table}

\begin{table}[!tb]
  \caption{\textbf{Training details for supervised semantic segmentation methods on ACDC.}}
  \label{table:supp:training:supervised}
  \centering
  \setlength\tabcolsep{2pt}
  \small
  \begin{tabular}{lcc}
  \toprule
  Method & Base LR & Training epochs\\
  \midrule
  RefineNet & $5\times{}10^{-5}$ & 60\\
  DeepLabv2 & $2.5\times{}10^{-4}$ & 60\\
  DeepLabv3+ & $10^{-4}$ & 60\\
  HRNet & $10^{-4}$ & 60\\
  \bottomrule
  \end{tabular}
\end{table}

\begin{table*}[!tb]
  \caption{\textbf{Comparison of state-of-the-art unsupervised domain adaptive semantic segmentation methods on Cityscapes$\to$ACDC adaptation for fog.} Performance of the model trained only on the source domain (Source model) and of the oracle with access to the target domain labels (Oracle) is also reported.}
  \label{table:supp:uda:fog}
  \centering
  \setlength\tabcolsep{2pt}
  \small
  \begin{tabular*}{\linewidth}{l @{\extracolsep{\fill}} cccccccccccccccccccc}
  \toprule
  Method & \ver{road} & \ver{sidew.} & \ver{build.} & \ver{wall} & \ver{fence} & \ver{pole} & \ver{light} & \ver{sign} & \ver{veget.} & \ver{terrain} & \ver{sky} & \ver{person} & \ver{rider} & \ver{car} & \ver{truck} & \ver{bus} & \ver{train} & \ver{motorc.} & \ver{bicycle} & mIoU\\
  \midrule
  Source model & 66.4 & 31.2 & 26.8 & 22.9 & 18.6 & 8.2 & 32.3 & 10.7 & 70.7 & 39.0 & 31.3 & 17.6 & 41.1 & 65.0 & 30.0 & 34.3 & 18.3 & 42.3 & 29.0 & 33.5\\
  \midrule
  AdaptSegNet & 35.4 & 45.9 & 35.4 & 25.6 & 17.5 & 9.0 & 32.5 & 23.1 & 70.5 & 47.4 & 11.6 & 22.3 & 28.2 & 44.4 & 43.9 & 35.0 & 46.0 & 15.6 & 15.0 & 31.8\\
  ADVENT & 44.2 & 38.9 & 26.4 & 20.7 & 20.1 & 7.9 & 34.4 & 23.6 & 70.7 & 35.6 & 8.3 & 17.3 & 43.5 & 60.0 & 48.6 & 46.8 & 40.5 & 19.9 & 17.6 & 32.9\\
  BDL & 36.9 & 37.8 & 47.0 & 28.2 & 21.6 & 13.7 & 37.2 & 34.5 & 67.2 & 49.4 & 27.6 & 29.1 & 51.3 & 58.5 & 49.4 & 51.8 & 30.3 & 21.4 & 22.5 & 37.7\\
  CLAN & 48.8 & 41.3 & 29.6 & 27.2 & 21.0 & 16.1 & 41.1 & 39.6 & 67.7 & 50.2 & 15.4 & 36.2 & 30.8 & 72.2 & 52.2 & 54.4 & 47.2 & 27.1 & 22.6 & 39.0\\
  FDA & 68.8 & 37.3 & 27.1 & 27.6 & 19.8 & 21.6 & 37.5 & 43.3 & 74.9 & 43.7 & 33.1 & 35.0 & 21.5 & 65.7 & 44.6 & 45.3 & 47.1 & 41.5 & 15.8 & 39.5\\
  SIM & 76.7 & 43.1 & 23.5 & 23.6 & 17.9 & 10.9 & 32.1 & 15.3 & 70.4 & 50.5 & 21.4 & 34.8 & 44.3 & 58.4 & 50.5 & 55.2 & 34.7 & 23.0 & 8.8 & 36.6\\
  MRNet & 78.6 & 26.1 & 19.6 & 29.0 & 13.5 & 12.0 & 41.9 & 49.0 & 78.2 & 59.0 & 6.6 & 39.8 & 26.1 & 72.5 & 44.8 & 37.9 & 59.6 & 19.1 & 24.1 & 38.8\\
  \midrule
  Oracle & 89.9 & 65.6 & 81.2 & 39.1 & 25.9 & 28.1 & 45.9 & 47.7 & 83.0 & 67.4 & 96.7 & 35.2 & 38.4 & 73.5 & 46.1 & 29.8 & 37.9 & 28.4 & 31.6 & 52.2\\
  \bottomrule
  \end{tabular*}
\end{table*}

\begin{table*}[!tb]
  \caption{\textbf{Comparison of state-of-the-art unsupervised domain adaptive semantic segmentation methods on Cityscapes$\to$ACDC adaptation for nighttime.} Performance of the model trained only on the source domain (Source model) and of the oracle with access to the target domain labels (Oracle) is also reported.}
  \label{table:supp:uda:night}
  \centering
  \setlength\tabcolsep{2pt}
  \small
  \begin{tabular*}{\linewidth}{l @{\extracolsep{\fill}} cccccccccccccccccccc}
  \toprule
  Method & \ver{road} & \ver{sidew.} & \ver{build.} & \ver{wall} & \ver{fence} & \ver{pole} & \ver{light} & \ver{sign} & \ver{veget.} & \ver{terrain} & \ver{sky} & \ver{person} & \ver{rider} & \ver{car} & \ver{truck} & \ver{bus} & \ver{train} & \ver{motorc.} & \ver{bicycle} & mIoU\\
  \midrule
  Source model & 77.0 & 22.9 & 56.3 & 13.5 & 9.2 & 23.8 & 22.9 & 25.6 & 41.4 & 16.1 & 2.9 & 44.1 & 17.5 & 64.1 & 11.9 & 34.5 & 42.4 & 22.6 & 22.7 & 30.1\\
  \midrule
  AdaptSegNet & 84.9 & 39.9 & 66.8 & 17.2 & 17.7 & 13.4 & 17.6 & 16.4 & 39.6 & 16.1 & 5.7 & 42.8 & 21.4 & 44.8 & 11.9 & 13.0 & 39.1 & 27.5 & 28.4 & 29.7\\
  ADVENT & 86.5 & 45.3 & 60.8 & 23.2 & 12.5 & 15.4 & 18.0 & 19.4 & 41.2 & 18.3 & 2.7 & 43.8 & 21.3 & 61.6 & 12.6 & 19.1 & 43.0 & 30.2 & 27.6 & 31.7\\
  BDL & 87.1 & 49.6 & 68.8 & 20.2 & 17.5 & 16.7 & 19.9 & 24.1 & 39.1 & 23.7 & 0.2 & 42.0 & 20.4 & 63.7 & 18.0 & 27.0 & 45.6 & 27.8 & 31.3 & 33.8\\
  CLAN & 82.3 & 28.8 & 65.9 & 15.1 & 9.3 & 22.1 & 16.1 & 26.5 & 39.2 & 23.4 & 0.4 & 45.9 & 25.4 & 63.6 & 9.5 & 24.2 & 39.8 & 31.5 & 31.1 & 31.6\\
  FDA & 82.7 & 39.4 & 57.0 & 14.7 & 7.6 & 26.1 & 37.8 & 30.5 & 53.2 & 14.0 & 15.3 & 48.0 & 28.8 & 62.6 & 26.6 & 47.5 & 51.5 & 27.0 & 35.0 & 37.1\\
  SIM & 87.0 & 48.4 & 42.1 & 6.3 & 8.3 & 15.8 & 8.4 & 17.6 & 21.7 & 22.8 & 0.1 & 39.3 & 22.1 & 60.3 & 8.7 & 18.2 & 42.3 & 30.1 & 32.9 & 28.0\\
  MRNet & 83.6 & 36.3 & 65.6 & 8.1 & 8.2 & 21.5 & 30.0 & 23.7 & 39.4 & 24.2 & 0.0 & 44.1 & 26.0 & 64.9 & 0.8 & 3.6 & 7.6 & 10.3 & 31.8 & 27.9\\
  \midrule
  Oracle & 90.5 & 63.7 & 78.0 & 30.0 & 29.6 & 32.9 & 37.0 & 41.2 & 61.9 & 25.2 & 75.3 & 47.9 & 23.4 & 69.5 & 2.7 & 15.4 & 60.3 & 39.7 & 37.9 & 45.4\\
  \bottomrule
  \end{tabular*}
\end{table*}

\begin{table*}[!tb]
  \caption{\textbf{Comparison of state-of-the-art unsupervised domain adaptive semantic segmentation methods on Cityscapes$\to$ACDC adaptation for rain.} Performance of the model trained only on the source domain (Source model) and of the oracle with access to the target domain labels (Oracle) is also reported.}
  \label{table:supp:uda:rain}
  \centering
  \setlength\tabcolsep{2pt}
  \small
  \begin{tabular*}{\linewidth}{l @{\extracolsep{\fill}} cccccccccccccccccccc}
  \toprule
  Method & \ver{road} & \ver{sidew.} & \ver{build.} & \ver{wall} & \ver{fence} & \ver{pole} & \ver{light} & \ver{sign} & \ver{veget.} & \ver{terrain} & \ver{sky} & \ver{person} & \ver{rider} & \ver{car} & \ver{truck} & \ver{bus} & \ver{train} & \ver{motorc.} & \ver{bicycle} & mIoU\\
  \midrule
  Source model & 71.2 & 26.7 & 73.8 & 20.8 & 27.1 & 29.9 & 39.3 & 44.4 & 87.3 & 25.2 & 82.0 & 42.0 & 14.3 & 76.2 & 36.3 & 26.6 & 49.8 & 30.3 & 42.2 & 44.5\\
  \midrule
  AdaptSegNet & 81.2 & 43.2 & 83.3 & 27.3 & 31.4 & 23.0 & 41.4 & 40.5 & 87.2 & 35.0 & 93.1 & 40.2 & 15.5 & 73.9 & 45.7 & 34.9 & 57.0 & 27.1 & 49.1 & 49.0\\
  ADVENT & 77.0 & 31.0 & 52.5 & 35.0 & 34.2 & 23.4 & 42.1 & 41.0 & 85.3 & 34.2 & 26.7 & 41.3 & 14.1 & 75.6 & 47.3 & 40.4 & 64.3 & 29.6 & 46.2 & 44.3\\
  BDL & 79.1 & 39.0 & 82.8 & 30.0 & 34.5 & 28.1 & 40.1 & 47.3 & 87.0 & 28.7 & 91.8 & 40.6 & 17.8 & 74.6 & 46.3 & 36.7 & 60.4 & 33.2 & 46.3 & 49.7\\
  CLAN & 77.5 & 40.0 & 46.8 & 24.9 & 30.3 & 28.1 & 37.7 & 48.3 & 83.8 & 37.0 & 6.6 & 45.7 & 17.4 & 79.7 & 43.7 & 42.9 & 63.7 & 35.0 & 46.1 & 44.0\\
  FDA & 76.6 & 45.0 & 82.9 & 37.0 & 35.6 & 34.8 & 49.8 & 52.0 & 88.7 & 37.8 & 88.8 & 43.6 & 17.4 & 76.8 & 46.5 & 53.6 & 64.8 & 34.5 & 45.5 & 53.3\\
  SIM & 76.6 & 29.6 & 85.7 & 20.4 & 28.7 & 21.3 & 37.4 & 34.2 & 87.3 & 34.8 & 94.0 & 29.4 & 16.6 & 73.2 & 46.1 & 22.3 & 46.2 & 21.8 & 39.3 & 44.5\\
  MRNet & 70.5 & 9.9 & 46.5 & 35.6 & 36.1 & 36.5 & 56.4 & 56.2 & 90.2 & 41.3 & 4.3 & 53.0 & 23.5 & 81.6 & 39.3 & 26.7 & 57.8 & 43.6 & 54.5 & 45.4\\
  \midrule
  Oracle & 87.3 & 63.9 & 89.0 & 50.3 & 40.6 & 38.4 & 52.2 & 53.4 & 89.2 & 42.2 & 96.7 & 51.5 & 13.0 & 81.9 & 47.9 & 47.2 & 72.2 & 29.1 & 48.8 & 57.6\\
  \bottomrule
  \end{tabular*}
\end{table*}

\begin{table*}[!tb]
  \caption{\textbf{Comparison of state-of-the-art unsupervised domain adaptive semantic segmentation methods on Cityscapes$\to$ACDC adaptation for snow.} Performance of the model trained only on the source domain (Source model) and of the oracle with access to the target domain labels (Oracle) is also reported.}
  \label{table:supp:uda:snow}
  \centering
  \setlength\tabcolsep{2pt}
  \small
  \begin{tabular*}{\linewidth}{l @{\extracolsep{\fill}} cccccccccccccccccccc}
  \toprule
  Method & \ver{road} & \ver{sidew.} & \ver{build.} & \ver{wall} & \ver{fence} & \ver{pole} & \ver{light} & \ver{sign} & \ver{veget.} & \ver{terrain} & \ver{sky} & \ver{person} & \ver{rider} & \ver{car} & \ver{truck} & \ver{bus} & \ver{train} & \ver{motorc.} & \ver{bicycle} & mIoU\\
  \midrule
  Source model & 68.5 & 26.6 & 52.7 & 18.8 & 26.9 & 22.2 & 35.7 & 40.7 & 76.5 & 3.6 & 49.9 & 50.4 & 27.1 & 73.7 & 27.6 & 39.1 & 60.9 & 21.1 & 42.5 & 40.2\\
  \midrule
  AdaptSegNet & 51.3 & 32.5 & 47.3 & 21.5 & 31.5 & 13.2 & 37.8 & 23.2 & 76.0 & 2.6 & 4.5 & 49.9 & 23.1 & 68.7 & 38.3 & 31.8 & 51.5 & 21.7 & 45.0 & 35.3\\
  ADVENT & 50.8 & 24.8 & 46.2 & 15.5 & 26.0 & 15.5 & 27.9 & 23.0 & 70.0 & 2.1 & 9.5 & 44.2 & 25.3 & 68.5 & 22.9 & 24.9 & 50.1 & 23.9 & 38.9 & 32.1\\
  BDL & 42.3 & 36.4 & 60.2 & 15.7 & 30.4 & 15.1 & 41.4 & 30.4 & 71.3 & 1.7 & 11.2 & 46.8 & 27.8 & 57.7 & 38.6 & 34.1 & 59.2 & 28.1 & 43.7 & 36.4\\
  CLAN & 71.8 & 26.0 & 37.3 & 12.5 & 27.0 & 21.1 & 32.0 & 41.1 & 78.5 & 1.9 & 0.9 & 50.9 & 23.9 & 82.4 & 43.2 & 39.5 & 61.6 & 25.2 & 39.4 & 37.7\\
  FDA & 74.6 & 30.9 & 56.1 & 20.5 & 34.8 & 28.7 & 53.9 & 47.8 & 80.5 & 1.1 & 55.9 & 53.1 & 37.9 & 79.7 & 40.5 & 51.9 & 67.4 & 34.3 & 41.8 & 46.9\\
  SIM & 72.1 & 26.7 & 39.4 & 13.3 & 29.5 & 15.3 & 26.4 & 17.9 & 76.4 & 4.8 & 5.1 & 45.9 & 32.0 & 76.2 & 29.8 & 26.6 & 48.3 & 23.2 & 24.2 & 33.3\\
  MRNet & 67.7 & 3.5 & 36.8 & 8.3 & 24.8 & 18.0 & 52.6 & 55.4 & 82.4 & 0.5 & 0.1 & 62.2 & 30.2 & 79.2 & 32.1 & 59.3 & 58.4 & 29.1 & 35.8 & 38.7\\
  \midrule
  Oracle & 89.1 & 61.7 & 82.7 & 26.4 & 40.9 & 35.5 & 56.5 & 54.1 & 85.2 & 39.0 & 95.1 & 55.0 & 25.7 & 84.3 & 38.6 & 53.8 & 77.6 & 29.0 & 49.5 & 56.8\\
  \bottomrule
  \end{tabular*}
\end{table*}

\begin{table*}[!tb]
  \caption{\textbf{Comparison of state-of-the-art unsupervised domain-adaptive object detection methods on Cityscapes$\to$ACDC for fog.} The first and second groups of rows present two-stage domain-adaptive detection and one-stage domain-adaptive detection methods, respectively. Performance of the model trained only on the source domain (Source model) and of the oracle with access to the target domain labels (Oracle) is also reported.}
  \label{table:supp:uda_det:fog}
  \centering
  \setlength\tabcolsep{2pt}
  \footnotesize
  \begin{tabular*}{\linewidth}{l @{\extracolsep{\fill}} cccccccccc}
  \toprule
  Method & \ver{person} & \ver{rider} & \ver{car} & \ver{truck} & \ver{bus} & \ver{train} & \ver{motorc.} & \ver{bicycle} & AP$_{50}^{box}$& AP$^{box}$\\
  \midrule
  Source model (Faster R-CNN) & 18.2 & 10.7 & 46.2 & 16.8 & 30.3 & 12.3 & 15.6 & 7.1 & 19.7 & 10.8  \\
  DA-Faster & 8.1 & 8.9 & 51.5 & 13.0 & 24.6 & 12.3 & 12.5 & 7.3 & 17.3 & 9.0 \\
  SADA & 23.3 & 3.9 & 60.8 & 11.7 & 24.9 & 8.2 & 16.6 & 6.7 & 19.5 & 10.0 \\
  MIC (SADA) &31.3 & 19.2 & 64.8 & 10.3 & 16.1 & 16.7 & 27.3 & 12.6 & 24.8 & 12.4 \\
  FRCNN-SIGMA++ & 19.1 & 14.4 & 54.8 & 16.7 & 33.1 & 22.6 & 16.6 & 8.4 & 23.2 & 12.2 \\
  Oracle & 27.5 & 13.1 & 58.5 & 29.8 & 41.0 & 26.6 & 22.7 & 12.1 & 28.9 & 16.4 \\
  \midrule
  Source model (FCOS) & 29.9 & 12.4 & 53.0 & 18.8 & 33.9 & 11.7 & 12.7 & 3.2 & 22.0 & 12.9 \\
  EPM   & 28.4 & 9.7 & 56.3 & 16.7 & 33.8 & 11.1 & 14.1 & 8.6 & 22.3  & 12.3 \\
  SIGMA & 32.1 & 16.7 & 59.2 & 17.9 & 25.1 & 17.7 & 27.3 & 7.0 & 25.4 & 14.2 \\
  Oracle & 30.4 & 12.2 & 64.8 & 26.7 & 32.0 & 23.6 & 29.4 & 9.5 & 28.6  & 16.9  \\
  \bottomrule
  \end{tabular*}
\end{table*}

\begin{table*}[!tb]
  \caption{\textbf{Comparison of state-of-the-art unsupervised domain-adaptive object detection methods on Cityscapes$\to$ACDC for nighttime.} The first and second groups of rows present two-stage domain-adaptive detection and one-stage domain-adaptive detection methods, respectively. Performance of the model trained only on the source domain (Source model) and of the oracle with access to the target domain labels (Oracle) is also reported.}
  \label{table:supp:uda_det:nighttime}
  \centering
  \setlength\tabcolsep{2pt}
  \footnotesize
  \begin{tabular*}{\linewidth}{l @{\extracolsep{\fill}} cccccccccc}
  \toprule
  Method & \ver{person} & \ver{rider} & \ver{car} & \ver{truck} & \ver{bus} & \ver{train} & \ver{motorc.} & \ver{bicycle} & AP$_{50}^{box}$& AP$^{box}$\\
  \midrule
  Source model (Faster R-CNN) & 19.0 & 17.0 & 27.3 & 3.2 & 28.3 & 8.1 & 3.4 & 8.8 & 14.4 & 7.2 \\
  DA-Faster & 15.1 & 14.5 & 20.1 & 2.1 & 13.3 & 8.4 & 3.8 & 15.2 & 11.6 & 5.3 \\
  SADA & 34.7 & 23.7 & 37.0 & 2.8 & 15.2 & 6.3 & 6.5 & 17.1 & 17.9 & 7.8 \\
  MIC (SADA) & 26.6 & 21.1 & 34.0 & 5.3 & 27.8 & 5.3 & 7.5 & 19.1 & 18.4  & 8.9 \\
  FRCNN-SIGMA++ & 24.5 & 24.0 & 41.7 & 10.1 & 40.4 & 16.9 & 6.6 & 21.4 & 23.2 & 11.1 \\
  Oracle &28.7 & 28.9 & 51.0 & 11.1 & 31.5 & 32.9 & 14.6 & 24.3 & 27.9 & 14.1 \\
  \midrule
  Source model (FCOS) & 23.5 & 15.9 & 25.9 & 2.5 & 26.8 & 6.7 & 5.5 & 8.8 & 14.4 & 7.2 \\
  EPM                 & 25.1 & 15.4 & 29.8 & 1.9 & 30.5 & 9.5 & 3.9 & 9.2 & 15.7  & 7.8 \\
  SIGMA               & 29.9 & 18.8 & 38.2 & 1.5 & 33.2 & 5.2 & 8.2 & 13.2 & 18.5 & 9.3  \\
  Oracle              & 39.0 & 30.2 & 54.2 & 3.6 & 39.4 & 28.9 & 15.2 & 19.1 & 28.7 & 15.1  \\
  \bottomrule
  \end{tabular*}
\end{table*}

\begin{table*}[!tb]
  \caption{\textbf{Comparison of state-of-the-art unsupervised domain-adaptive object detection methods on Cityscapes$\to$ACDC for rain.} The first and second groups of rows present two-stage domain-adaptive detection and one-stage domain-adaptive detection methods, respectively. Performance of the model trained only on the source domain (Source model) and of the oracle with access to the target domain labels (Oracle) is also reported.}
  \label{table:supp:uda_det:rain}
  \centering
  \setlength\tabcolsep{2pt}
  \footnotesize
  \begin{tabular*}{\linewidth}{l @{\extracolsep{\fill}} cccccccccc}
  \toprule
  Method & \ver{person} & \ver{rider} & \ver{car} & \ver{truck} & \ver{bus} & \ver{train} & \ver{motorc.} & \ver{bicycle} & AP$_{50}^{box}$& AP$^{box}$\\
  \midrule
  Source model (Faster R-CNN) & 23.1 & 8.1 & 66.2 & 29.6 & 2.9 & 20.4 & 25.1 & 15.5 & 23.9 & 11.2 \\
  DA-Faster & 19.5 & 8.3 & 64.1 & 24.5 & 4.2 & 16.7 & 22.1 & 14.0 & 21.7 & 9.7 \\
  SADA & 34.8 & 11.4 & 78.0 & 20.3 & 0.4 & 7.4 & 22.6 & 17.0 & 24.0 & 11.3  \\
  MIC (SADA) & 38.7 & 13.4 & 76.7 & 19.9 & 0.2 & 15.3 & 26.0 & 18.4 & 26.1 & 12.5 \\
  FRCNN-SIGMA++ & 30.3 & 7.9 & 69.0 & 36.2 & 1.0 & 29.3 & 28.5 & 17.2 & 27.4 & 12.7 \\
  Oracle & 36.7 & 12.5 & 73.8 & 49.0 & 12.6 & 37.4 & 37.1 & 28.1 & 35.9 & 17.8 \\
  \midrule
  Source model (FCOS) & 27.3 & 6.2 & 68.2 & 20.3 & 2.8 & 18.6 & 20.8 & 16.5 & 22.6 & 11.2 \\
  EPM                 & 29.3 & 9.3 & 65.8 & 17.1 & 1.5 & 16.6 & 19.6 & 15.8 & 21.9 & 10.6 \\
  SIGMA               & 28.0 & 5.3 & 72.3 & 25.1 & 1.7 & 26.2 & 16.5 & 20.1 & 24.4 & 12.1  \\
  Oracle              & 44.4 & 15.0 & 79.0 & 38.8 & 13.3 & 40.1 & 31.8 & 26.9 & 36.2 & 18.9 \\
  \bottomrule
  \end{tabular*}
\end{table*}

\begin{table*}[!tb]
  \caption{\textbf{Comparison of state-of-the-art unsupervised domain-adaptive object detection methods on Cityscapes$\to$ACDC for snow.} The first and second groups of rows present two-stage domain-adaptive detection and one-stage domain-adaptive detection methods, respectively. Performance of the model trained only on the source domain (Source model) and of the oracle with access to the target domain labels (Oracle) is also reported.}
  \label{table:supp:uda_det:snow}
  \centering
  \setlength\tabcolsep{2pt}
  \footnotesize
  \begin{tabular*}{\linewidth}{l @{\extracolsep{\fill}} cccccccccc}
  \toprule
  Method & \ver{person} & \ver{rider} & \ver{car} & \ver{truck} & \ver{bus} & \ver{train} & \ver{motorc.} & \ver{bicycle} & AP$_{50}^{box}$& AP$^{box}$\\
  \midrule
  Source model (Faster R-CNN) & 33.4 & 17.6 & 66.8 & 25.5 & 29.7 & 23.2 & 21.6 & 15.7 & 29.2 & 14.7 \\
  DA-Faster                   & 37.3 & 12.3 & 67.5 & 21.4 & 31.2 & 23.4 & 21.4 & 24.8 & 29.9 & 14.3 \\
  SADA                        & 48.1 & 20.2 & 74.6 & 7.2  & 7.2  & 11.5 & 23.8 & 32.6 & 28.2 & 12.3 \\
  MIC (SADA)                  & 46.3 & 30.1 & 76.4 & 8.1  & 19.3 & 19.9 & 23.9 & 28.3 & 31.5 & 15.9 \\
  FRCNN-SIGMA++               & 41.5 & 19.1 & 69.3 & 19.4 & 33.4 & 28.4 & 33.9 & 25.1 & 33.8 & 16.4 \\
  Oracle                      & 49.4 & 19.2 & 73.2 & 32.0 & 37.0 & 48.5 & 41.7 & 33.7 & 41.9 & 20.8 \\
  \midrule
  Source model (FCOS) & 40.9 & 18.3 & 68.1 & 23.3 & 24.4 & 18.6 & 19.3 & 14.3 & 28.4 & 15.2  \\
  EPM                 & 41.8 & 22.2 & 70.9 & 13.4 & 18.6 & 15.7 & 13.5 & 10.5 & 25.8 & 14.3  \\
  SIGMA               & 40.6 & 8.1  & 57.6 & 0.5  & 14.9 & 15.8 & 17.4 & 4.8  & 19.9 & 10.1  \\
  Oracle              & 56.6 & 22.8 & 76.2 & 36.4 & 30.5 & 38.6 & 26.0 & 26.2 & 39.2 & 21.5  \\
  \bottomrule
  \end{tabular*}
\end{table*}

\begin{table*}[!tb]
  \caption{\textbf{Comparison of state-of-the-art supervised semantic segmentation methods on ACDC for fog.} The first group of rows presents condition-specific expert models trained only on fog, while the second group presents uber models trained on all conditions.}
  \label{table:supp:supervised:fog}
  \centering
  \setlength\tabcolsep{2pt}
  \footnotesize
  \begin{tabular*}{\linewidth}{l @{\extracolsep{\fill}} cccccccccccccccccccc}
  \toprule
  Method & \ver{road} & \ver{sidew.} & \ver{build.} & \ver{wall} & \ver{fence} & \ver{pole} & \ver{light} & \ver{sign} & \ver{veget.} & \ver{terrain} & \ver{sky} & \ver{person} & \ver{rider} & \ver{car} & \ver{truck} & \ver{bus} & \ver{train} & \ver{motorc.} & \ver{bicycle} & mIoU\\
  \midrule
  RefineNet & 93.2 & 75.5 & 86.1 & 44.1 & 37.6 & 46.0 & 64.2 & 64.8 & 85.5 & 70.8 & 97.9 & 46.1 & 34.8 & 79.3 & 59.4 & 64.8 & 82.4 & 36.6 & 38.8 &          63.6\\
  DeepLabv2 & 89.9 & 65.6 & 81.2 & 39.1 & 25.9 & 28.1 & 45.9 & 47.7 & 83.0 & 67.4 & 96.7 & 35.2 & 38.4 & 73.5 & 46.1 & 29.8 & 37.9 & 28.4 & 31.6 &     52.2\\
  DeepLabv3+ & 93.8 & 77.4 & 88.8 & 51.0 & 43.3 & 54.2 & 68.2 & 71.7 & 87.7 & 74.6 & 98.2 & 53.5 & 32.1 & 83.8 & 69.3 & 84.4 & 85.3 & 47.2 & 40.1 &     68.7\\
  HRNet & 94.6 & 79.6 & 89.9 & 53.6 & 44.9 & 59.4 & 74.3 & 76.1 & 88.9 & 77.6 & 98.3 & 61.5 & 53.3 & 86.0 & 66.6 & 80.0 & 88.5 & 41.1 & 30.2 & 70.8\\
  \midrule
  RefineNet & 93.5 & 75.6 & 87.2 & 42.3 & 39.2 & 49.8 & 68.5 & 67.2 & 85.6 & 70.1 & 97.9 & 52.6 & 48.2 & 81.0 & 62.6 & 62.0 & 69.1 & 57.7 & 37.4 & 65.7\\
  DeepLabv2 & 90.9 & 67.2 & 81.6 & 38.7 & 29.5 & 29.7 & 51.2 & 50.7 & 81.4 & 61.9 & 96.0 & 34.8 & 40.5 & 74.1 & 53.4 & 53.1 & 59.9 & 8.3 & 32.5 & 54.5\\
  DeepLabv3+ & 93.6 & 77.6 & 89.2 & 54.0 & 44.8 & 55.8 & 67.6 & 72.0 & 88.0 & 73.5 & 98.2 & 49.5 & 24.4 & 83.9 & 72.2 & 84.2 & 89.2 & 52.8 & 42.4 & 69.1\\
  HRNet & 94.9 & 81.0 & 90.5 & 58.9 & 53.7 & 61.9 & 79.0 & 78.7 & 89.3 & 78.7 & 98.3 & 63.2 & 54.6 & 87.2 & 72.3 & 87.8 & 90.6 & 58.7 & 38.9 & 74.7\\
  \bottomrule
  \end{tabular*}
\end{table*}

\begin{table*}[!tb]
  \caption{\textbf{Comparison of state-of-the-art supervised semantic segmentation methods on ACDC for nighttime.} The first group of rows presents condition-specific expert models trained only on nighttime, while the second group presents uber models trained on all conditions.}
  \label{table:supp:supervised:night}
  \centering
  \setlength\tabcolsep{2pt}
  \footnotesize
  \begin{tabular*}{\linewidth}{l @{\extracolsep{\fill}} cccccccccccccccccccc}
  \toprule
  Method & \ver{road} & \ver{sidew.} & \ver{build.} & \ver{wall} & \ver{fence} & \ver{pole} & \ver{light} & \ver{sign} & \ver{veget.} & \ver{terrain} & \ver{sky} & \ver{person} & \ver{rider} & \ver{car} & \ver{truck} & \ver{bus} & \ver{train} & \ver{motorc.} & \ver{bicycle} & mIoU\\
  \midrule
  RefineNet & 93.4 & 70.3 & 78.6 & 34.3 & 34.1 & 46.9 & 52.2 & 54.2 & 66.3 & 18.7 & 78.1 & 60.3 & 35.5 & 76.2 & 4.7 & 47.8 & 59.4 & 36.0 & 45.3 &          52.2\\
  DeepLabv2 & 90.5 & 63.7 & 78.0 & 30.0 & 29.6 & 32.9 & 37.0 & 41.2 & 61.9 & 25.2 & 75.3 & 47.9 & 23.4 & 69.5 & 2.7 & 15.4 & 60.3 & 39.7 & 37.9 &     45.4\\
  DeepLabv3+ & 94.7 & 75.9 & 85.0 & 48.4 & 38.6 & 52.2 & 55.8 & 54.4 & 76.1 & 30.3 & 84.2 & 67.4 & 41.1 & 85.0 & 8.3 & 62.3 & 80.6 & 35.6 & 49.8 &     59.2\\
  HRNet & 95.5 & 78.8 & 86.5 & 49.2 & 44.1 & 58.0 & 64.5 & 63.2 & 75.6 & 41.0 & 83.9 & 71.7 & 48.8 & 84.6 & 15.5 & 76.9 & 81.2 & 25.9 & 55.9 & 63.2\\
  \midrule
  RefineNet & 93.5 & 70.9 & 80.3 & 32.0 & 32.0 & 46.0 & 53.9 & 54.1 & 69.2 & 31.9 & 78.0 & 61.0 & 35.4 & 80.2 & 11.6 & 60.0 & 69.4 & 48.9 & 46.8 & 55.5\\
  DeepLabv2 & 86.6 & 57.8 & 71.7 & 30.3 & 23.6 & 31.8 & 37.4 & 38.9 & 60.0 & 26.8 & 72.8 & 47.6 & 25.1 & 71.1 & 16.9 & 27.8 & 65.1 & 30.6 & 38.5 & 45.3\\
  DeepLabv3+ & 94.7 & 75.3 & 84.9 & 46.9 & 37.8 & 53.8 & 57.3 & 52.1 & 75.7 & 41.2 & 82.9 & 66.6 & 40.2 & 83.6 & 24.7 & 67.9 & 80.8 & 41.7 & 49.4 & 60.9\\
  HRNet & 95.7 & 79.0 & 86.2 & 46.8 & 43.5 & 59.2 & 64.9 & 64.5 & 75.3 & 40.3 & 82.7 & 72.1 & 52.6 & 86.9 & 18.8 & 78.8 & 83.6 & 52.5 & 57.3 & 65.3\\
  \bottomrule
  \end{tabular*}
\end{table*}

\begin{table*}[!tb]
  \caption{\textbf{Comparison of state-of-the-art supervised semantic segmentation methods on ACDC for rain.} The first group of rows presents condition-specific expert models trained only on rain, while the second group presents uber models trained on all conditions.}
  \label{table:supp:supervised:rain}
  \centering
  \setlength\tabcolsep{2pt}
  \footnotesize
  \begin{tabular*}{\linewidth}{l @{\extracolsep{\fill}} cccccccccccccccccccc}
  \toprule
  Method & \ver{road} & \ver{sidew.} & \ver{build.} & \ver{wall} & \ver{fence} & \ver{pole} & \ver{light} & \ver{sign} & \ver{veget.} & \ver{terrain} & \ver{sky} & \ver{person} & \ver{rider} & \ver{car} & \ver{truck} & \ver{bus} & \ver{train} & \ver{motorc.} & \ver{bicycle} & mIoU\\
  \midrule
  RefineNet & 89.2 & 69.8 & 91.7 & 52.2 & 51.3 & 57.9 & 71.0 & 69.9 & 93.6 & 50.5 & 98.4 & 65.8 & 25.1 & 88.1 & 49.4 & 55.4 & 74.8 & 47.0 & 60.2 &        66.4\\
  DeepLabv2 & 87.3 & 63.9 & 89.0 & 50.3 & 40.6 & 38.4 & 52.2 & 53.4 & 89.2 & 42.2 & 96.7 & 51.5 & 13.0 & 81.9 & 47.9 & 47.2 & 72.2 & 29.1 & 48.8 &   57.6\\
  DeepLabv3+ & 92.8 & 77.4 & 93.9 & 67.3 & 58.1 & 64.1 & 74.4 & 75.9 & 94.2 & 50.8 & 98.6 & 70.8 & 33.4 & 90.4 & 67.7 & 79.2 & 86.8 & 54.6 & 66.1 &   73.5\\
  HRNet & 94.8 & 81.8 & 94.9 & 69.6 & 63.7 & 69.5 & 79.6 & 80.7 & 94.8 & 51.2 & 98.7 & 73.5 & 27.0 & 93.1 & 75.4 & 40.9 & 61.4 & 59.6 & 70.8 & 72.7\\
  \midrule
  RefineNet & 91.5 & 73.5 & 91.1 & 51.0 & 51.6 & 58.3 & 72.5 & 73.7 & 92.9 & 51.2 & 97.9 & 65.5 & 29.5 & 89.2 & 59.8 & 68.2 & 80.3 & 48.0 & 59.5 & 68.7\\
  DeepLabv2 & 87.4 & 64.8 & 88.1 & 48.2 & 40.4 & 38.4 & 52.0 & 56.9 & 89.3 & 40.2 & 96.5 & 52.3 & 17.4 & 83.9 & 55.5 & 63.0 & 75.8 & 28.9 & 47.2 & 59.3\\
  DeepLabv3+ & 92.7 & 76.5 & 93.5 & 64.8 & 58.0 & 63.8 & 75.8 & 77.3 & 94.1 & 50.0 & 98.0 & 70.5 & 33.1 & 91.2 & 75.9 & 85.1 & 86.2 & 55.8 & 65.0 & 74.1\\
  HRNet & 95.6 & 83.1 & 94.2 & 60.1 & 66.3 & 71.2 & 82.3 & 82.4 & 94.6 & 55.1 & 98.6 & 75.2 & 39.7 & 93.4 & 73.8 & 86.2 & 85.9 & 66.4 & 71.3 & 77.7\\
  \bottomrule
  \end{tabular*}
\end{table*}

\begin{table*}[!tb]
  \caption{\textbf{Comparison of state-of-the-art supervised semantic segmentation methods on ACDC for snow.} The first group of rows presents condition-specific expert models trained only on snow, while the second group presents uber models trained on all conditions.}
  \label{table:supp:supervised:snow}
  \centering
  \setlength\tabcolsep{2pt}
  \footnotesize
  \begin{tabular*}{\linewidth}{l @{\extracolsep{\fill}} cccccccccccccccccccc}
  \toprule
  Method & \ver{road} & \ver{sidew.} & \ver{build.} & \ver{wall} & \ver{fence} & \ver{pole} & \ver{light} & \ver{sign} & \ver{veget.} & \ver{terrain} & \ver{sky} & \ver{person} & \ver{rider} & \ver{car} & \ver{truck} & \ver{bus} & \ver{train} & \ver{motorc.} & \ver{bicycle} & mIoU\\
  \midrule
  RefineNet & 90.1 & 65.7 & 86.4 & 31.2 & 48.1 & 58.0 & 76.7 & 70.3 & 89.7 & 45.7 & 97.3 & 70.8 & 15.4 & 87.1 & 35.0 & 43.1 & 79.1 & 38.7 & 59.9 &        62.5\\
  DeepLabv2 & 89.1 & 61.7 & 82.7 & 26.4 & 40.9 & 35.5 & 56.5 & 54.1 & 85.2 & 39.0 & 95.1 & 55.0 & 25.7 & 84.3 & 38.6 & 53.8 & 77.6 & 29.0 & 49.5 &   56.8\\
  DeepLabv3+ & 91.9 & 70.9 & 90.1 & 48.9 & 52.0 & 62.2 & 79.2 & 74.5 & 92.0 & 47.0 & 97.6 & 78.2 & 35.9 & 90.4 & 61.7 & 64.3 & 89.2 & 43.9 & 69.4 &   70.5\\
  HRNet & 93.6 & 75.2 & 89.0 & 42.0 & 55.6 & 67.7 & 83.3 & 78.9 & 93.0 & 48.9 & 97.8 & 78.1 & 16.4 & 92.6 & 54.8 & 61.6 & 87.0 & 50.0 & 68.9 & 70.2\\
  \midrule
  RefineNet & 90.2 & 65.7 & 86.5 & 33.7 & 50.6 & 57.8 & 78.0 & 71.5 & 89.2 & 44.5 & 97.0 & 73.8 & 46.0 & 88.4 & 50.0 & 48.0 & 79.9 & 40.6 & 60.3 & 65.9\\
  DeepLabv2 & 88.7 & 62.5 & 82.5 & 35.3 & 41.7 & 35.0 & 59.0 & 52.8 & 84.4 & 36.0 & 95.2 & 58.1 & 29.8 & 84.8 & 48.9 & 30.9 & 77.9 & 32.9 & 48.4 & 57.1\\
  DeepLabv3+ & 91.4 & 69.6 & 88.8 & 48.8 & 53.9 & 60.6 & 79.5 & 72.9 & 90.5 & 44.7 & 97.4 & 77.4 & 37.2 & 90.0 & 64.3 & 55.0 & 87.8 & 41.7 & 70.0 & 69.6\\
  HRNet & 94.4 & 77.3 & 91.5 & 53.1 & 63.6 & 70.2 & 85.1 & 81.4 & 92.1 & 57.7 & 97.7 & 83.3 & 69.6 & 93.6 & 71.8 & 54.5 & 86.3 & 52.7 & 73.1 & 76.3\\
  \bottomrule
  \end{tabular*}
\end{table*}

\begin{table*}[!tb]
  \caption{\textbf{Comparison of state-of-the-art supervised instance segmentation methods on ACDC for fog.} The first group of rows presents condition-specific expert models trained only on fog, while the second group presents uber models trained on all conditions. For each condition we report the performance in $AP^{mask}$.}
  \label{table:supp:supervised:inst_fog}
  \centering
  \setlength\tabcolsep{2pt}
  \footnotesize
  \begin{tabular*}{\linewidth}{l @{\extracolsep{\fill}} cccccccccc}
  \toprule
  Method & \ver{person} & \ver{rider} & \ver{car} & \ver{truck} & \ver{bus} & \ver{train} & \ver{motorc.} & \ver{bicycle} & $AP^{mask}$ \\
  \midrule
  Mask R-CNN          & 14.7 & 1.5 & 41.3 & 17.5 & 21.3 & 17.3 & 8.5 & 2.8 & 15.6 \\
  Cascaded Mask R-CNN & 15.5 & 0.8 & 42.3 & 21.7 & 23.6 & 13.2 & 10.3 & 2.4 & 16.2 \\
  HTC                 & 17.4 & 1.3 & 43.9 & 21.8 & 28.1 & 14.7 & 8.0 & 3.1 & 17.3 \\
  Detectors           & 16.2 & 1.4 & 44.0 & 22.0 & 25.9 & 20.0 & 6.8 & 2.6 & 17.4 \\
  \midrule
  Mask R-CNN          & 22.7 & 9.8 & 46.8 & 23.8 & 31.3 & 33.5 & 20.6 & 7.1 & 24.4 \\
  Cascaded Mask R-CNN & 22.6 & 9.7 & 47.7 & 25.1 & 33.9 & 31.9 & 15.5 & 8.0 & 24.3 \\
  HTC                 & 26.6 & 9.3 & 49.4 & 27.3 & 35.8 & 33.9 & 18.4 & 7.1 & 26.0 \\
  Detectors           & 23.8 & 8.0 & 49.3 & 26.8 & 35.1 & 37.6 & 15.4 & 6.3 & 25.3 \\
  \bottomrule
  \end{tabular*}
\end{table*}

\begin{table*}[!tb]
  \caption{\textbf{Comparison of state-of-the-art supervised instance segmentation methods on ACDC for nighttime.} The first group of rows presents condition-specific expert models trained only on nighttime, while the second group presents uber models trained on all conditions. For each condition we report the performance in $AP^{mask}$.}
  \label{table:supp:supervised:inst_night}
  \centering
  \setlength\tabcolsep{2pt}
  \footnotesize
  \begin{tabular*}{\linewidth}{l @{\extracolsep{\fill}} cccccccccc}
  \toprule
  Method & \ver{person} & \ver{rider} & \ver{car} & \ver{truck} & \ver{bus} & \ver{train} & \ver{motorc.} & \ver{bicycle} & $AP^{mask}$ \\
  \midrule
  Mask R-CNN          & 13.7 & 3.4 & 36.6 & 2.2 & 8.1 & 14.4 & 2.9 & 3.9 & 10.7 \\
  Cascaded Mask R-CNN & 13.8 & 3.4 & 36.9 & 2.2 & 8.7 & 17.8 & 4.8 & 4.2 & 11.5 \\
  HTC                 & 14.9 & 4.7 & 39.1 & 2.5 & 10.6 & 17.5 & 5.3 & 4.5 & 12.4 \\
  Detectors           & 15.1 & 3.8 & 39.4 & 5.5 & 12.6 & 18.3 & 5.9 & 4.3 & 13.1 \\
  \midrule
  Mask R-CNN          & 16.9 & 4.9 & 40.7 & 8.3 & 9.5 & 21.1 & 5.8 & 6.3 & 14.2 \\
  Cascaded Mask R-CNN & 17.1 & 4.8 & 41.6 & 3.5 & 9.4 & 22.7 & 5.6 & 6.3 & 13.9 \\
  HTC                 & 18.6 & 6.8 & 43.0 & 2.2 & 15.7 & 23.3 & 6.6 & 7.3 & 15.4 \\
  Detectors           & 19.3 & 6.7 & 42.5 & 5.7 & 15.9 & 27.6 & 6.0 & 8.0 & 16.5 \\
  \bottomrule
  \end{tabular*}
\end{table*}

\begin{table*}[!tb]
  \caption{\textbf{Comparison of state-of-the-art supervised instance segmentation methods on ACDC for rain.} The first group of rows presents condition-specific expert models trained only on rain, while the second group presents uber models trained on all conditions. For each condition we report the performance in $AP^{mask}$.}
  \label{table:supp:supervised:inst_raing}
  \centering
  \setlength\tabcolsep{2pt}
  \footnotesize
  \begin{tabular*}{\linewidth}{l @{\extracolsep{\fill}} cccccccccc}
  \toprule
  Method & \ver{person} & \ver{rider} & \ver{car} & \ver{truck} & \ver{bus} & \ver{train} & \ver{motorc.} & \ver{bicycle} & $AP^{mask}$ \\
  \midrule
  Mask R-CNN          & 20.7 & 1.4 & 56.1 & 26.1 & 20.9 & 27.9 & 9.8 & 7.7 & 21.3 \\
  Cascaded Mask R-CNN & 20.1 & 1.0 & 56.6 & 24.3 & 21.0 & 28.0 & 11.2 & 7.0 & 21.2 \\
  HTC                 & 22.2 & 1.0 & 58.9 & 25.2 & 19.7 & 30.5 & 11.1 & 9.3 & 22.3 \\
  Detectors           & 21.0 & 3.4 & 59.1 & 26.4 & 25.4 & 31.5 & 10.5 & 9.0 & 23.3 \\
  \midrule
  Mask R-CNN          & 20.2 & 1.4 & 57.1 & 27.1 & 20.7 & 27.1 & 10.7 & 8.4 & 21.6 \\
  Cascaded Mask R-CNN & 20.4 & 1.4 & 58.0 & 26.9 & 24.5 & 29.3 & 11.4 & 8.2 & 22.5 \\
  HTC                 & 22.6 & 2.5 & 60.3 & 25.0 & 22.7 & 32.1 & 11.0 & 9.7 & 23.2 \\
  Detectors           & 23.2 & 3.0 & 60.6 & 30.5 & 26.1 & 32.7 & 12.7 & 10.7 & 24.9 \\
  \bottomrule
  \end{tabular*}
\end{table*}

\begin{table*}[!tb]
  \caption{\textbf{Comparison of state-of-the-art supervised instance segmentation methods on ACDC for snow.} The first group of rows presents condition-specific expert models trained only on snow, while the second group presents uber models trained on all conditions. For each condition we report the performance in $AP^{mask}$.}
  \label{table:supp:supervised:inst_snow}
  \centering
  \setlength\tabcolsep{2pt}
  \footnotesize
  \begin{tabular*}{\linewidth}{l @{\extracolsep{\fill}} cccccccccc}
  \toprule
  Method & \ver{person} & \ver{rider} & \ver{car} & \ver{truck} & \ver{bus} & \ver{train} & \ver{motorc.} & \ver{bicycle} & $AP^{mask}$ \\
  \midrule
  Mask R-CNN          & 28.6 & 5.1 & 52.9 & 17.7 & 19.0 & 21.5 & 17.1 & 4.5 & 20.8 \\
  Cascaded Mask R-CNN & 28.8 & 5.9 & 52.6 & 21.3 & 28.3 & 26.5 & 9.0 & 5.8 & 22.3 \\
  HTC                 & 29.8 & 5.3 & 55.0 & 21.2 & 28.5 & 28.0 & 13.0 & 6.2 & 23.4 \\
  Detectors           & 29.2 & 5.7 & 55.5 & 23.1 & 29.3 & 26.7 & 12.2 & 5.8 & 23.4 \\
  \midrule
  Mask R-CNN          & 30.0 & 7.3 & 58.4 & 27.2 & 37.3 & 30.4 & 18.2 & 10.1 & 27.4 \\
  Cascaded Mask R-CNN & 30.5 & 10.3 & 59.5 & 27.2 & 40.1 & 30.8 & 17.0 & 10.1 & 28.2 \\
  HTC                 & 33.0 & 10.1 & 61.9 & 32.2 & 40.1 & 35.5 & 17.9 & 11.2 & 30.2 \\
  Detectors           & 33.8 & 11.7 & 61.2 & 28.9 & 37.3 & 37.9 & 17.9 & 9.5 & 29.8 \\
  \bottomrule
  \end{tabular*}
\end{table*}

\begin{table}[!tb]
  \caption{\textbf{Comparison of state-of-the-art supervised panoptic segmentation methods on ACDC for fog.} The first group of rows presents condition-specific expert models trained only on fog, while the second group presents uber models trained on all conditions.}
  \label{table:supp:supervised:panoptic_fog}
  \centering
  \setlength\tabcolsep{2pt}
  \small
  \begin{tabular*}{\linewidth}{l @{\extracolsep{\fill}} ccccc}
  \toprule
  Method           & PQ & PQ$^{things}$ & PQ$^{stuff}$ & SQ & RQ\\
  \midrule
  PanopticFPN      & 38.4 & 25.2 & 48.0 & 72.8 & 47.3 \\
  K-Net            & 37.9 & 16.1 & 53.8 & 68.8 & 47.1 \\
  Panoptic-Deeplab & 42.4 & 23.9 & 55.8 & 79.9 & 51.2 \\
  Mask2Former      & 44.9 & 23.8 & 60.3 & 79.0 & 54.5 \\
  \midrule
  PanopticFPN      & 43.9  & 33.3 & 51.6 & 79.0 & 53.4 \\
  K-Net            & 47.8  & 32.3 & 59.1 & 78.9 & 59.1 \\
  Panoptic-Deeplab & 49.1  & 33.8 & 60.1 & 80.1 & 58.9 \\
  Mask2Former      & 52.9  & 37.0 & 64.5 & 82.0 & 63.2 \\
  \bottomrule
  \end{tabular*}
\end{table}

\begin{table}[!tb]
  \caption{\textbf{Comparison of state-of-the-art supervised panoptic segmentation methods on ACDC for nighttime.} The first group of rows presents condition-specific expert models trained only on nighttime, while the second group presents uber models trained on all conditions.}
  \label{table:supp:supervised:panoptic_night}
  \centering
  \setlength\tabcolsep{2pt}
  \small
  \begin{tabular*}{\linewidth}{l @{\extracolsep{\fill}} ccccc}
  \toprule
  Method           & PQ & PQ$^{things}$ & PQ$^{stuff}$ & SQ & RQ\\
  \midrule
  PanopticFPN      & 29.8 & 22.0 & 35.4 & 67.4 & 39.5 \\
  K-Net            & 30.7 & 15.6 & 41.7 & 67.3 & 41.0 \\
  Panoptic-Deeplab & 34.1 & 20.2 & 44.3 & 68.9 & 44.3 \\
  Mask2Former      & 34.0 & 18.0 & 45.7 & 69.5 & 44.1 \\
  \midrule
  PanopticFPN      & 32.6 & 26.6 & 37.0 & 73.4 & 42.9 \\
  K-Net            & 33.4 & 18.3 & 44.4 & 70.6 & 44.7 \\
  Panoptic-Deeplab & 37.2 & 22.9 & 47.7 & 74.9 & 47.9 \\
  Mask2Former      & 39.4 & 26.5 & 48.8 & 74.9 & 50.6 \\
  \bottomrule
  \end{tabular*}
\end{table}

\begin{table}[!tb]
  \caption{\textbf{Comparison of state-of-the-art supervised panoptic segmentation methods on ACDC for rain.} The first group of rows presents condition-specific expert models trained only on rain, while the second group presents uber models trained on all conditions.}
  \label{table:supp:supervised:panoptic_rain}
  \centering
  \setlength\tabcolsep{2pt}
  \small
  \begin{tabular*}{\linewidth}{l @{\extracolsep{\fill}} ccccc}
  \toprule
  Method           & PQ & PQ$^{things}$ & PQ$^{stuff}$ & SQ & RQ\\
  \midrule
  PanopticFPN      & 46.7 & 37.9 & 53.0 & 77.9 & 57.5 \\
  K-Net            & 48.5 & 29.6 & 62.2 & 78.0 & 60.1 \\
  Panoptic-Deeplab & 52.7 & 37.9 & 63.5 & 80.0 & 63.6 \\
  Mask2Former      & 53.0 & 34.7 & 66.4 & 80.8 & 64.0 \\
  \midrule
  PanopticFPN      & 43.9 & 33.3 & 51.6 & 79.0 & 53.4 \\
  K-Net            & 47.1 & 28.8 & 60.4 & 76.4 & 59.3 \\
  Panoptic-Deeplab & 53.1 & 38.2 & 63.9 & 79.9 & 63.9 \\
  Mask2Former      & 54.2 & 36.3 & 67.3 & 81.2 & 65.2 \\
  \bottomrule
  \end{tabular*}
\end{table}

\begin{table}[!tb]
  \caption{\textbf{Comparison of state-of-the-art supervised panoptic segmentation methods on ACDC for snow.} The first group of rows presents condition-specific expert models trained only on snow, while the second group presents uber models trained on all conditions.}
  \label{table:supp:supervised:panoptic_snow}
  \centering
  \setlength\tabcolsep{2pt}
  \small
  \begin{tabular*}{\linewidth}{l @{\extracolsep{\fill}} ccccc}
  \toprule
  Method           & PQ & PQ$^{things}$ & PQ$^{stuff}$ & SQ & RQ\\
  \midrule
  PanopticFPN      & 44.8 & 36.3 & 51.0 & 74.1 & 55.1 \\
  K-Net            & 48.0 & 32.4 & 59.4 & 74.2 & 59.4 \\
  Panoptic-Deeplab & 51.6 & 38.4 & 61.2 & 81.6 & 61.9 \\
  Mask2Former      & 52.5 & 37.0 & 63.8 & 80.6 & 63.4 \\
  \midrule
  PanopticFPN      & 49.1 & 44.2 & 52.7 & 79.0 & 59.9 \\
  K-Net            & 53.2 & 40.7 & 62.3 & 78.9 & 65.6 \\
  Panoptic-Deeplab & 55.1 & 43.2 & 63.8 & 81.6 & 65.7 \\
  Mask2Former      & 58.6 & 46.0 & 67.7 & 82.2 & 69.8 \\
  \bottomrule
  \end{tabular*}
\end{table}

\subsection{Normal-to-Adverse Adaptation}
\label{supp:sec:training:adaptation}
\subsubsection{Domain adaptive semantic segmentation}

For the comparison in Table~\ref{table:uda:seg_all:adverse}, we use as source-domain model the DeepLabv2~\cite{DeepLab:v2} model that is used as the Cityscapes oracle in AdaptSegNet~\cite{adapt:structured:output:cvpr18}, with a performance of 65.1\% mIoU on the Cityscapes validation set. For all eight unsupervised domain adaptation (UDA) methods that are compared, we use their default training configurations, including the learning rate schedule and the weights of the various losses. The number of training iterations run for each method as well as the number of self-supervised learning rounds that are used by some of the methods are reported in Table~\ref{table:supp:training:uda:all}. For FDA, SIM and MRNet, we run a first training round without self-training followed by a second training round with self-training, as per default implementation of these methods. For FDA, we train three separate models in each training round, one for each different value of the $\beta$ parameter from the set $\{0.01,\,0.05,\,0.09\}$, and use the average prediction of the three models at test time. In all cases, we use the model weights corresponding to the final training iteration for testing.

The same source-domain model is also used for the experiment on adaptation to individual conditions presented in Table~\ref{table:uda:seg_individual}. Again, we use the default training configurations for all examined methods and across all four conditions. The number of training iterations run for each method to adapt to each condition as well as the number of self-supervised learning rounds that are used by some of the methods are reported in Table~\ref{table:supp:training:uda:individual}. For MRNet and fog, the self-supervised training round includes 35k iterations instead of 40k. In addition, for MRNet and rain, the first training round without self-supervised training includes 25k iterations instead of 40k.

\subsubsection{Domain adaptive object detection}
For the comparison in Table~\ref{table:uda:det_all}, we use the representative FCOS and Faster R-CNN as the source-domain models for object detection. For a fair and consistent comparison, each model is trained with a ResNet-50 backbone. For all compared UDA object detection methods, we use their default training configurations for Cityscapes to Foggy Cityscapes adaptation task as it is a common normal-to-adverse setting in existing UDA object detection works. All hyperparameters including the learning rate scheduling, the loss weights and the training iterations are consistent with the original configurations. Following SIGMA~\cite{li2022sigma}, we use the ACDC validation set for each condition to select the model weights for testing.

\subsection{Supervised Learning on Adverse Conditions}
\label{supp:sec:training:supervised}

\subsubsection{Supervised Semantic Segmentation}
For training the four semantic segmentation methods that are compared in Tables~9\ref{table:supervised:all} and \ref{table:supervised:experts_vs_ubers}, we have generally used the default configuration for each method both in the case of condition experts and uber models. For DeepLabv2~\cite{DeepLab:v2}, we use the architecture employed in AdaptSegNet~\cite{adapt:structured:output:cvpr18} in the context of domain adaptation and not the original architecture. We have used the default learning rate schedule for each method, with the base learning rates that are reported in Table~\ref{table:supp:training:supervised}. We generally use 60 training epochs for all four methods, which yields 96k training iterations for uber models and 24k training iterations for condition experts. Exceptions to this rule are RefineNet and fog where we use 30 epochs, DeepLabv2 and fog where we use 45 epochs, DeepLabv2 and night where we use 240 epochs, and the DeepLabv3+ uber model for which we use 30 epochs. For HRNet, we use the snapshot with the best mIoU performance on the respective validation set of ACDC for predicting on the test set, while for the rest of the methods we use the final training snapshot for the same purpose.

\subsubsection{Supervised Instance Segmentation}

For training the four instance segmentation methods that are compared in Tables~\ref{table:supervised:instance_all} and \ref{table:supervised:instance_experts_vs_ubers}, we have generally used the default configuration for each method both in the case of condition experts and uber models. We use the consistent ResNet-50 backbone for each model and train each model on data of each condition for 60 epochs. We use the model weights corresponding to the final training iteration for testing.

\subsubsection{Supervised Panoptic Segmentation}

For training the four panoptic segmentation methods that are compared in Tables~\ref{table:supervised:panoptic_all} and \ref{table:supervised:panoptic_experts_vs_ubers}, we have generally used the default configuration for each method both in the case of condition experts and uber models. We also use the consistent ResNet-50 backbone for each model and train each model on data of each condition for 60 epochs. The model weights corresponding to the final training iteration are reported for testing.

\subsection{Uncertainty-Aware Semantic Segmentation}
\label{supp:sec:training:uass}

We have used the two-head model designed in \cite{simultaneous:segmentation:outliers} and trained it on the entire training set of ACDC for 60 epochs. We use the default learning rate schedule of \cite{simultaneous:segmentation:outliers}, with a base learning rate of $4\times{}10^{-4}$, which is equal to the default. For predicting on the test set, we use the final training snapshot.

\section{Detailed Class-level Results}
\label{supp:sec:results}

We provide class-level performance for the experiments for which only mean performance over all classes is reported in the main paper.

\subsection{Normal-to-Adverse Adaptation}
\label{supp:sec:results:adaptation}

In Tables~\ref{table:supp:uda:fog}--\ref{table:supp:uda:snow}, we present the class-level IoU performance of the UDA semantic segmentation methods that are examined in the setting of adaptation to individual conditions in Table~\ref{table:uda:seg_all:adverse}.

In Tables~\ref{table:supp:uda_det:fog}--\ref{table:supp:uda_det:snow}, the class-wise $AP_{50}^{box}$ for each UDA object detection methods are reported, which corresponds to the results in Table~\ref{table:uda:det_all}.

\subsection{Supervised Learning on Adverse Conditions}
\label{supp:sec:results:supervised}

In Tables~\ref{table:supp:supervised:fog}--\ref{table:supp:supervised:snow}, we present the class-level IoU performance of the supervised semantic segmentation methods that are examined in Table~\ref{table:supervised:experts_vs_ubers}. In particular, we consider the individual conditions of ACDC separately for evaluation, and evaluate on each condition both the respective condition experts that have been trained only on that condition and uber models trained on all conditions.

In Tables~\ref{table:supp:supervised:inst_fog}--\ref{table:supp:supervised:inst_snow}, we present the class-level $AP^{mask}$ performance of the supervised instance segmentation methods that are examined in Table~\ref{table:supervised:instance_experts_vs_ubers}. The performance of condition experts and uber models are reported for each condition respectively.

In Tables~\ref{table:supp:supervised:panoptic_fog}--\ref{table:supp:supervised:panoptic_snow}, we present the detailed performance of the supervised panoptic segmentation methods that are examined in Table~\ref{table:supervised:panoptic_experts_vs_ubers}. The performance of condition experts and uber models are reported for each condition respectively.

\subsection{Evaluation of Pre-trained Models on ACDC}
\label{supp:sec:results:pretrained}

In Tables~\ref{table:supp:pretrained:all}--\ref{table:supp:pretrained:snow}, we present the class-level IoU performance of the externally pre-trained semantic segmentation models that are evaluated in Table~\ref{table:pretrained_seg}.

In Tables~\ref{table:supp:pretrained:inst_all}--\ref{table:supp:pretrained:inst_snow}, we present the class-level $AP^{box}$ and $AP^{mask}$ performance of the externally pre-trained instance segmentation models that are evaluated in Table~\ref{table:pretrained_inst}.

In Tables~\ref{table:supp:pretrained:panoptic_all}--\ref{table:supp:pretrained:panoptic_snow}, we present the detailed performance of the externally pre-trained panoptic segmentation models that are evaluated in Table~\ref{table:pretrained_pan}.

\begin{table*}[!tb]
  \caption{\textbf{Comparison of externally pre-trained semantic segmentation models on the complete test set of ACDC including all conditions.} The three groups of rows present models pre-trained on normal, foggy, and nighttime conditions respectively. CS: Cityscapes, FC: Foggy Cityscapes, FC-DBF: Foggy Cityscapes-DBF, FZ: Foggy Zurich, ND: Nighttime Driving, DZ: Dark Zurich.}
  \label{table:supp:pretrained:all}
  \centering
  \setlength\tabcolsep{2pt}
  \footnotesize
  \begin{tabular*}{\linewidth}{l @{\extracolsep{\fill}} lcccccccccccccccccccc}
  \toprule
  Method & Trained on & \ver{road} & \ver{sidew.} & \ver{build.} & \ver{wall} & \ver{fence} & \ver{pole} & \ver{light} & \ver{sign} & \ver{veget.} & \ver{terrain} & \ver{sky} & \ver{person} & \ver{rider} & \ver{car} & \ver{truck} & \ver{bus} & \ver{train} & \ver{motorc.} & \ver{bicycle} & mIoU\\
  \midrule
  RefineNet & CS & 66.3 & 28.9 & 67.6 & 19.2 & 25.9 & 36.7 & 50.0 & 47.5 & 69.4 & 28.8 & 83.0 & 42.1 & 17.7 & 72.6 & 30.9 & 31.6 & 48.9 & 26.1 & 36.7 &     43.7\\
  DeepLabv2 & CS & 71.9 & 26.2 & 51.1 & 18.8 & 22.5 & 19.7 & 33.0 & 27.7 & 67.9 & 28.6 & 44.2 & 43.1 & 22.1 & 71.2 & 29.8 & 33.3 & 48.4 & 26.2 & 35.8 &     38.0\\
  DeepLabv3+ & CS & 75.1 & 32.8 & 65.9 & 17.5 & 20.2 & 32.2 & 46.7 & 45.2 & 70.5 & 33.5 & 80.9 & 23.9 & 14.7 & 71.5 & 40.1 & 20.3 & 51.2 & 20.2 & 28.8 &   41.6\\
  DANet & CS & 58.0 & 6.0 & 57.3 & 6.8 & 22.3 & 27.7 & 41.3 & 42.1 & 66.4 & 19.9 & 69.2 & 32.2 & 10.2 & 46.5 & 22.4 & 19.1 & 43.1 & 13.2 & 25.5 &           33.1\\
  HRNet & CS & 55.6 & 10.9 & 55.4 & 7.7 & 15.9 & 21.7 & 37.8 & 42.5 & 67.4 & 13.3 & 59.0 & 38.7 & 14.0 & 68.3 & 23.8 & 48.0 & 48.3 & 17.9 & 23.6 &            35.3\\
  \midrule
  SFSU & FC & 72.9 & 28.8 & 68.3 & 19.6 & 23.9 & 37.3 & 49.3 & 47.0 & 60.4 & 33.4 & 72.3 & 43.1 & 14.8 & 72.7 & 31.7 & 31.2 & 47.0 & 25.4 & 35.5 & 42.9\\
  CMAda & FC-DBF+FZ & 79.9 & 32.5 & 69.5 & 14.7 & 24.7 & 41.1 & 53.6 & 51.3 & 67.4 & 34.8 & 83.8 & 49.0 & 19.9 & 77.0 & 34.1 & 38.5 & 51.1 & 29.6 & 42.7 & 47.1\\
  \midrule
  DMAda & ND & 75.3 & 35.5 & 67.4 & 19.2 & 27.1 & 40.0 & 53.7 & 50.9 & 74.6 & 30.9 & 84.9 & 48.8 & 23.1 & 76.6 & 39.7 & 37.4 & 52.5 & 29.1 & 42.1 & 47.9\\
  GCMA & CS+DZ & 79.7 & 48.7 & 71.5 & 21.6 & 29.9 & 42.5 & 56.7 & 57.7 & 75.8 & 39.5 & 87.2 & 57.4 & 29.7 & 80.6 & 44.9 & 46.2 & 62.0 & 37.2 & 46.5 & 53.4\\
  MGCDA & CS+DZ & 76.0 & 49.4 & 72.0 & 11.3 & 21.7 & 39.5 & 52.0 & 54.9 & 73.7 & 24.7 & 88.6 & 54.1 & 27.2 & 78.2 & 30.9 & 41.9 & 58.2 & 31.1 & 44.4 & 48.9\\
  \bottomrule
  \end{tabular*}
\end{table*}

\begin{table*}[!tb]
  \caption{\textbf{Comparison of externally pre-trained semantic segmentation models on ACDC for fog.} The three groups of rows present models pre-trained on normal, foggy, and nighttime conditions respectively. CS: Cityscapes, FC: Foggy Cityscapes, FC-DBF: Foggy Cityscapes-DBF, FZ: Foggy Zurich, ND: Nighttime Driving, DZ: Dark Zurich.}
  \label{table:supp:pretrained:fog}
  \centering
  \setlength\tabcolsep{2pt}
  \footnotesize
  \begin{tabular*}{\linewidth}{l @{\extracolsep{\fill}} lcccccccccccccccccccc}
  \toprule
  Method & Trained on & \ver{road} & \ver{sidew.} & \ver{build.} & \ver{wall} & \ver{fence} & \ver{pole} & \ver{light} & \ver{sign} & \ver{veget.} & \ver{terrain} & \ver{sky} & \ver{person} & \ver{rider} & \ver{car} & \ver{truck} & \ver{bus} & \ver{train} & \ver{motorc.} & \ver{bicycle} & mIoU\\
  \midrule
  RefineNet & CS & 64.4 & 40.0 & 69.6 & 24.2 & 19.7 & 36.5 & 52.7 & 55.2 & 71.1 & 35.4 & 93.9 & 27.4 & 19.2 & 72.7 & 42.0 & 42.1 & 69.3 & 30.3 & 15.8 &     46.4\\
  DeepLabv2 & CS & 66.4 & 31.2 & 26.8 & 22.9 & 18.6 & 8.2 & 32.3 & 10.7 & 70.7 & 39.0 & 31.3 & 17.6 & 41.1 & 65.0 & 30.0 & 34.3 & 18.3 & 42.3 & 29.0 &     33.5\\
  DeepLabv3+ & CS & 82.3 & 57.6 & 61.5 & 18.1 & 16.4 & 33.3 & 49.6 & 54.5 & 76.0 & 44.1 & 90.0 & 9.6 & 28.7 & 69.0 & 35.1 & 34.5 & 28.9 & 41.7 & 37.5 &    45.7\\
  DANet & CS & 52.1 & 14.5 & 49.7 & 5.5 & 16.9 & 30.0 & 47.9 & 51.5 & 72.2 & 23.3 & 80.1 & 24.2 & 3.0 & 44.7 & 32.4 & 27.5 & 65.1 & 10.8 & 7.7 &        34.7\\
  HRNet & CS & 57.3 & 19.3 & 49.1 & 12.8 & 17.8 & 27.3 & 44.0 & 54.7 & 72.8 & 15.5 & 81.7 & 28.3 & 3.9 & 66.6 & 28.4 & 52.0 & 72.7 & 7.2 & 18.1 &     38.4\\
  \midrule
  SFSU & FC & 72.3 & 37.9 & 74.4 & 28.9 & 19.3 & 37.5 & 49.4 & 54.6 & 58.0 & 43.7 & 77.9 & 28.6 & 5.3 & 73.6 & 42.4 & 44.0 & 72.7 & 31.4 & 14.9 &    45.6\\
  CMAda & FC-DBF+FZ & 81.7 & 43.5 & 72.8 & 25.6 & 19.5 & 39.8 & 51.0 & 58.9 & 80.5 & 51.3 & 95.3 & 36.9 & 12.7 & 76.5 & 45.2 & 51.2 & 77.1 & 33.2 & 19.9 &    51.2\\
  \midrule
  DMAda & ND & 75.5 & 44.7 & 72.6 & 26.4 & 20.8 & 38.3 & 52.9 & 57.8 & 75.9 & 38.6 & 96.3 & 35.5 & 26.8 & 75.8 & 47.7 & 50.7 & 73.9 & 35.8 & 17.3 & 50.7\\
  GCMA & CS+DZ & 80.8 & 53.5 & 70.1 & 29.2 & 20.7 & 38.4 & 53.0 & 60.9 & 70.2 & 46.5 & 95.4 & 44.2 & 38.0 & 76.6 & 52.4 & 49.7 & 56.8 & 41.0 & 17.6 & 52.4\\
  MGCDA & CS+DZ & 71.7 & 47.3 & 65.7 & 18.2 & 15.3 & 34.4 & 48.6 & 59.9 & 64.9 & 24.7 & 95.4 & 44.8 & 23.8 & 73.3 & 36.1 & 45.4 & 63.9 & 23.9 & 15.4 & 45.9\\
  \bottomrule
  \end{tabular*}
\end{table*}

\begin{table*}[!tb]
  \caption{\textbf{Comparison of externally pre-trained semantic segmentation models on ACDC for nighttime.} The three groups of rows present models pre-trained on normal, foggy, and nighttime conditions respectively. CS: Cityscapes, FC: Foggy Cityscapes, FC-DBF: Foggy Cityscapes-DBF, FZ: Foggy Zurich, ND: Nighttime Driving, DZ: Dark Zurich.}
  \label{table:supp:pretrained:night}
  \centering
  \setlength\tabcolsep{2pt}
  \footnotesize
  \begin{tabular*}{\linewidth}{l @{\extracolsep{\fill}} lcccccccccccccccccccc}
  \toprule
  Method & Trained on & \ver{road} & \ver{sidew.} & \ver{build.} & \ver{wall} & \ver{fence} & \ver{pole} & \ver{light} & \ver{sign} & \ver{veget.} & \ver{terrain} & \ver{sky} & \ver{person} & \ver{rider} & \ver{car} & \ver{truck} & \ver{bus} & \ver{train} & \ver{motorc.} & \ver{bicycle} & mIoU\\
  \midrule
  RefineNet & CS & 66.5 & 24.0 & 50.3 & 16.9 & 11.6 & 26.4 & 34.2 & 25.5 & 44.2 & 21.6 & 0.1 & 40.8 & 24.8 & 57.4 & 6.8 & 37.3 & 20.5 & 23.9 & 19.1 &     29.0\\
  DeepLabv2 & CS & 77.0 & 22.9 & 56.3 & 13.5 & 9.2 & 23.8 & 22.9 & 25.6 & 41.4 & 16.1 & 2.9 & 44.1 & 17.5 & 64.1 & 11.9 & 34.5 & 42.4 & 22.6 & 22.7 &    30.1\\
  DeepLabv3+ & CS & 73.0 & 20.8 & 50.4 & 22.2 & 5.4 & 22.6 & 31.8 & 23.0 & 42.9 & 16.1 & 6.6 & 19.2 & 11.7 & 48.9 & 0.9 & 13.9 & 42.4 & 10.5 & 13.7 &     25.0\\
  DANet & CS & 67.1 & 4.5 & 46.7 & 5.5 & 5.1 & 13.1 & 29.3 & 19.6 & 36.6 & 15.6 & 0.1 & 29.3 & 12.4 & 29.1 & 4.5 & 12.3 & 9.0 & 10.3 & 13.3 & 19.1\\
  HRNet & CS & 50.0 & 10.1 & 59.9 & 0.7 & 6.0 & 14.2 & 25.6 & 22.3 & 19.1 & 3.4 & 0.1 & 37.6 & 7.9 & 49.4 & 6.9 & 45.9 & 13.9 & 7.8 & 11.3 &    20.6\\
  \midrule
  SFSU & FC & 76.9 & 26.2 & 50.4 & 18.1 & 9.6 & 27.4 & 33.3 & 25.3 & 41.0 & 21.5 & 0.0 & 41.5 & 25.3 & 58.7 & 7.3 & 40.7 & 17.9 & 22.0 & 17.9 & 29.5\\
  CMAda & FC-DBF+FZ & 82.6 & 25.4 & 53.9 & 10.1 & 11.2 & 30.5 & 36.7 & 30.0 & 38.7 & 16.5 & 0.1 & 46.0 & 26.2 & 65.8 & 13.9 & 50.9 & 20.4 & 24.8 & 23.8 & 32.0\\
  \midrule
  DMAda & ND & 74.7 & 29.5 & 49.4 & 17.1 & 12.6 & 31.0 & 38.2 & 30.0 & 48.0 & 22.8 & 0.2 & 47.0 & 25.4 & 63.8 & 12.8 & 46.1 & 23.1 & 24.7 & 24.6 & 32.7\\
  GCMA & CS+DZ & 78.6 & 45.9 & 58.5 & 17.7 & 18.6 & 37.5 & 43.6 & 43.5 & 58.7 & 39.2 & 22.4 & 57.9 & 29.9 & 72.1 & 21.5 & 56.2 & 41.8 & 35.7 & 35.4 & 42.9\\
  MGCDA & CS+DZ & 74.5 & 52.5 & 69.4 & 7.7 & 10.8 & 38.4 & 40.2 & 43.3 & 61.5 & 36.3 & 37.6 & 55.3 & 25.6 & 71.2 & 10.9 & 46.4 & 32.6 & 27.3 & 33.8 & 40.8\\
  DANNet & CS+DZ & 90.7 & 61.1 & 75.5 & 35.9 & 28.8 & 26.6 & 31.4 & 30.6 & 70.8 & 39.4 & 78.7 & 49.9 & 28.8 & 65.9 & 24.7 & 44.1 & 61.1 & 25.9 & 34.5 & 47.6\\
  \bottomrule
  \end{tabular*}
\end{table*}

\begin{table*}[!tb]
  \caption{\textbf{Comparison of externally pre-trained semantic segmentation models on ACDC for rain.} The three groups of rows present models pre-trained on normal, foggy, and nighttime conditions respectively. CS: Cityscapes, FC: Foggy Cityscapes, FC-DBF: Foggy Cityscapes-DBF, FZ: Foggy Zurich, ND: Nighttime Driving, DZ: Dark Zurich.}
  \label{table:supp:pretrained:rain}
  \centering
  \setlength\tabcolsep{2pt}
  \footnotesize
  \begin{tabular*}{\linewidth}{l @{\extracolsep{\fill}} lcccccccccccccccccccc}
  \toprule
  Method & Trained on & \ver{road} & \ver{sidew.} & \ver{build.} & \ver{wall} & \ver{fence} & \ver{pole} & \ver{light} & \ver{sign} & \ver{veget.} & \ver{terrain} & \ver{sky} & \ver{person} & \ver{rider} & \ver{car} & \ver{truck} & \ver{bus} & \ver{train} & \ver{motorc.} & \ver{bicycle} & mIoU\\
  \midrule
  RefineNet & CS & 73.9 & 29.9 & 82.9 & 26.3 & 37.2 & 46.3 & 61.8 & 57.9 & 89.4 & 42.5 & 96.6 & 44.2 & 13.2 & 80.5 & 40.7 & 22.9 & 66.8 & 32.0 & 53.5 &     52.6\\
  DeepLabv2 & CS & 71.2 & 26.7 & 73.8 & 20.8 & 27.1 & 29.9 & 39.3 & 44.4 & 87.3 & 25.2 & 82.0 & 42.0 & 14.3 & 76.2 & 36.3 & 26.6 & 49.8 & 30.3 & 42.2 &     44.5\\
  DeepLabv3+ & CS & 74.4 & 29.8 & 82.3 & 18.1 & 28.8 & 41.7 & 54.3 & 55.6 & 88.7 & 32.8 & 97.2 & 36.7 & 8.5 & 84.7 & 51.7 & 34.0 & 61.5 & 29.7 & 40.0 &    50.0\\
  DANet & CS & 59.9 & 2.4 & 75.9 & 12.9 & 31.5 & 37.7 & 49.5 & 53.3 & 85.5 & 35.5 & 91.1 & 35.4 & 8.4 & 53.5 & 26.0 & 16.4 & 57.8 & 17.9 & 38.9 &    41.5\\
  HRNet & CS & 65.0 & 6.7 & 70.3 & 16.1 & 20.2 & 29.5 & 48.5 & 54.7 & 87.5 & 36.1 & 80.1 & 40.6 & 8.6 & 78.2 & 34.1 & 44.6 & 67.3 & 29.4 & 34.6 &      44.8\\
  \midrule
  SFSU & FC & 74.6 & 29.9 & 81.4 & 24.1 & 33.8 & 46.2 & 59.9 & 56.7 & 86.8 & 40.8 & 93.4 & 46.4 & 15.1 & 80.5 & 40.5 & 18.6 & 65.7 & 33.6 & 52.5 & 51.6\\
  CMAda & FC-DBF+FZ & 78.1 & 34.8 & 80.7 & 18.9 & 33.3 & 50.0 & 63.1 & 62.2 & 87.4 & 38.8 & 96.6 & 51.1 & 16.9 & 83.3 & 37.9 & 21.9 & 68.7 & 36.5 & 55.1 & 53.4\\
  \midrule
  DMAda & ND & 78.3 & 37.7 & 82.5 & 24.2 & 36.8 & 49.0 & 64.5 & 61.5 & 90.6 & 42.8 & 97.3 & 49.6 & 18.2 & 83.4 & 45.1 & 21.6 & 70.2 & 35.2 & 54.8 & 54.9\\
  GCMA & CS+DZ & 81.1 & 48.0 & 84.8 & 25.0 & 37.3 & 49.8 & 66.5 & 66.2 & 92.1 & 43.5 & 97.6 & 54.5 & 20.4 & 85.5 & 47.3 & 34.6 & 71.3 & 40.3 & 56.7 & 58.0\\
  MGCDA & CS+DZ & 80.5 & 46.5 & 79.9 & 16.0 & 28.8 & 44.9 & 60.0 & 61.5 & 90.3 & 44.8 & 97.1 & 51.1 & 23.1 & 82.3 & 33.4 & 30.2 & 69.1 & 36.5 & 53.8 & 54.2\\
  \bottomrule
  \end{tabular*}
\end{table*}

\begin{table*}[!tb]
  \caption{\textbf{Comparison of externally pre-trained semnatic segmentation models on ACDC for snow.} The three groups of rows present models pre-trained on normal, foggy, and nighttime conditions respectively. CS: Cityscapes, FC: Foggy Cityscapes, FC-DBF: Foggy Cityscapes-DBF, FZ: Foggy Zurich, ND: Nighttime Driving, DZ: Dark Zurich.}
  \label{table:supp:pretrained:snow}
  \centering
  \setlength\tabcolsep{2pt}
  \footnotesize
  \begin{tabular*}{\linewidth}{l @{\extracolsep{\fill}} lcccccccccccccccccccc}
  \toprule
  Method & Trained on & \ver{road} & \ver{sidew.} & \ver{build.} & \ver{wall} & \ver{fence} & \ver{pole} & \ver{light} & \ver{sign} & \ver{veget.} & \ver{terrain} & \ver{sky} & \ver{person} & \ver{rider} & \ver{car} & \ver{truck} & \ver{bus} & \ver{train} & \ver{motorc.} & \ver{bicycle} & mIoU\\
  \midrule
  RefineNet & CS & 61.0 & 25.5 & 73.7 & 11.7 & 31.1 & 37.2 & 53.1 & 57.7 & 71.3 & 0.9 & 92.7 & 44.1 & 14.7 & 77.0 & 30.3 & 26.9 & 57.2 & 18.4 & 38.5 &      43.3\\
  DeepLabv2 & CS & 68.5 & 26.6 & 52.7 & 18.8 & 26.9 & 22.2 & 35.7 & 40.7 & 76.5 & 3.6 & 49.9 & 50.4 & 27.1 & 73.7 & 27.6 & 39.1 & 60.9 & 21.1 & 42.5 &      40.2\\
  DeepLabv3+ & CS & 73.9 & 32.6 & 71.3 & 11.1 & 25.6 & 31.4 & 50.6 & 54.4 & 77.8 & 4.1 & 87.0 & 25.1 & 14.6 & 82.7 & 39.5 & 17.2 & 55.2 & 12.0 & 31.2 &    42.0\\
  DANet & CS & 47.6 & 5.4 & 57.5 & 2.9 & 29.1 & 29.3 & 41.4 & 51.2 & 71.1 & 0.5 & 64.8 & 32.7 & 11.7 & 56.5 & 14.5 & 27.9 & 53.7 & 8.1 & 25.9 &     33.3\\
  HRNet & CS & 59.6 & 9.3 & 43.9 & 4.0 & 17.8 & 17.6 & 35.6 & 47.0 & 77.0 & 0.0 & 32.5 & 39.4 & 39.2 & 74.2 & 13.4 & 54.0 & 61.1 & 15.9 & 26.1 &     35.1\\
  \midrule
  SFSU & FC & 64.5 & 24.0 & 72.6 & 10.9 & 28.8 & 37.8 & 54.9 & 58.1 & 62.4 & 0.8 & 78.4 & 44.2 & 9.5 & 76.0 & 29.5 & 25.6 & 55.2 & 16.7 & 37.3 & 41.4\\
  CMAda & FC-DBF+FZ & 74.6 & 31.6 & 73.6 & 9.4 & 30.3 & 43.1 & 61.9 & 61.7 & 75.7 & 0.7 & 93.5 & 53.1 & 19.1 & 79.6 & 29.7 & 31.6 & 61.9 & 22.9 & 50.3 & 47.6\\
  \midrule
  DMAda & ND & 73.6 & 34.4 & 74.9 & 12.3 & 33.4 & 41.1 & 58.4 & 60.1 & 79.9 & 0.6 & 95.4 & 53.1 & 23.0 & 80.4 & 40.3 & 34.5 & 62.9 & 22.7 & 48.6 & 48.9\\
  GCMA & CS+DZ & 79.7 & 49.5 & 75.3 & 17.5 & 37.9 & 43.2 & 59.0 & 61.9 & 78.8 & 2.2 & 95.5 & 62.5 & 33.6 & 83.2 & 42.5 & 43.4 & 72.1 & 32.2 & 51.1 & 53.7\\
  MGCDA & CS+DZ & 80.1 & 49.5 & 70.2 & 6.1 & 27.8 & 39.6 & 55.4 & 58.0 & 76.0 & 0.3 & 95.5 & 57.5 & 35.7 & 81.0 & 28.6 & 48.9 & 70.3 & 27.8 & 50.5 & 50.5\\
  \bottomrule
  \end{tabular*}
\end{table*}

\begin{table*}[!tb]
  \caption{\textbf{Comparison of externally pre-trained instance segmentation models on ACDC including all conditions.} The two groups of rows present performance in $AP^{box}$ and $AP^{mask}$ respectively. CS: Cityscapes.}
  \label{table:supp:pretrained:inst_all}
  \centering
  \setlength\tabcolsep{1pt}
  \small
  \begin{tabular*}{\linewidth}{l @{\extracolsep{\fill}} ccccccccccc}
  \toprule
  Method & Trained on & \ver{person} & \ver{rider} & \ver{car} & \ver{truck} & \ver{bus} & \ver{train} & \ver{motorc.} & \ver{bicycle} & AP \\
  \midrule
  Mask R-CNN          & CS & 12.8 & 6.3 & 29.9 & 8.2 & 8.2 & 5.2 & 6.5 & 4.5 & 10.2 \\
  Cascaded Mask R-CNN & CS & 15.4 & 6.2 & 29.6 & 8.0 & 8.2 & 6.9 & 3.9 & 6.6 & 10.6 \\
  HTC                 & CS & 8.6  & 1.7 & 21.8 & 5.3 & 5.5 & 4.6 & 1.6 & 2.9 & 6.5 \\
  Detectors           & CS & 12.5 & 4.6 & 28.3 & 6.4 & 8.8 & 4.3 & 4.8 & 5.2 & 9.4 \\
  \midrule
  Mask R-CNN          & CS & 9.9 & 3.4 & 27.5 & 8.1 & 8.8 & 5.7 & 4.7 & 2.4 & 8.8 \\
  Cascaded Mask R-CNN & CS & 11.8 & 2.7 & 26.6 & 7.8 & 8.6 & 8.1 & 3.3 & 3.1 & 9.0 \\
  HTC                 & CS & 6.8 & 1.2 & 20.7 & 5.3 & 5.7 & 4.7 & 0.9 & 1.8 & 5.9 \\
  Detectors           & CS & 8.3 & 2.1 & 24.8 & 6.2 & 9.0 & 5.5 & 3.8 & 2.5 & 7.8 \\
  \bottomrule
  \end{tabular*}
\end{table*}

\begin{table*}[!tb]
  \caption{\textbf{Comparison of externally pre-trained instance segmentation models on ACDC for fog.} The two groups of rows present performance in $AP^{box}$ and $AP^{mask}$ respectively. CS: Cityscapes.}
  \label{table:supp:pretrained:inst_fog}
  \centering
  \setlength\tabcolsep{1pt}
  \small
  \begin{tabular*}{\linewidth}{l @{\extracolsep{\fill}} ccccccccccc}
  \toprule
  Method & Trained on & \ver{person} & \ver{rider} & \ver{car} & \ver{truck} & \ver{bus} & \ver{train} & \ver{motorc.} & \ver{bicycle} & AP \\
  \midrule
  Mask R-CNN          & CS & 13.1 & 7.5 & 27.5 & 8.4 & 21.4 & 1.6 & 5.8 & 3.5 & 11.1 \\
  Cascaded Mask R-CNN & CS & 17.2 & 10.0 & 26.8 & 4.6 & 17.6 & 3.1 & 3.3 & 6.3 & 11.1 \\
  HTC                 & CS & 8.9 & 5.3 & 21.8 & 3.1 & 11.8 & 2.9 & 2.4 & 4.9 & 7.6 \\
  Detectors           & CS & 15.7 & 7.5 & 31.1 & 5.4 & 22.5 & 4.0 & 5.9 & 4.4 & 12.1 \\
  \midrule
  Mask R-CNN          & CS & 9.9 & 2.5 & 26.4 & 8.2 & 21.0 & 1.1 & 5.8 & 3.2 & 9.8 \\
  Cascaded Mask R-CNN & CS & 12.3 & 5.3 & 24.6 & 4.1 & 17.0 & 6.6 & 4.8 & 4.0 & 9.8 \\
  HTC                 & CS & 6.8 & 4.0 & 20.6 & 3.2 & 12.9 & 2.1 & 2.4 & 3.7 & 7.0 \\
  Detectors           & CS & 11.5 & 4.4 & 29.0 & 5.5 & 20.3 & 4.0 & 3.1 & 2.8 & 10.1 \\
  \bottomrule
  \end{tabular*}
\end{table*}

\begin{table*}[!tb]
  \caption{\textbf{Comparison of externally pre-trained instance segmentation models on ACDC for nighttime.} The two groups of rows present performance in $AP^{box}$ and $AP^{mask}$ respectively. CS: Cityscapes.}
  \label{table:supp:pretrained:inst_night}
  \centering
  \setlength\tabcolsep{1pt}
  \small
  \begin{tabular*}{\linewidth}{l @{\extracolsep{\fill}} ccccccccccc}
  \toprule
  Method & Trained on & \ver{person} & \ver{rider} & \ver{car} & \ver{truck} & \ver{bus} & \ver{train} & \ver{motorc.} & \ver{bicycle} & AP \\
  \midrule
  Mask R-CNN          & CS & 10.6 & 6.1 & 8.7 & 0.4 & 6.2 & 1.2 & 2.6 & 2.9 & 4.8 \\
  Cascaded Mask R-CNN & CS & 12.1 & 7.1 & 8.6 & 0.1 & 6.9 & 1.5 & 1.1 & 5.2 & 5.3 \\
  HTC                 & CS & 6.3 & 2.0 & 3.0 & 0.1 & 6.1 & 0.5 & 1.5 & 1.6 & 2.6 \\
  Detectors           & CS & 8.6 & 3.6 & 6.1 & 3.5 & 3.4 & 0.2 & 2.4 & 2.4 & 3.8 \\
  \midrule
  Mask R-CNN          & CS & 7.4 & 2.7 & 7.6 & 0.1 & 6.7 & 0.8 & 1.6 & 1.7 & 3.6 \\
  Cascaded Mask R-CNN & CS & 8.3 & 2.7 & 7.7 & 0.0 & 7.3 & 1.5 & 1.3 & 2.3 & 3.9 \\
  HTC                 & CS & 4.6 & 1.3 & 2.7 & 0.1 & 7.7 & 0.2 & 1.3 & 0.9 & 2.3 \\
  Detectors           & CS & 5.2 & 1.3 & 5.3 & 1.5 & 3.5 & 0.2 & 2.4 & 1.2 & 2.6 \\
  \bottomrule
  \end{tabular*}
\end{table*}

\begin{table*}[!tb]
  \caption{\textbf{Comparison of externally pre-trained instance segmentation models on ACDC for rain.} The two groups of rows present performance in $AP^{box}$ and $AP^{mask}$ respectively. CS: Cityscapes.}
  \label{table:supp:pretrained:inst_rain}
  \centering
  \setlength\tabcolsep{1pt}
  \small
  \begin{tabular*}{\linewidth}{l @{\extracolsep{\fill}} ccccccccccc}
  \toprule
  Method & Trained on & \ver{person} & \ver{rider} & \ver{car} & \ver{truck} & \ver{bus} & \ver{train} & \ver{motorc.} & \ver{bicycle} & AP \\
  \midrule
  Mask R-CNN          & CS & 12.7 & 4.5 & 43.9 & 12.9 & 2.7 & 8.8 & 10.2 & 6.4 & 12.8 \\
  Cascaded Mask R-CNN & CS & 14.3 & 2.5 & 44.9 & 13.5 & 2.6 & 13.4 & 7.9 & 7.5 & 13.3 \\
  HTC                 & CS & 9.6 & 0.5 & 34.8 & 8.1 & 4.9 & 9.6 & 1.6 & 4.7 & 9.2 \\
  Detectors           & CS & 14.1 & 5.5 & 41.6 & 11.0 & 2.8 & 10.2 & 8.1 & 10.0 & 12.9 \\
  \midrule
  Mask R-CNN          & CS & 10.5 & 1.9 & 39.9 & 13.1 & 3.4 & 9.9 & 6.6 & 2.9 & 11.0 \\
  Cascaded Mask R-CNN & CS & 12.3 & 0.5 & 40.2 & 13.8 & 3.8 & 13.5 & 6.2 & 3.7 & 11.8 \\
  HTC                 & CS & 8.0 & 0.2 & 33.4 & 8.2 & 4.5 & 9.4 & 0.7 & 2.8 & 8.4 \\
  Detectors           & CS & 10.0 & 2.5 & 35.7 & 11.1 & 4.6 & 13.7 & 6.1 & 3.9 & 10.9 \\
  \bottomrule
  \end{tabular*}
\end{table*}

\begin{table*}[!tb]
  \caption{\textbf{Comparison of externally pre-trained instance segmentation models on ACDC for snow.} The two groups of rows present performance in $AP^{box}$ and $AP^{mask}$ respectively. CS: Cityscapes.}
  \label{table:supp:pretrained:inst_snow}
  \centering
  \setlength\tabcolsep{1pt}
  \small
  \begin{tabular*}{\linewidth}{l @{\extracolsep{\fill}} ccccccccccc}
  \toprule
  Method & Trained on & \ver{person} & \ver{rider} & \ver{car} & \ver{truck} & \ver{bus} & \ver{train} & \ver{motorc.} & \ver{bicycle} & AP \\
  \midrule
  Mask R-CNN          & CS & 18.6 & 18.5 & 38.9 & 7.7 & 9.6 & 6.8 & 7.3 & 6.8 & 14.3 \\
  Cascaded Mask R-CNN & CS & 25.0 & 16.4 & 37.3 & 9.3 & 12.4 & 7.5 & 5.7 & 10.2 & 15.5 \\
  HTC                 & CS & 13.9 & 4.9 & 28.4 & 6.0 & 8.7 & 4.1 & 3.3 & 4.6 & 9.2 \\
  Detectors           & CS & 18.8 & 13.6 & 35.0 & 4.2 & 14.7 & 2.5 & 5.8 & 7.1 & 12.7 \\
  \midrule
  Mask R-CNN          & CS & 15.6 & 13.4 & 35.8 & 7.1 & 10.8 & 7.9 & 8.0 & 4.3 & 12.9 \\
  Cascaded Mask R-CNN & CS & 19.2 & 9.4 & 33.4 & 8.4 & 12.0 & 10.0 & 3.2 & 4.6 & 12.5 \\
  HTC                 & CS & 11.7 & 3.3 & 27.2 & 5.9 & 8.0 & 5.6 & 1.3 & 3.3 & 8.3 \\
  Detectors           & CS & 12.4 & 7.7 & 29.8 & 3.2 & 16.0 & 3.0 & 4.5 & 4.0 & 10.1 \\
  \bottomrule
  \end{tabular*}
\end{table*}

\begin{table}[!tb]
  \caption{\textbf{Comparison of externally pre-trained panoptic segmentation models on ACDC including all conditions.} CS: Cityscapes.}
  \label{table:supp:pretrained:panoptic_all}
  \centering
  \setlength\tabcolsep{2pt}
  \small
  \begin{tabular*}{\linewidth}{l @{\extracolsep{\fill}} cccccc}
  \toprule
  Method      & Trained on & PQ & PQ$^{things}$ & PQ$^{stuff}$ & SQ & RQ\\
  \midrule
  PanopticFPN      & CS    & 13.0 & 11.1 & 14.5 & 69.3 & 17.6 \\
  K-Net            & CS    & 16.7 & 14.6 & 18.3 & 70.2 & 23.3 \\
  Panoptic-Deeplab & CS    & 4.7 & 0.7 & 7.7 & 47.2 & 6.8 \\
  Mask2Former      & CS    & 37.7 & 29.0 & 44.1 & 77.5 & 47.4 \\
  \bottomrule
  \end{tabular*}
\end{table}

\begin{table}[!tb]
  \caption{\textbf{Comparison of externally pre-trained panoptic segmentation models on ACDC for fog.} CS: Cityscapes.}
  \label{table:supp:pretrained:panoptic_fog}
  \centering
  \setlength\tabcolsep{2pt}
  \small
  \begin{tabular*}{\linewidth}{l @{\extracolsep{\fill}} cccccc}
  \toprule
  Method      & Trained on & PQ & PQ$^{things}$ & PQ$^{stuff}$ & SQ & RQ\\
  \midrule
  PanopticFPN      & CS    & 15.9 & 17.1 & 15.0 & 70.2 & 21.5 \\
  K-Net            & CS    & 17.3 & 17.0 & 17.6 & 65.7 & 24.2 \\
  Panoptic-Deeplab & CS    & 6.5  & 1.9  & 9.9  & 40.4 & 9.1 \\
  Mask2Former      & CS    & 42.7 & 30.8 & 51.4 & 79.1 & 52.7 \\
  \bottomrule
  \end{tabular*}
\end{table}

\begin{table}[!tb]
  \caption{\textbf{Comparison of externally pre-trained panoptic segmentation models on ACDC for nighttime.} CS: Cityscapes.}
  \label{table:supp:pretrained:panoptic_night}
  \centering
  \setlength\tabcolsep{2pt}
  \small
  \begin{tabular*}{\linewidth}{l @{\extracolsep{\fill}} cccccc}
  \toprule
  Method      & Trained on & PQ  & PQ$^{things}$ & PQ$^{stuff}$ & SQ & RQ\\
  \midrule
  PanopticFPN      & CS    & 4.0 & 3.2 & 4.8 & 49.4 & 6.0 \\
  K-Net            & CS    & 6.0 & 3.8 & 7.6 & 48.9 & 9.0 \\
  Panoptic-Deeplab & CS    & 1.6 & 0.4 & 2.5 & 29.7 & 2.6 \\
  Mask2Former      & CS    & 19.9 & 17.2 & 22.0 & 71.7 & 26.5 \\
  \bottomrule
  \end{tabular*}
\end{table}

\begin{table}[!tb]
  \caption{\textbf{Comparison of externally pre-trained panoptic segmentation models on ACDC for rain.} CS: Cityscapes.}
  \label{table:supp:pretrained:panoptic_rain}
  \centering
  \setlength\tabcolsep{2pt}
  \small
  \begin{tabular*}{\linewidth}{l @{\extracolsep{\fill}} cccccc}
  \toprule
  Method      & Trained on & PQ & PQ$^{things}$ & PQ$^{stuff}$ & SQ & RQ\\
  \midrule
  PanopticFPN      & CS    & 18.6 & 14.2 & 21.9 & 67.7 & 25.2 \\
  K-Net            & CS    & 23.0 & 18.7 & 26.1 & 69.4 & 31.7 \\
  Panoptic-Deeplab & CS    & 8.3  & 0.5  & 13.9 & 44.6 & 11.7 \\
  Mask2Former      & CS    & 41.4 & 30.8 & 49.2 & 77.0 & 52.1 \\
  \bottomrule
  \end{tabular*}
\end{table}

\begin{table}[!tb]
  \caption{\textbf{Comparison of externally pre-trained panoptic segmentation models on ACDC for snow.} CS: Cityscapes.}
  \label{table:supp:pretrained:panoptic_snow}
  \centering
  \setlength\tabcolsep{2pt}
  \small
  \begin{tabular*}{\linewidth}{l @{\extracolsep{\fill}} cccccc}
  \toprule
  Method      & Trained on & PQ & PQ$^{things}$ & PQ$^{stuff}$ & SQ & RQ\\
  \midrule
  PanopticFPN      & CS    & 13.1 & 12.5 & 13.6 & 62.0 & 17.4 \\
  K-Net            & CS    & 18.7 & 19.8 & 18.0 & 66.6 & 25.8 \\
  Panoptic-Deeplab & CS    & 1.6  & 0.1  & 2.7  & 27.0 & 2.5 \\
  Mask2Former      & CS    & 42.0 & 36.9 & 45.8 & 77.9 & 52.6 \\
  \bottomrule
  \end{tabular*}
\end{table}

\subsection{Uncertainty-aware Semantic Segmentation}
\label{supp:sec:results:uncertainsemseg}

In Tables~\ref{table:supp:uiou:all}--\ref{table:supp:uiou:snow}, we present the class-level average uncertainty-aware IoU (AUIoU) performance of the baselines and oracles that are examined in Table~\ref{table:uiou:baselines}. More specifically, Table~\ref{table:supp:uiou:all} considers methods trained jointly on all conditions of ACDC and also evaluated jointly on all conditions, while Tables~\ref{table:supp:uiou:fog}--\ref{table:supp:uiou:snow} present methods trained and evaluated on individual conditions. The results corresponding to the baseline that uses constant confidence equal to 1 are omitted, as they are identical by definition to IoU results and are thus already included in Table~\ref{table:supervised:all} and Tables~\ref{table:supp:supervised:fog}--\ref{table:supp:supervised:snow}.

\begin{table*}[!tb]
  \caption{\textbf{Uncertainty-aware semantic segmentation baseline results on the complete test set of ACDC including all conditions.} Supervised methods for standard semantic segmentation are trained and evaluated jointly on all conditions for semantic label prediction. Confidence prediction baselines: max-softmax network outputs (Max-Softmax) and ground-truth invalid masks (GT).}
  \label{table:supp:uiou:all}
  \centering
  \setlength\tabcolsep{2pt}
  \footnotesize
  \begin{tabular*}{\linewidth}{l @{\extracolsep{\fill}} lcccccccccccccccccccc}
  \toprule
  Method & Confidence & \ver{road} & \ver{sidew.} & \ver{build.} & \ver{wall} & \ver{fence} & \ver{pole} & \ver{light} & \ver{sign} & \ver{veget.} & \ver{terrain} & \ver{sky} & \ver{person} & \ver{rider} & \ver{car} & \ver{truck} & \ver{bus} & \ver{train} & \ver{motorc.} & \ver{bicycle} & mAUIoU\\
  \midrule
  RefineNet & Max-Softmax & 91.3 & 67.6 & 84.4 & 34.3 & 42.1 & 49.9 & 64.7 & 64.2 & 85.8 & 54.6 & 95.3 & 59.6 & 34.4 & 84.6 & 51.9 & 60.6 & 70.6 & 43.3 & 48.9 &     62.5\\
  RefineNet & GT & 92.9 & 73.1 & 89.1 & 43.1 & 50.7 & 57.0 & 72.9 & 70.7 & 90.1 & 63.4 & 97.7 & 67.6 & 43.1 & 87.3 & 57.3 & 61.4 & 77.1 & 54.1 & 58.3 & 68.8\\
  DeepLabv2 & Max-Softmax & 87.1 & 60.4 & 79.7 & 36.1 & 35.7 & 32.6 & 47.3 & 48.7 & 80.2 & 49.2 & 92.2 & 49.0 & 24.7 & 79.0 & 51.1 & 43.3 & 72.3 & 26.3 & 45.1 &     54.7\\
  DeepLabv2 & GT & 88.5 & 64.4 & 84.2 & 40.9 & 41.8 & 37.8 & 54.0 & 54.2 & 86.4 & 54.9 & 96.0 & 53.6 & 30.3 & 81.8 & 52.5 & 42.7 & 73.6 & 33.3 & 47.6 & 58.9\\
  DeepLabv3+ & Max-Softmax & 92.1 & 71.3 & 88.2 & 49.0 & 47.3 & 54.9 & 68.7 & 65.6 & 88.0 & 60.7 & 96.0 & 65.0 & 33.9 & 87.5 & 66.7 & 72.6 & 81.3 & 43.8 & 55.0 & 67.8\\
  DeepLabv3+ & GT & 93.8 & 76.5 & 91.4 & 56.6 & 55.4 & 62.3 & 75.0 & 72.3 & 91.8 & 66.5 & 98.0 & 72.0 & 41.0 & 89.5 & 71.1 & 74.0 & 86.5 & 55.4 & 63.7 & 73.3\\
  \bottomrule
  \end{tabular*}
\end{table*}

\begin{table*}[!tb]
  \caption{\textbf{Uncertainty-aware semantic segmentation baseline results on ACDC for fog.} Supervised methods for standard semantic segmentation are trained and evaluated on fog for semantic label prediction. Confidence prediction baselines: max-softmax network outputs (Max-Softmax) and ground-truth invalid masks (GT).}
  \label{table:supp:uiou:fog}
  \centering
  \setlength\tabcolsep{2pt}
  \footnotesize
  \begin{tabular*}{\linewidth}{l @{\extracolsep{\fill}} lcccccccccccccccccccc}
  \toprule
  Method & Confidence & \ver{road} & \ver{sidew.} & \ver{build.} & \ver{wall} & \ver{fence} & \ver{pole} & \ver{light} & \ver{sign} & \ver{veget.} & \ver{terrain} & \ver{sky} & \ver{person} & \ver{rider} & \ver{car} & \ver{truck} & \ver{bus} & \ver{train} & \ver{motorc.} & \ver{bicycle} & mAUIoU\\
  \midrule
  RefineNet & Max-Softmax & 92.6 & 71.9 & 82.9 & 40.7 & 35.8 & 42.7 & 62.1 & 62.6 & 84.1 & 64.1 & 97.5 & 45.0 & 26.8 & 77.1 & 57.8 & 59.9 & 79.8 & 35.2 & 33.4 &     60.6\\
  RefineNet & GT & 93.4 & 76.5 & 87.6 & 48.7 & 45.5 & 49.4 & 68.2 & 68.9 & 87.3 & 73.0 & 98.1 & 55.6 & 40.3 & 80.9 & 61.3 & 65.4 & 83.7 & 53.6 & 51.7 & 67.9\\
  DeepLabv2 & Max-Softmax & 89.7 & 63.0 & 79.2 & 39.4 & 25.9 & 25.0 & 41.4 & 46.6 & 82.5 & 66.7 & 95.6 & 36.4 & 35.6 & 72.7 & 49.5 & 29.6 & 44.5 & 29.2 & 33.3 & 51.9\\
  DeepLabv2 & GT & 90.2 & 66.7 & 82.8 & 44.2 & 35.3 & 31.5 & 49.5 & 52.2 & 84.8 & 69.4 & 96.9 & 44.2 & 44.5 & 76.0 & 48.3 & 30.1 & 39.0 & 48.0 & 42.7 & 56.7\\
  DeepLabv3+ & Max-Softmax & 92.9 & 74.8 & 87.2 & 51.3 & 41.7 & 49.9 & 65.6 & 69.8 & 87.1 & 72.3 & 97.6 & 51.9 & 27.1 & 82.8 & 67.4 & 79.1 & 84.1 & 42.6 & 36.4 & 66.4\\
  DeepLabv3+ & GT & 93.9 & 78.3 & 90.0 & 55.5 & 52.0 & 57.9 & 72.3 & 75.9 & 89.2 & 76.6 & 98.4 & 63.2 & 38.5 & 85.0 & 71.7 & 85.1 & 86.7 & 66.0 & 53.3 & 73.1\\
  \bottomrule
  \end{tabular*}
\end{table*}

\begin{table*}[!tb]
  \caption{\textbf{Uncertainty-aware semantic segmentation baseline results on ACDC for nighttime.} Supervised methods for standard semantic segmentation are trained and evaluated on nighttime for semantic label prediction. Confidence prediction baselines: max-softmax network outputs (Max-Softmax) and ground-truth invalid masks (GT).}
  \label{table:supp:uiou:night}
  \centering
  \setlength\tabcolsep{2pt}
  \footnotesize
  \begin{tabular*}{\linewidth}{l @{\extracolsep{\fill}} lcccccccccccccccccccc}
  \toprule
  Method & Confidence & \ver{road} & \ver{sidew.} & \ver{build.} & \ver{wall} & \ver{fence} & \ver{pole} & \ver{light} & \ver{sign} & \ver{veget.} & \ver{terrain} & \ver{sky} & \ver{person} & \ver{rider} & \ver{car} & \ver{truck} & \ver{bus} & \ver{train} & \ver{motorc.} & \ver{bicycle} & mAUIoU\\
  \midrule
  RefineNet & Max-Softmax & 92.3 & 66.4 & 78.6 & 31.8 & 37.2 & 46.2 & 48.3 & 53.3 & 73.5 & 16.9 & 83.6 & 54.9 & 34.6 & 77.4 & 8.5 & 43.1 & 53.6 & 35.2 & 41.6 &     51.4\\
  RefineNet & GT & 93.6 & 72.4 & 88.2 & 42.0 & 53.0 & 55.5 & 61.6 & 61.7 & 89.0 & 31.3 & 97.1 & 63.3 & 41.9 & 80.0 & 18.2 & 50.3 & 60.8 & 49.5 & 51.9 & 61.1\\
  DeepLabv2 & Max-Softmax & 90.2 & 62.2 & 78.6 & 29.9 & 32.9 & 33.7 & 36.5 & 40.3 & 65.6 & 25.2 & 77.9 & 45.2 & 23.2 & 70.2 & 5.0 & 14.6 & 62.1 & 40.3 & 38.8 & 45.9\\
  DeepLabv2 & GT & 90.8 & 65.8 & 87.2 & 37.8 & 45.3 & 43.3 & 48.1 & 49.6 & 87.8 & 37.5 & 97.0 & 51.1 & 29.8 & 74.3 & 17.3 & 17.3 & 63.0 & 51.8 & 43.8 & 54.7\\
  DeepLabv3+ & Max-Softmax & 93.8 & 73.3 & 85.2 & 47.0 & 43.4 & 51.3 & 53.7 & 54.3 & 80.7 & 28.7 & 87.9 & 62.1 & 40.9 & 84.8 & 10.4 & 65.2 & 78.8 & 34.7 & 47.2 & 59.1\\
  DeepLabv3+ & GT & 94.9 & 77.5 & 91.5 & 54.7 & 53.4 & 60.2 & 64.8 & 62.5 & 92.7 & 41.3 & 98.5 & 70.2 & 49.3 & 88.3 & 22.4 & 65.5 & 82.4 & 50.5 & 55.0 & 67.1\\
  \bottomrule
  \end{tabular*}
\end{table*}

\begin{table*}[!tb]
  \caption{\textbf{Uncertainty-aware semantic segmentation baseline results on ACDC for rain.} Supervised methods for standard semantic segmentation are trained and evaluated on rain for semantic label prediction. Confidence prediction baselines: max-softmax network outputs (Max-Softmax) and ground-truth invalid masks (GT).}
  \label{table:supp:uiou:rain}
  \centering
  \setlength\tabcolsep{2pt}
  \footnotesize
  \begin{tabular*}{\linewidth}{l @{\extracolsep{\fill}} lcccccccccccccccccccc}
  \toprule
  Method & Confidence & \ver{road} & \ver{sidew.} & \ver{build.} & \ver{wall} & \ver{fence} & \ver{pole} & \ver{light} & \ver{sign} & \ver{veget.} & \ver{terrain} & \ver{sky} & \ver{person} & \ver{rider} & \ver{car} & \ver{truck} & \ver{bus} & \ver{train} & \ver{motorc.} & \ver{bicycle} & mAUIoU\\
  \midrule
  RefineNet & Max-Softmax & 86.0 & 67.8 & 89.9 & 44.9 & 45.7 & 53.2 & 65.1 & 67.3 & 92.1 & 48.4 & 97.8 & 58.6 & 23.6 & 86.6 & 44.1 & 53.1 & 65.6 & 40.3 & 56.6 &    62.5\\
  RefineNet & GT & 89.5 & 70.8 & 92.1 & 54.1 & 53.2 & 59.9 & 72.6 & 72.3 & 93.9 & 52.1 & 98.4 & 67.4 & 26.6 & 88.7 & 52.4 & 56.4 & 75.5 & 51.4 & 62.9 & 67.9\\
  DeepLabv2 & Max-Softmax & 85.9 & 62.3 & 87.2 & 48.3 & 38.9 & 35.8 & 48.6 & 51.5 & 87.3 & 41.8 & 95.9 & 47.2 & 13.5 & 80.8 & 46.2 & 50.2 & 69.3 & 23.9 & 50.0 & 56.0\\
  DeepLabv2 & GT & 87.8 & 65.1 & 89.4 & 52.1 & 42.5 & 40.2 & 53.7 & 56.1 & 89.6 & 43.6 & 96.8 & 53.4 & 13.8 & 82.7 & 50.2 & 48.1 & 72.9 & 33.3 & 51.4 & 59.1\\
  DeepLabv3+ & Max-Softmax & 91.2 & 75.3 & 92.8 & 62.2 & 53.7 & 60.0 & 71.3 & 72.2 & 93.2 & 50.0 & 98.0 & 65.4 & 30.8 & 90.0 & 63.5 & 77.0 & 83.1 & 48.0 & 63.9 & 70.6\\
  DeepLabv3+ & GT & 93.2 & 78.4 & 94.2 & 68.8 & 60.0 & 66.0 & 75.8 & 78.2 & 94.5 & 52.5 & 98.6 & 72.4 & 35.0 & 91.0 & 70.4 & 80.4 & 87.4 & 58.8 & 69.0 & 75.0\\
  \bottomrule
  \end{tabular*}
\end{table*}

\begin{table*}[!tb]
  \caption{\textbf{Uncertainty-aware semantic segmentation baseline results on ACDC for snow.} Supervised methods for standard semantic segmentation are trained and evaluated on snow for semantic label prediction. Confidence prediction baselines: max-softmax network outputs (Max-Softmax) and ground-truth invalid masks (GT).}
  \label{table:supp:uiou:snow}
  \centering
  \setlength\tabcolsep{2pt}
  \footnotesize
  \begin{tabular*}{\linewidth}{l @{\extracolsep{\fill}} lcccccccccccccccccccc}
  \toprule
  Method & Confidence & \ver{road} & \ver{sidew.} & \ver{build.} & \ver{wall} & \ver{fence} & \ver{pole} & \ver{light} & \ver{sign} & \ver{veget.} & \ver{terrain} & \ver{sky} & \ver{person} & \ver{rider} & \ver{car} & \ver{truck} & \ver{bus} & \ver{train} & \ver{motorc.} & \ver{bicycle} & mAUIoU\\
  \midrule
  RefineNet & Max-Softmax & 89.1 & 59.9 & 83.8 & 25.8 & 43.8 & 53.1 & 72.6 & 69.2 & 88.6 & 43.5 & 96.8 & 65.9 & 11.7 & 85.8 & 39.5 & 48.4 & 74.1 & 36.9 & 48.8 &     59.9\\
  RefineNet & GT & 91.3 & 69.1 & 86.8 & 32.4 & 49.9 & 59.0 & 78.2 & 72.8 & 90.0 & 52.5 & 97.3 & 71.8 & 16.1 & 87.6 & 37.6 & 44.7 & 79.5 & 39.8 & 60.1 & 64.0\\
  DeepLabv2 & Max-Softmax & 89.1 & 61.7 & 82.7 & 26.4 & 40.9 & 35.5 & 56.5 & 54.1 & 85.2 & 39.0 & 95.1 & 55.0 & 25.7 & 84.3 & 38.6 & 53.8 & 77.6 & 29.0 & 49.5 & 56.8\\
  DeepLabv2 & GT & 90.3 & 65.1 & 83.1 & 27.6 & 42.7 & 36.5 & 57.9 & 56.7 & 85.5 & 46.3 & 95.1 & 56.4 & 26.4 & 85.0 & 41.1 & 55.0 & 78.2 & 30.2 & 49.8 & 58.4\\
  DeepLabv3+ & Max-Softmax & 90.6 & 67.0 & 88.8 & 45.1 & 48.9 & 57.8 & 76.6 & 72.9 & 90.8 & 45.7 & 97.0 & 74.8 & 28.4 & 89.2 & 63.3 & 67.8 & 87.8 & 36.7 & 61.1 & 67.9\\
  DeepLabv3+ & GT & 92.9 & 74.0 & 90.4 & 50.3 & 53.9 & 63.4 & 80.5 & 77.4 & 92.2 & 53.6 & 97.6 & 79.2 & 36.6 & 90.9 & 64.4 & 65.9 & 90.0 & 45.2 & 69.8 & 72.0\\
  \bottomrule
  \end{tabular*}
\end{table*}

\section{Additional Details on ACDC Dataset}
\label{supp:sec:dataset}

We provide additional details on the construction and the characteristics of ACDC.

\subsection{Collection}
\label{supp:sec:dataset:collection}

Our recordings were performed in Switzerland. Therefore, the geographic distribution of ACDC is similar to Cityscapes, which was also recorded in central Europe. This eliminates geographic location from the set of factors that introduce a domain shift between Cityscapes and ACDC and allows to study in isolation the effect of visual conditions at time of capture on the performance of semantic segmentation methods, both in the supervised setting and the unsupervised domain adaptation setting.

\subsection{Correspondence Establishment}
\label{supp:sec:dataset:correspondences}

We present in Algorithm~\ref{alg:dp} the dynamic programming algorithm that we use for matching the GPS sequences of adverse-condition recordings and normal-condition recordings of ACDC. The algorithm takes into account the sequential nature of the GPS measurements from the two recordings in computing the correspondence function $A$. In particular, we enforce $k < i \Rightarrow A(k) \leq A(i)$. That is, for a given sample $i$ of the adverse-condition sequence $P$, its matched sample $A(i)$ of the normal-condition sequence $R$ is restricted to not precede in time any sample of $R$ that has been matched to a sample $k$ of $P$ that precedes $i$. This constraint is based on the fact that the routes of the two recordings are driven in the same direction and thus in the same order. Consequently, for routes that contain loops, our formulation prevents the matching of samples that are nearest neighbors but correspond to \emph{different} passes from the same location and are thus potentially associated with different driving directions and 3D rotations of the camera.

\begin{algorithm*}
  \caption{Dynamic programming algorithm for GPS sequence matching}
  \label{alg:dp}
  \begin{algorithmic}[1]
    \Require{Adverse-condition GPS sequence $P=(\mathbf{p}_1,\,\dots,\mathbf{p}_n)$, normal-condition GPS sequence $R=(\mathbf{r}_1,\,\dots,\mathbf{r}_m)$}
    \Ensure{Correspondence function $A: \{1,\,\dots,n\} \to \{1,\,\dots,m\}$}
    \LineComment{Compute pairwise Euclidean distances of GPS samples}
    \State $d_{ij} \gets \|\mathbf{p}_i - \mathbf{r}_j\|,\,1 \leq i \leq n,\,1 \leq j \leq m$
    \LineComment{Compute cost matrix $C$ ($n\times{}m$)}
    \State $C_{1j} \gets d_{1j},\,1 \leq j \leq m$
    \State $C_{ij} \gets \displaystyle\min_{k \leq j}\{C_{i-1,k}\} + d_{ij},\,2 \leq i \leq n,\,1 \leq j \leq m$
    \LineComment{Compute backtracking indices matrix $\alpha$}
    \State $\alpha_{ij} \gets \arg\displaystyle\min_{k \leq j}\{C_{i-1,k}\},\,2 \leq i \leq n,\,1 \leq j \leq m$
    \LineComment{Backtracking}
    \State $A(n) \gets \arg\displaystyle\min_j\{C_{nj}\}$
    \State $A(i) \gets \alpha_{i+1,A(i+1)},\,1 \leq i \leq n-1$
  \end{algorithmic}
\end{algorithm*}

\subsection{Annotation}
\label{supp:sec:dataset:annotation}

\begin{figure*}[!tb]
    \centering
    \begin{tikzpicture}
    \tikzstyle{every node}=[font=\footnotesize]
    \begin{axis}[
      ybar,
      width=\textwidth,
      height=4.5cm,
      xmin=0,
      xmax=26,
      ymin=0,
      ymax=20,
      ymajorgrids=true,
      ylabel={\% of labeled pixels},
      ytick={0,5,10,15,20},
      yticklabels={$0$,$5$,$10$,$15$,$20$},
      xtick={1.5,5,8.5,13.5,18,21,24.5},
      minor xtick={3,7,10,17,19,23},
      xticklabels = {
        flat,
        construction,
        nature,
        vehicle,
        sky,
        object,
        human,
      },
      major x tick style = {opacity=0},
      minor x tick num = 1,
      xtick pos=left,
      every node near coord/.append style={
      anchor=west,
      rotate=90,
      font=\scriptsize,
      }
    ]
    
    \addplot[bar shift=0pt,draw=road,          fill opacity=0.9,fill=road!80!white           , nodes near coords=road                 ] plot coordinates{ ( 1,     0.6158  ) };
    \addplot[bar shift=0pt,draw=sidewalk,      fill opacity=0.8,fill=sidewalk!80!white       , nodes near coords=sidewalk             ] plot coordinates{ ( 2,     3.5701   ) };

    \addplot[bar shift=0pt,draw=building,      fill opacity=0.8,fill=building!80!white       , nodes near coords=building               ] plot coordinates{ ( 4,     6.3418  ) };
    \addplot[bar shift=0pt,draw=wall,          fill opacity=0.8,fill=wall!80!white           , nodes near coords=wall                 ] plot coordinates{ ( 5,     5.7847  ) };
    \addplot[bar shift=0pt,draw=fence,         fill opacity=0.8,fill=fence!80!white          , nodes near coords=fence                ] plot coordinates{ ( 6,     10.7119   ) };

    \addplot[bar shift=0pt,draw=vegetation,    fill opacity=0.8,fill=vegetation!80!white     , nodes near coords=veget.               ] plot coordinates{ ( 8,    13.3094   ) };
    \addplot[bar shift=0pt,draw=terrain,       fill opacity=0.8,fill=terrain!80!white        , nodes near coords=terrain              ] plot coordinates{ ( 9,    5.4041    ) };

    \addplot[bar shift=0pt,draw=car,           fill opacity=0.8,fill=car!80!white            , nodes near coords=car           ] plot coordinates{ ( 11,    1.9046   ) };
    \addplot[bar shift=0pt,draw=train,         fill opacity=0.8,fill=train!80!white          , nodes near coords=train         ] plot coordinates{ ( 12,    0.8356        ) };
    \addplot[bar shift=0pt,draw=truck,         fill opacity=0.8,fill=truck!80!white          , nodes near coords=truck         ] plot coordinates{ ( 13,    3.3480     ) };
    \addplot[bar shift=0pt,draw=bus,           fill opacity=0.8,fill=bus!80!white            , nodes near coords=bus           ] plot coordinates{ ( 14,    1.0454     ) };
    \addplot[bar shift=0pt,draw=bicycle,       fill opacity=0.8,fill=bicycle!80!white        , nodes near coords=bicycle       ] plot coordinates{ ( 15,    6.0160      ) };
    \addplot[bar shift=0pt,draw=motorcycle,    fill opacity=0.8,fill=motorcycle!80!white     , nodes near coords=motorc.    ] plot coordinates{ ( 16,    13.5986        ) };

    \addplot[bar shift=0pt,draw=sky,           fill opacity=0.8,fill=sky!80!white            , nodes near coords=sky                  ] plot coordinates{ ( 18,    13.0483 ) };

    \addplot[bar shift=0pt,draw=pole,          fill opacity=0.8,fill=pole!80!white           , nodes near coords=pole                 ] plot coordinates{ ( 20,    5.0052   ) };
    \addplot[bar shift=0pt,draw=traffic sign,  fill opacity=0.8,fill=traffic sign!80!white   , nodes near coords=traffic sign         ] plot coordinates{ ( 21,    5.6577   ) };
    \addplot[bar shift=0pt,draw=traffic light, fill opacity=0.8,fill=traffic light!80!white  , nodes near coords=traffic light        ] plot coordinates{ ( 22,    5.0104    ) };

    \addplot[bar shift=0pt,draw=person,        fill opacity=0.8,fill=person!80!white         , nodes near coords=person        ] plot coordinates{ ( 24,    3.2333     ) };
    \addplot[bar shift=0pt,draw=rider,         fill opacity=0.8,fill=rider!80!white          , nodes near coords=rider         ] plot coordinates{ ( 25,    5.7358        ) };

    \end{axis}
    \end{tikzpicture}
    \caption{Per-class percentages of labeled pixels that are marked as invalid in the adverse-condition part ACDC.}
    \label{fig:supp:dataset:stats:invalid}
    \vspace{-0.4cm}
\end{figure*}

\begin{figure}
    \centering
    \includegraphics[width=\linewidth]{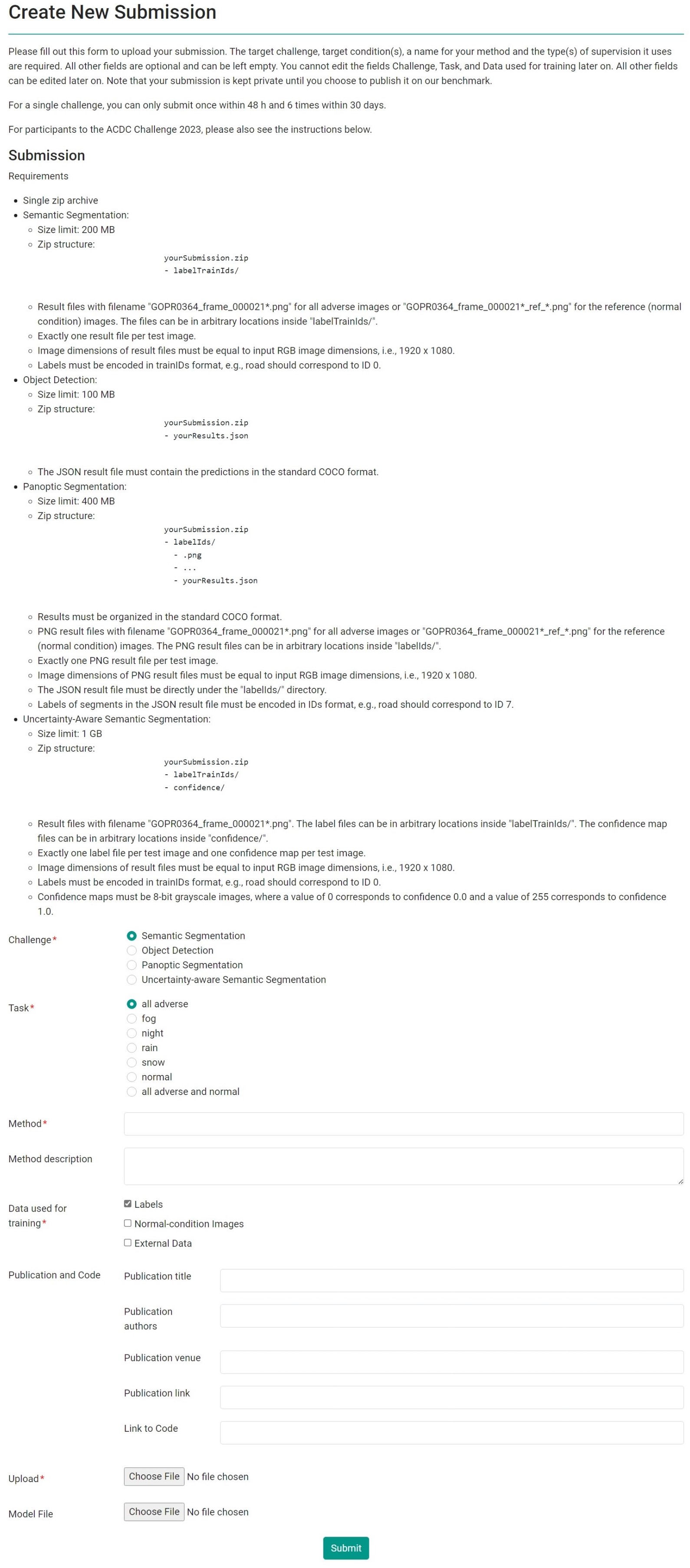}
    \caption{\textbf{The submission page of our benchmark website.} Our evaluation server supports four tasks, i.e.\ semantic segmentation, object detection, panoptic segmentation, and uncertainty-aware semantic segmentation, and seven condition configurations of ACDC, accepting submissions for each of the four individual adverse conditions, for normal conditions, all adverse conditions, and all adverse and normal conditions. Best viewed on a screen.}
    \label{fig:supp:dataset:server:screenshot}
\end{figure}

In Fig.~\ref{fig:supp:dataset:stats:invalid}, we show for the adverse-condition part of ACDC (4006 images) the percentage of the pixels of each semantic class that are marked as invalid in the ground-truth invalid mask $J$. For the majority of the classes, a notable percentage of more than 5\% of the pixels are labeled as invalid, which demonstrates the ability of our specialized annotation protocol with privileged information to assign a legitimate semantic label even to invalid regions with ambiguous semantic content.

\begin{table*}[tb]
  \caption{\textbf{Overall annotation statistics for ACDC.} We report the total number of pixels assigned to a legitimate semantic label (Labeled) and of pixels not assigned to any semantic label (Unlabeled) as well as the respective percentages for the adverse-condition part of the dataset with 4006 images (Adverse), the normal-condition part of the dataset with 1503 images (Normal), and their union (Full).}
  \label{table:supp:dataset:stats:total}
  \centering
  \setlength\tabcolsep{3pt}
  \small
  \begin{tabular}{lcccccc}
  \toprule
  & \multicolumn{2}{c}{Adverse} & \multicolumn{2}{c}{Normal} & \multicolumn{2}{c}{Full} \\
  & \#pixels & \% of pixels & \#pixels & \% of pixels  & \#pixels & \% of pixels \\
  \midrule
  Labeled & $7.682\times{}10^9$ & 92.47 & $3.015\times{}10^9$ & 96.77 & $10.697\times{}10^9$ & 93.64\\
  -out of which Valid & $7.055\times{}10^9$ & 84.93 & $3.015\times{}10^9$ & 96.77 & $10.071\times{}10^9$ & 88.16\\
  -out of which Invalid & $0.627\times{}10^9$ & 7.54 & 0 & 0 & $0.627\times{}10^9$ & 5.48\\
  Unlabeled & $0.625\times{}10^9$ & 7.53 & $0.101\times{}10^9$ & 3.23 & $0.726\times{}10^9$ & 6.36\\
  \midrule
  Total & $8.307\times{}10^9$ & 100.00 & $3.117\times{}10^9$ & 100.00 & $11.423\times{}10^9$ & 100.00\\
  \bottomrule
  \end{tabular}
\end{table*}

The total number of annotated pixels in ACDC is presented in Table~\ref{table:supp:dataset:stats:total}. Note that labeled pixels that are marked as valid in the ground-truth invalid masks $J$ constitute ca.\ 85\% of the pixels in the adverse-condition part of the dataset. From the remaining 15\% of pixels in the adverse-condition part that did not receive a legitimate semantic label in stage~1 of the annotation because of their ambiguity, it was possible to label \emph{half} of them (7.5\%) with a legitimate semantic label in stage~2 of the annotation, by making use of the additional privileged information in the form of corresponding normal-condition images and original adverse-condition videos. Note that for stage~2 of the annotation, we explicitly set the time budget (excluding quality control) to 20 minutes and asked the annotators to prioritize labeling of (i) traffic participants and (ii) distant and/or unclear objects that were affected the most by the adverse conditions at the time of capture. The normal-condition part of the dataset was annotated with the standard semantic segmentation protocol, so none of the labeled pixels is invalid. It is worth noting that, probably due to the normality of the conditions in this part of the dataset, a slightly larger percentage of pixels (96.8\%) was possible to label compared to the adverse-condition part. Overall, more than 10 billion pixels in ACDC have received panoptic labels.

\subsection{Evaluation Server}
\label{supp:sec:dataset:server}

We have implemented a website and evaluation server for the ACDC benchmark and have made it publicly available at \url{https://acdc.vision.ee.ethz.ch}. An indicative screenshot from the submission page of the website is provided in Fig.~\ref{fig:supp:dataset:server:screenshot}.